\documentclass[journal]{IEEEtran}
\usepackage{amsmath,graphicx}
\usepackage{lineno,hyperref}
\usepackage{multirow}
\usepackage{wasysym}
\usepackage{tabularx}
\usepackage{rotating}
\usepackage{mathrsfs}

\usepackage{makecell}
\usepackage{algorithm}
\usepackage{algorithmic}
\usepackage{xcolor}
\usepackage{epstopdf}
\usepackage{subfigure}
\usepackage{epsfig}
\usepackage{xcolor}
\usepackage{amsfonts}
\usepackage{bm}
\usepackage{color}
\usepackage{subfigure}
\usepackage{epsfig}
\usepackage{booktabs}
\usepackage{array}
\usepackage{amsmath}
\usepackage{cite}
\usepackage[greek,english]{babel} 

\usepackage{bbding}

\ifCLASSINFOpdf
\else
\fi
\begin{document}
%
\title{Multi-scale Attentive Image De-raining Networks via Neural Architecture Search}
%
%
%

\author{Lei~Cai, Yuli~Fu, Wanliang~Huo, Youjun~Xiang, Tao~Zhu, Ying~Zhang, Huanqiang~Zeng,~\IEEEmembership{Senior Member,~IEEE}, and Delu Zeng
	\thanks{This work was supported in part by the National Key Research and Development Program of China under the grant 2021YFE0205400, in part by the Natural Science Foundation of Guangdong Province under the Grant 2019A1515010861, in part by Guangzhou Technical Project under the Grant 201902020008, in part by NSFC under the Grant 61471174, and in part by Fundamental Research Program of Guangdong under the Grant 2020B1515310023. \emph{(Corresponding author: Youjun~Xiang)}.}
	\thanks{L. Cai, Y. L. Fu, W. L. Huo, Y. J. Xiang, Y. Zhang, and D. L. Zeng are with School of Electronic and Information Engineering, South China University of Technology, Guangzhou 510640, China (e-mail: eelcai@mail.scut.edu.cn; fuyuli@scut.edu.cn; eewanlianghuo@mail.scut.edu.cn; yjxiang@scut.edu.cn; eezhangying9804@mail.scut.edu.cn, dlzeng@scut.edu.cn).}
	\thanks{T. Zhu is with School of Electronic Engineering and Automation, Guilin University of Electronic Technology, Guilin, 541004, China (e-mail: zt21@guet.edu.cn).}
	\thanks{H. Q. Zeng is with School of Engineering and School of Information Science and Engineering, Huaqiao University, China (e-mail: zeng0043@hqu.edu.cn).}
	\thanks{Copyright~\textcircled{c} 2022 IEEE. Personal use of this material is permitted. However, permission to use this material for any other purposes must be obtained from the IEEE by sending an email to pubs-permissions@ieee.org.}
}

\maketitle

\begin{abstract}
Multi-scale architectures and attention modules have shown effectiveness in many deep learning-based image de-raining methods. However, manually designing and integrating these two components into a neural network requires a bulk of labor and extensive expertise. In this article, a high-performance~\emph{multi-scale attentive neural architecture search} (MANAS) framework is technically developed for image de-raining. The proposed method formulates a new~\emph{multi-scale attention search space} with multiple flexible modules that are favorite to the image de-raining task. Under the search space, multi-scale attentive cells are built, which are further used to construct a powerful image de-raining network. The internal multi-scale attentive architecture of the de-raining network is searched automatically through a gradient-based search algorithm, which avoids the daunting procedure of the manual design to some extent.
Moreover, in order to obtain a~\emph{robust} image de-raining model, a practical and effective~\emph{multi-to-one training strategy} is also presented to allow the de-raining network to get sufficient background information from multiple rainy images with the same background scene, and meanwhile, multiple loss functions including~\emph{external loss},~\emph{internal loss},~\emph{architecture regularization loss}, and~\emph{model complexity loss} are jointly optimized to achieve robust de-raining performance and controllable model complexity. Extensive experimental results on both synthetic and realistic rainy images, as well as the down-stream vision applications (\emph{i.e.}, objection detection and segmentation) consistently demonstrate the superiority of our proposed method. The code is publicly available at https://github.com/lcai-gz/MANAS.
\end{abstract}


%
\IEEEpeerreviewmaketitle

\section{Introduction}
\label{sec:introduction}
%
%
%
%

\IEEEPARstart{W}{hen} people take photos in an outdoor environment, the captured images could suffer from visibility degradation owning to various bad weather conditions such as rain~\cite{Jiang2020Decomposition,Zhu2021Learning}, haze~\cite{Yin2020Color,Agrawal2021Dense}, or snow~\cite{Jaw2021DesnowGAN}. Particularly, on a rainy day, the presence of rain can seriously degrade the visual quality of outdoor images, and as a result affects the visual authenticity of human perception and the performance of many outdoor vision systems~\cite{Cai2021Joint}. In this context, image de-raining becomes very necessary and meaningful, it can serve as an essential pre-step for various computer vision tasks such as object detection~\cite{Chen2020High}, object segmentation~\cite{Liu2020Guided}, autonomous driving~\cite{Feng2021Deep}, and more. However, image de-raining remains a formidable challenge due to its intractability and complexity, \emph{e.g.}, the rain streaks in the real world could present different shapes, sizes, density, orientations,~\emph{etc}. Therefore, how to design an effective de-raining algorithm is crucial and has drawn much attention in the computer vision field.

The current research on rain removal can be divided into two directions, namely video de-raining methods~\cite{Chen2004Detection,Garg2006Photorealistic,Zhang2006Rain,Brewer2008Using,Barnum2010Analysis,Bossu2011Rain,Chen2013Generalized,Eigen2013Restoring,Wei2017Should} and image de-raining methods~\cite{Luo2015Removing,Li2016Rain,Kim2013Single,Fu2017Clearing,Yang2017Deep,Fan2018Residual,Zhang2018Density,Kang2012Automatic,Chen2014Visual,Li2018Recurrent,Ren2019Progressive,Yang2019Joint,Zhang2019Image,Jiang2020Multi,Deng2020Detail,Wang2020Model,Yang2019Scale,Yang2020Single,Wang2020Rethinking,Lin2020Rain,Lin2020Utilizing,Fu2019Lightweight,Li2019Heavy,Hu2019Depth,Du2020Conditional,Zamir2021Multi,Ahn2021Eagnet,Hu2021Single,Wang2021Context,Fu2021Successive,Zhang2022Single,Wei2021DerainCycleGAN}. The former need to exploit the additional~\emph{temporal correlation} in adjacent multi-frames to help restore the pixel corrupted by rain. However, once the temporal information becomes unreliable (\emph{e.g.}, unstable video) or unavailable (\emph{e.g.}, single frame), the performance of the existing video de-raining methods could be severely degraded. Furthermore, for rain removal from videos acquired from a moving camera, the video de-raining methods could also produce poor performance~\cite{Chen2013Generalized}. Unlike video de-raining methods, image de-raining methods aim to remove rain directly from a given rainy image and restore its clean background, without the temporal information. 
\begin{figure*}[!t]
	\centering
	\subfigure[\scriptsize Rainy Images]{
		\begin{minipage}[t]{0.195\linewidth}
			\centering
			\centerline{\includegraphics[width=1.4in]{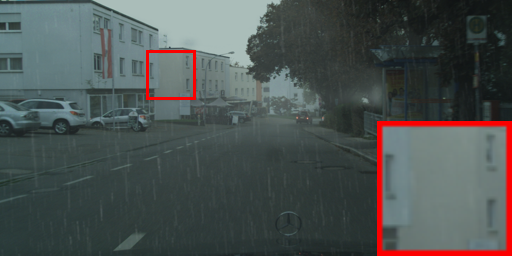}}
			\vspace{2pt}
			\centerline{\includegraphics[width=1.4in]{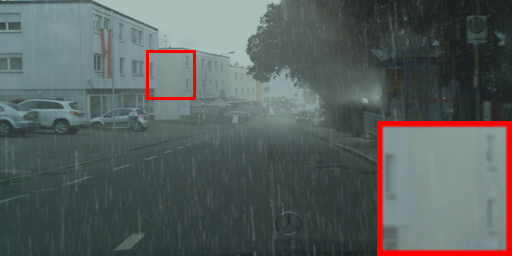}}
			\vspace{2pt}
			\centerline{\includegraphics[width=1.4in]{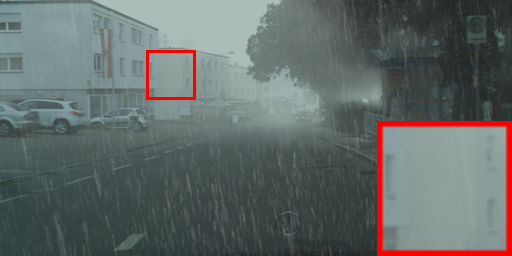}}
			\vspace{4pt}
		\end{minipage}%
	}%
	\subfigure[\scriptsize CLEARER~\cite{Gou2020CLEARER}]{
		\begin{minipage}[t]{0.195\linewidth}
			\centering
			\centerline{\includegraphics[width=1.4in]{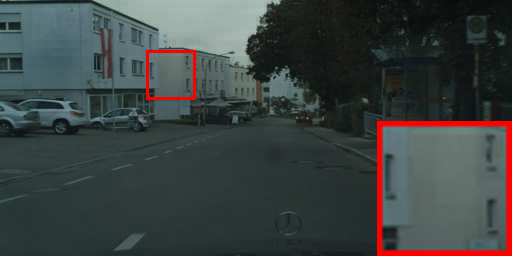}}
			\vspace{2pt}
			\centerline{\includegraphics[width=1.4in]{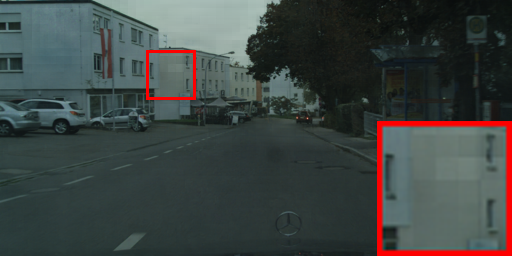}}
			\vspace{2pt}
			\centerline{\includegraphics[width=1.4in]{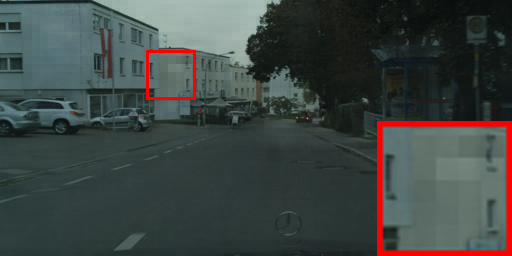}}
			\vspace{4pt}
		\end{minipage}%
	}%
	\subfigure[\scriptsize DGNL-Net-fast~\cite{Hu2021Single}]{
		\begin{minipage}[t]{0.195\linewidth}
			\centering
			\centerline{\includegraphics[width=1.4in]{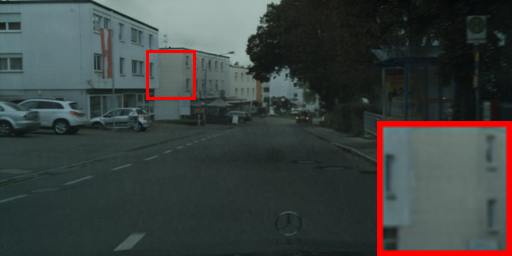}}
			\vspace{2pt}
			\centerline{\includegraphics[width=1.4in]{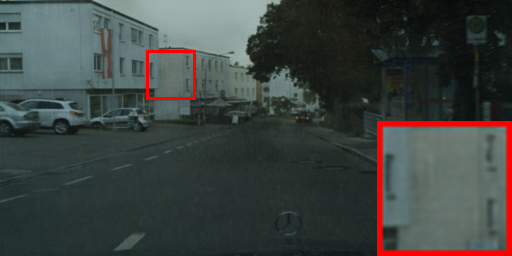}}
			\vspace{2pt}
			\centerline{\includegraphics[width=1.4in]{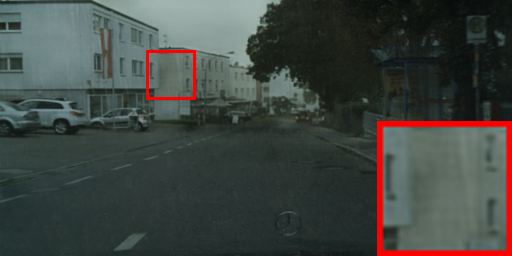}}
			\vspace{4pt}
		\end{minipage}%
	}%
	\subfigure[\scriptsize \textbf{MANAS (Ours)}]{
		\begin{minipage}[t]{0.195\linewidth}
			\centering
			\centerline{\includegraphics[width=1.4in]{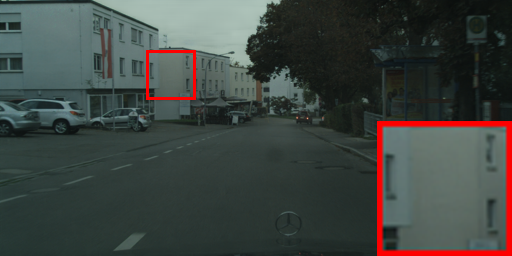}}
			\vspace{2pt}
			\centerline{\includegraphics[width=1.4in]{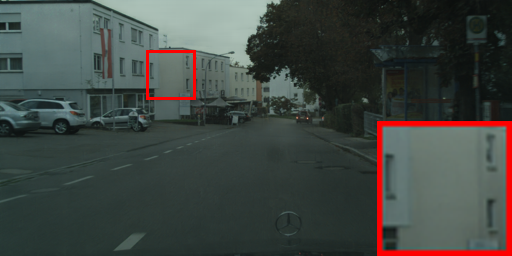}}
			\vspace{2pt}
			\centerline{\includegraphics[width=1.4in]{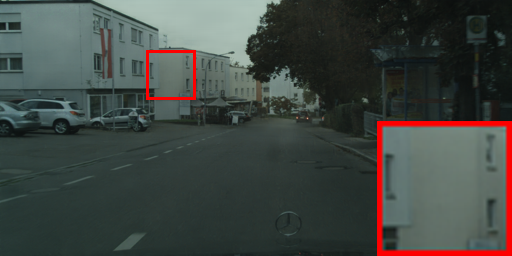}}
			\vspace{4pt}
		\end{minipage}%
	}%
	\subfigure[\scriptsize Ground Truth]{
		\begin{minipage}[t]{0.195\linewidth}
			\centering
			\centerline{\includegraphics[width=1.4in]{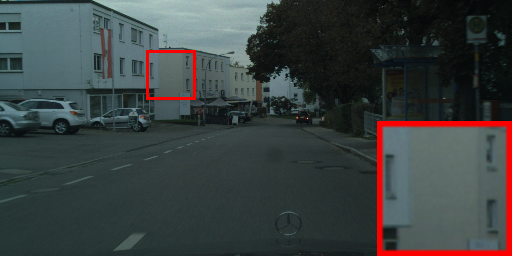}}
			\vspace{2pt}
			\centerline{\includegraphics[width=1.4in]{./Figs/gt_1.png}}
			\vspace{2pt}
			\centerline{\includegraphics[width=1.4in]{./Figs/gt_1.png}}
			\vspace{4pt}
		\end{minipage}%
	}%
	\centering
	\caption{Three examples of rainy images (a) come from RainCityscapes~\cite{Hu2021Single} dataset, as well as their corresponding de-rained images derived by CLEARER~\cite{Gou2020CLEARER} (b), DGNL-Net-fast~\cite{Hu2021Single} (c), and our MANAS (d). One can see that our MANAS method can deliver a more robust de-raining effect.}
	\label{fig:consistency}
\end{figure*}


Over the past decade, a variety of successful algorithms, ranging from early hand-crafted prior-based methods~\cite{Luo2015Removing,Li2016Rain,Kim2013Single,Kang2012Automatic,Chen2014Visual} to the latest deep learning-based methods~\cite{Que2021Attentive,Wang2021Deep,Fu2017Clearing,Yang2017Deep,Zhang2018Density,Li2018Recurrent,Ren2019Progressive,Yang2019Joint,Zhang2019Image,Jiang2020Multi,Deng2020Detail,Wang2020Model,Yang2019Scale,Yang2020Single,Wang2020Rethinking,Lin2020Rain,Fu2019Lightweight,Li2019Heavy,Hu2019Depth,Du2020Conditional,Zamir2021Multi,Ahn2021Eagnet,Hu2021Single,Cai2021Joint}, have been proposed to handle the image de-raining task.  
The prior-based de-raining methods first leverage effective regularizers to characterize the property of the background and rain streak layers, and then separate them by solving an objective function with proper optimization algorithms. The representative methods include dictionary learning~\cite{Kang2012Automatic}, discriminative sparse coding~\cite{Luo2015Removing}, Gaussian mixture model~\cite{Li2016Rain}, nonlocal means filter~\cite{Kim2013Single}, and more. However, these conventional methods have limited capability to model and remove rain, thus cannot satisfy the increasingly complex de-raining task~\cite{Hu2021Single}. For deep learning-based de-raining methods, the major development over the years lies on the design of various effective neural network architectures. Among them, the~\emph{multi-scale architecture}~\cite{Yang2017Deep,Zhang2018Density,Yang2019Joint,Fu2019Lightweight,Zhang2019Image,Wang2020Rethinking,Wang2021Deep,Cai2021Joint} and the~\emph{attention module}~\cite{Qian2018Attentive,Que2021Attentive,Ahn2021Eagnet,Hu2019Depth,Hu2021Single} or their combination~\cite{Jiang2020Multi,Qian2018Attentive,Shao2021Uncertainty,Zamir2021Multi} have been demonstrated effective in boosting the performance. Take a few examples, Zhang~\emph{et al.}~\cite{Zhang2018Density} developed a multi-stream densely connected de-raining network to efficiently learn features at different scales. Fu~\emph{et al.}~\cite{Fu2019Lightweight} proposed light pyramid image de-raining networks, according to a multi-scale Gaussian Laplacian pyramid decomposition technology. Ahn~\emph{et al.}~\cite{Ahn2021Eagnet} recently designed an image de-raining network by employing the elementwise attentive gating block as a basic unit. To achieve depth-attentive image de-raining, Hu~\emph{et al.}~\cite{Hu2019Depth,Hu2021Single} successfully designed an end-to-end de-raining network, where a depth-guided attention mechanism was introduced into their network to learn the depth-attentional features. 
Rather than separately exploiting the multi-scale architecture or the attention module, Jiang~\emph{et al.}~\cite{Jiang2020Multi} constructed a multi-scale pyramid structure and further introduced U-shaped residual attention blocks to obtain a multi-scale progressive fusion network for image de-raining.


Even though multi-scale architectures and attention modules are demonstrated to be helpful for image de-raining, it is difficult to design their neural architectures and most of them rely heavily on human design. Such a hand-crafted manner has the following limitations.
Firstly, manual design is very labor-intensive, especially for those networks with multi-scale architectures and attention modules. Secondly, one more daunting issue of the manual design is unknown when to fuse the low-scale and high-scale features or what kind of attention operation should be applied. Thirdly, the networks with multi-scale architectures and attention modules are generally more complex than the plain ones, thus it is necessary to find an elegant trade-off between the performance and the model complexity~\cite{Gou2020CLEARER}. Obviously, it is very difficult to tackle the above limitations through human design.

In this work, we propose 
a novel~\emph{Multi-scale Attentive Neural Architecture Search} (MANAS) framework to automatically search and integrate the multi-scale attentive neural architectures for image de-raining. 
The motivation behind our MANAS method is based on the consideration that both the multi-scale architecture and the attention module can strengthen the representation ability of a neural network, but from a different aspect.
The multi-scale architecture can not only capture the global structure of input images but also retain their local details~\cite{Gou2020CLEARER}, while attention modules can well
handle long-range dependencies which enables the neural network
to give more attention to useful information within a
context~\cite{Ma2020Auto}.
Regarding the image de-raining task, it would be better to combine them, rather than separately employing them, so that their merits are taken and their demerits are overcome. Unfortunately, to the best of our knowledge, the current multi-scale attentive neural networks (\emph{e.g.},~\cite{Jiang2020Multi,Qian2018Attentive,Shao2021Uncertainty,Zamir2021Multi}) for image de-raining are almost designed and integrated in a hand-crafted manner. 

On the other hand, when a well-designed de-raining network is ready to be used for image de-raining, how to train it is also considerably important. Under the premise of sufficient training data, it is highly expected to design an appropriate training strategy to encourage the outputs of the network to be close to the real clean background~\cite{Chang2019Single}. In previous image de-raining methods, the most possible way is to estimate~\emph{mean square error} (MSE) or other intricate distance measurements on one-to-one image pairs. 
As shown in Fig.~\ref{fig:consistency} (b) and (c), two recently-developed image de-raining methods, namely~CLEARER~\cite{Gou2020CLEARER} and DGNL-Net-fast~\cite{Hu2021Single}, which were trained with MSE loss on one-to-one image pairs, are applied to remove the rain streaks and fog from three synthetic rainy images shown in Fig.~\ref{fig:consistency} (a). As can be seen, although these input rainy images have the same background scene, parts of their corresponding de-rained results suffer from undesirable local artifacts or color distortion or preserve some rain streaks. This means that these two de-raining methods, which were trained by a one-to-one training paradigm, are not robust enough for image de-raining. 
See again Fig.~\ref{fig:consistency} (a), as different degrees of rain (\emph{i.e.}, the rain density, intensity, shapes, sizes, and orientations vary) could be presented in multiple images with the same static background, thus we know that these input rainy images should have the same clean background after de-raining. To this end, based on this prior knowledge, we present a~\emph{multi-to-one training strategy} to allow our de-raining network to get sufficient background information from multiple rainy images with the same background, and simultaneously impose~\emph{external} and~\emph{internal} constraints on the de-raining model to direct it to be robust.

The main contributions of this article are:
\begin{enumerate}
	\item [$\bullet$]{We make the first attempt to incorporate the~\emph{multi-scale architecture search} and the~\emph{attention search} into a unified~\emph{neural architecture search} (NAS) framework to automatically discover the high-performance multi-scale attentive image de-raining networks. This proposed framework, called MANAS, is expected to get rid of the daunting designing procedure of the multi-scale attentive neural networks to some extent. 	
	}
	\item [$\bullet$]{We propose a novel multi-scale attention search space and integrate it into a differentiable form through a continuous relaxation operation. Moreover, we also introduce a gradient-based search algorithm to search for the best paths in the de-raining network to determine when to fuse the high-scale and low-scale features and what kind of attention operation should be applied.
	}
	\item [$\bullet$]{We provide a practical and effective~\emph{multi-to-one training strategy} for image de-raining, 
	where multiple loss functions, including~\emph{external loss},~\emph{internal loss},~\emph{architecture regularization loss}, and~\emph{model complexity loss}, are jointly used to train our model to achieve robust de-raining performance and controllable model complexity.
	Both quantitative and qualitative results show that our proposed method outperforms multiple state-of-the-art image de-raining methods. 
	 
	}
\end{enumerate}

\section{Related Work}
\label{sec:related}

\subsection{Image De-raining Methods}
\label{ssec:image}
Existing image de-raining methods can be broadly divided into two categories: 1) hand-crafted prior-based de-raining methods and 2) hand-crafted neural network-based de-raining methods.

\subsubsection{Hand-crafted Prior-based De-raining Methods} 
The hand-crafted prior-based de-raining methods mainly rely on the statistic analysis of rain streaks and background scenes. Such methods usually use effective regularizers to characterize the rain streak and background layers, and then separate them by solving an objective function with proper optimization algorithms. Specifically, Kang~\emph{et al.}~\cite{Kang2012Automatic} decomposed high frequency parts of rainy images into rain and rain-free components by conducting dictionary learning and sparse coding. Luo~\emph{et al.}~\cite{Luo2015Removing} presented a discriminative sparse coding framework for layer decomposition. Li~\emph{et al.}~\cite{Li2016Rain} used a Gaussian mixture model to approximate priors of the background and rain-streak layers and then decomposed them with a Maximum A Posterior (MAP). Kim~\emph{et al.}~\cite{Kim2013Single} proposed a nonlocal means filter-based method to remove rain streaks within the detected rain streak regions. However, the above prior-based methods are insufficient in characterizing the background and rain-streak layers, thus cannot clearly remove rain from diverse rainy images. 


\begin{figure*}[!t]
	\centering
	\includegraphics[width=1.0\linewidth]{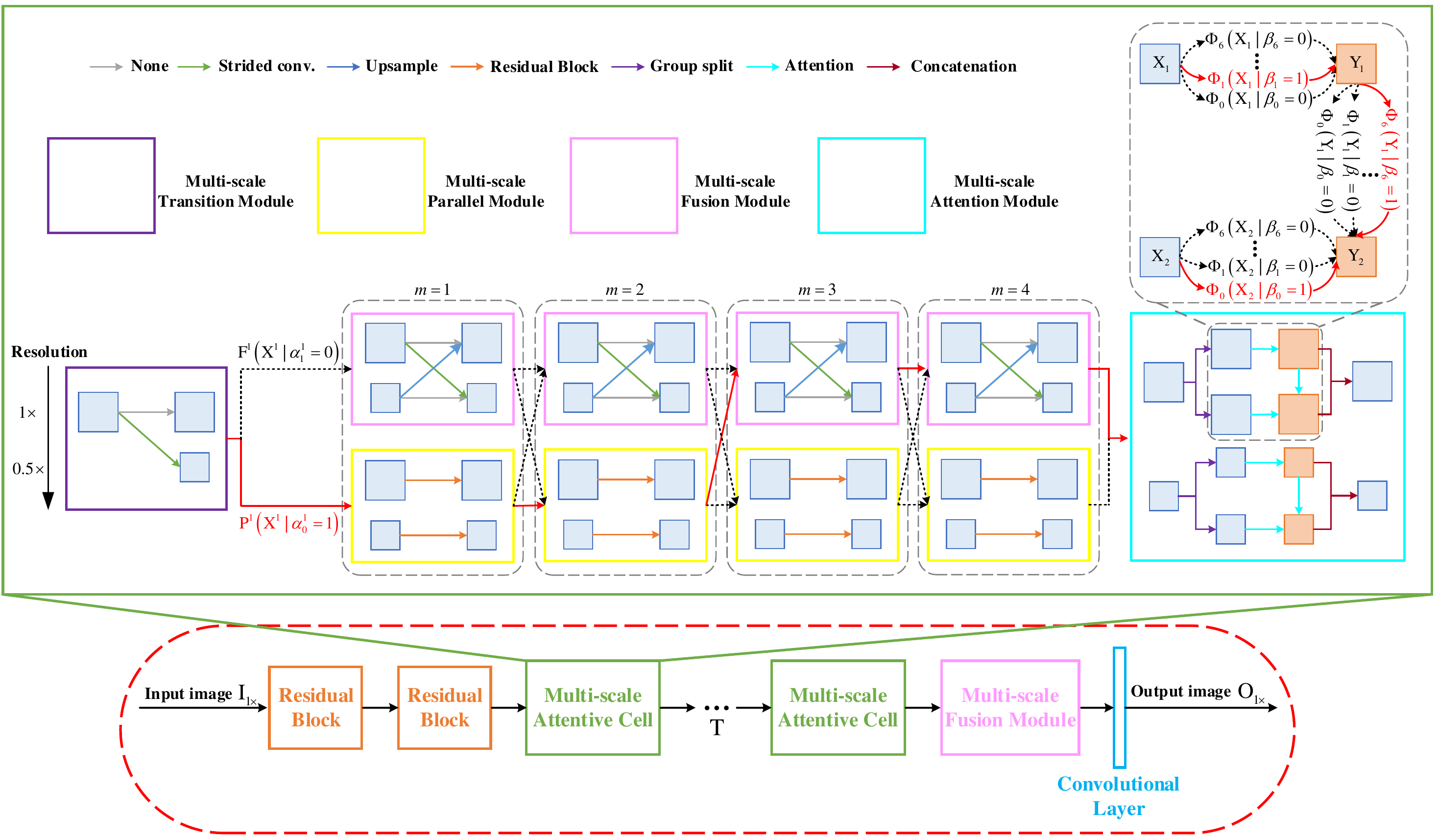}
	\caption{The proposed multi-scale attentive neural architecture search (MANAS) framework. In brief, multiple typical modules, including the multi-scale transition module, multi-scale parallel module, multi-scale fusion module, and multi-scale attention module, are utilized to build multi-scale attentive cells. These built cells are further employed to construct a multi-scale attentive image de-raining network (\emph{Bottom}). We take the first cell in the de-raining network as a showcase (\emph{Upper}). It starts with a multi-scale transition module, followed by a structure with $4$ columns (each column contains a multi-scale fusion module represented by the pink box or a multi-scale parallel module represented by the yellow box), and finally ends with a multi-scale attention module represented by the cyan box. In the first cell, there are two scales of resolution,~\emph{e.g.}, $1\times$ represents the original resolution, whereas $0.5\times$ represents the original resolution reduced by $0.5$. The red arrows in the cell denote the paths to be searched.
	}
	\label{fig:framework}
\end{figure*}

\subsubsection{Hand-crafted Neural Network-based De-raining Methods}
The major development of deep-learning-based de-raining methods lies on the manual design of various neural networks~\cite{Yang2017Deep,Yang2019Joint,Fu2019Lightweight,Zhang2019Image,Wang2020Rethinking,Wang2021Deep,Cai2021Joint,Que2021Attentive,Ahn2021Eagnet,Hu2019Depth,Qian2018Attentive,Zamir2021Multi,Jiang2020Multi,Hu2021Single,Shao2021Uncertainty,Zamir2021Multi}. In these de-raining networks, multi-scale architecture~\cite{Zhang2018Density,Yang2017Deep,Yang2019Joint,Yasarla2019Uncertainty,Fu2019Lightweight,Zhang2019Image,Wang2020Rethinking,Wang2021Deep,Cai2021Joint} and attention module~\cite{Ren2020Dually,Que2021Attentive,Ahn2021Eagnet} or their combination~\cite{Jiang2020Multi,Qian2018Attentive,Shao2021Uncertainty,Zamir2021Multi} have been widely incorporated into the design of various image de-raining networks and have achieved the promising performance. The related works are succinctly
described as follows:~\emph{1) Image de-raining using multi-scale neural networks:} Wang~\emph{et al.}~\cite{Wang2021Deep} developed a modeling Haze-Like effect-based deep neural network for image de-raining, in which a SSP module~\cite{He2019Spatial} is introduced to extract multi-scale features to help remove the haze-like effect. Zhang~\emph{et al.}~\cite{Zhang2018Density} proposed a multi-stream densely connected network to effectively exploit multi-scale features to characterize rain streaks with different scales and shapes. Fu~\emph{et al.}~\cite{Fu2019Lightweight} developed light-weight Pyramid networks by introducing multi-scale decomposition of Gaussian Laplacian pyramid, which simplifies the learning
process of image de-raining. Yasarla~\emph{et al.}~\cite{Yasarla2019Uncertainty} presented an uncertainty guided multi-scale residual learning network to learn the rain content at different scales.~\emph{2) Image de-raining using attentive networks:} Jiang~\emph{et al.}~\cite{Jiang2020Decomposition} proposed an improved attention-guided de-raining network for rain streak removal, where a mixed attention block was designed to guide the fusion of rain layers by focusing on the local and global overlaps. Zhu~\emph{et al.}~\cite{Zhu2021Learning} developed a non-local guided attention module in their de-raining network to learn attentional non-local features for the global residual image prediction. Ahn~\emph{et al.}~\cite{Ahn2021Eagnet} leveraged the elementwise attentive gating block as a basic unit to construct an elementwise attentive gating network for single image de-raining. Que~\emph{et al.}~\cite{Que2021Attentive} proposed an attentive composite residual network for image de-raining, where a channel-wise attention mechanism is built using a squeeze-and-excitation (SE)-Res2Net. To be well adapted to the stochastic distribution of real rain streaks, Wang~\emph{et al.}~\cite{Wang2019Spatial} developed a spatial attentive network to learn the representative and discriminative features in a local-to-global attentive manner.
\emph{3) Image de-raining using multi-scale attentive networks:} To fully explore the advantages of both the multi-scale architecture and the attention mechanism for image de-raining, several multi-scale attentive networks have been manually designed. Jiang~\emph{et al.}~\cite{Jiang2020Multi} proposed a~\emph{Multi-scale Progressive Fusion Network} (MSPFN) for image de-raining. In this method, they constructed a multi-scale pyramid structure, and further incorporated the attention module to guide the fine fusion of the correlated information from different scales. Qian~\emph{et al.}~\cite{Qian2018Attentive} present an attentive generative adversarial network for raindrop removal, where visual attention was injected into both the multi-scale generative and discriminative networks. To guide the removal of raindrops at different scales, Shao~\emph{et al.}~\cite{Shao2021Uncertainty} designed a multi-scale pyramid structure and an iterative attention mechanism. Zamir~\emph{et al.}~\cite{Zamir2021Multi} designed a multi-stage progressive network architecture for image de-raining. In their method, an encoder-decoder is applied in the earlier stage to learn multi-scale contextual information. And a supervised attention module is plugged between very two stages to enable progressive learning.

Although the above de-raining networks can provide a promising solution for image de-raining, their internal neural architectures are all designed and integrated in a hand-crafted manner, which requires a bulk of labor and extensive expertise. Different from these hand-crafted de-raining networks, we employ~\emph{neural architecture search} (NAS) to construct the high-performance multi-scale attentive neural networks for image de-raining, where the internal~\emph{multi-scale attentive neural architecture} of our de-raining network can be searched and integrated automatically through a gradient-based search algorithm.

\subsection{Neural Architecture Search}
\label{ssec:neural}
NAS aims to automate the procedure of discovering neural architectures using advanced search algorithms such as evolutionary algorithm (EA), reinforcement learning (RL), gradient-based algorithm, and more. Benefit from the development of neural architecture search~\cite{Suganuma2018Exploiting,Guo2020Hierarchical,Zhang2020Memory,Zhang2020Memory,Gou2020CLEARER,Ma2020Auto,Fang2020Densely}. Some works make attempts to handle the image restoration task by automatically searching effective neural architectures. The seminal work of NAS-based image restoration is E-CAE~\cite{Suganuma2018Exploiting}. It exploits an evolutionary algorithm to search for good architectures of the Convolutional Auto-Encoders (CAEs). Later, Gou~\emph{et al.}~\cite{Gou2020CLEARER} designed a multi-resolution search space, and used a data-driven strategy to search their image restoration networks. Zhang~\emph{et al.}~\cite{Zhang2020Memory} proposed the Hierarchical NAS (HiNAS) for image de-noising. Their method adopted a gradient-based search algorithm and created a hierarchical search space by employing operations with adaptive receptive field. Similarly, Guo~\emph{et al.}~\cite{Guo2020Hierarchical} developed a Hierarchical Neural Architecture Search (HNAS) method for image super-resolution. 
Different from the above-mentioned methods, we make the first attempt to integrate the~\emph{multi-scale architecture search} and the~\emph{attention search} into a unified NAS framework to find the high-performance multi-scale attentive neural network for image de-raining. Moreover, we specifically design a practical and effective~\emph{multi-to-one training strategy} for training our model. Such a training paradigm is also the first trial in the neural architecture search.

\section{Proposed Multi-scale Attention Neural Architecture Search for Image De-raining}
\label{sec:proposed}
\begin{figure*}[t]
	\centering
	\includegraphics[width=0.95\linewidth]{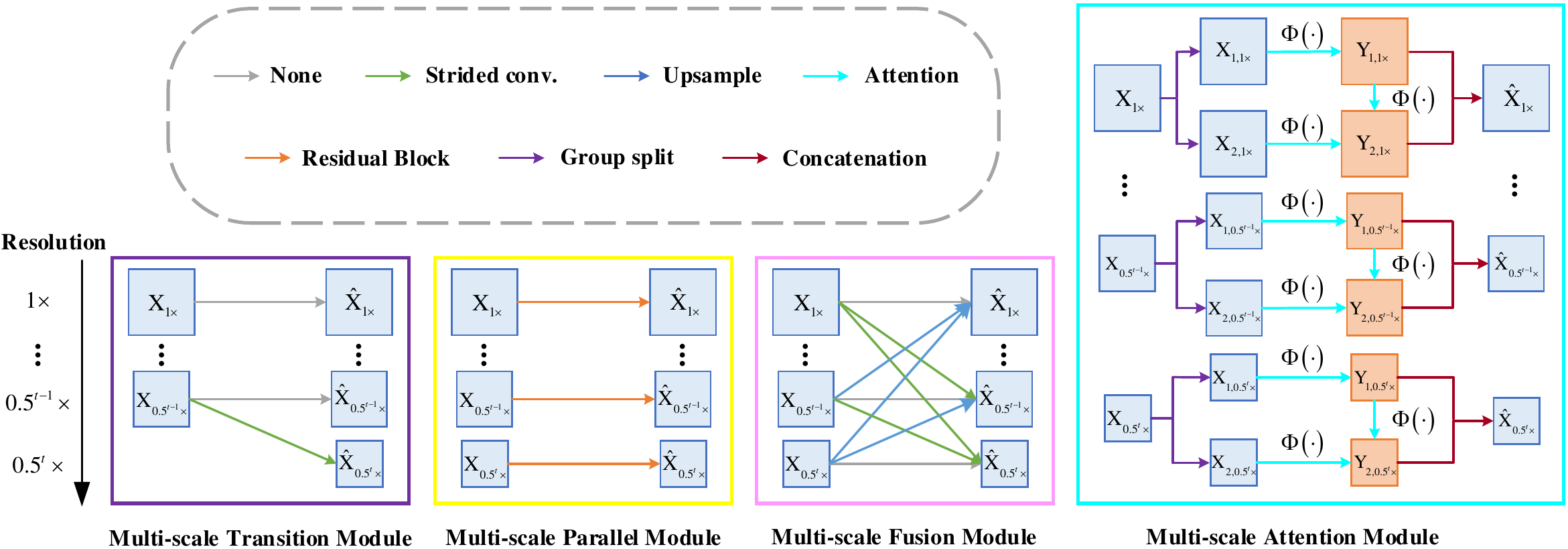}
	\caption{The generalized structures of the multi-scale transition module, multi-scale parallel module, multi-scale fusion module, and multi-scale attention module, respectively. $1\times$ represents the original resolution of features, whereas $0.5^{t}\times$ denotes the original resolution reduced by $0.5^{t}$. $\Phi(\cdot)$ denotes an attention operation, and $\{\textmd{Y}_{1},\textmd{Y}_{2}\}$ represent the intermediate features resulted from $\Phi(\cdot)$. 
	}
	\label{fig:modules}
\end{figure*}

In this section, we first define a~\emph{multi-scale attention search space} that involves three basic terms, namely~\emph{multi-scale module},~\emph{multi-scale attentive cell}, and~\emph{multi-scale attentive image de-raining network}.
Then, we elaborate the method of the multi-scale attentive neural architecture search.
Finally, we provide a~\emph{multi-to-one training strategy} for image de-raining. The framework of the proposed~\emph{multi-scale attentive neural architecture search} (MANAS) method is shown in Fig.~\ref{fig:framework}.

\subsection{Multi-scale Attention Search Space}
\label{ssec:space}
We introduce four typical multi-scale modules, namely multi-scale~\emph{transition} module, multi-scale~\emph{parallel} module, multi-scale~\emph{fusion} module, and multi-scale~\emph{attention} module, into our search space. Their generalized structures are depicted in Fig.~\ref{fig:modules}.


\subsubsection{Multi-scale Transition Module}
As shown in Fig.~\ref{fig:modules}, the multi-scale transition module is responsible for transmitting the input features to lower-scale ones through a~\emph{strided} convolution operation and simultaneously keeping the original resolution scale unchanged in the horizontal direction via none operation. Mathematically, it can be formulated as: 
\begin{equation}
\begin{split}
	\label{equ:esim1}
&	\bm{\hat{\textmd{X}}}_{1\times}=\bm{\textmd{X}}_{1\times}, \\
&	\;\;\;\;\;\;\;\;\vdots \\
&	\bm{\hat{\textmd{X}}}_{0.5^{t-1}\times}=\bm{\textmd{X}}_{0.5^{t-1}\times}, \\
&	\bm{\hat{\textmd{X}}}_{0.5^{t}\times}=f_{s}(\bm{\textmd{X}}_{0.5^{t-1}\times}), 
\end{split}
\end{equation}
where $\bm{\textmd{X}}_{1\times}$ represents the features at the original resolution, $\bm{\textmd{X}}_{0.5^{t-1}\times}$ denotes the features whose resolution is only $0.5^{t-1}$ times to the original resolution, and $t\in\{1,\dots,T\}$ denotes the index of the multi-scale attentive cell in the de-raining network. $f_{s}(\cdot)$ denotes a strided convolution operation to reduce the resolution of input features. For example, for the first cell (\emph{i.e.}, $t=1$), the input features are with single resolution scale, but after the multi-scale transition module, there will produce two resolution scales,~\emph{i.e.}, $\bm{\hat{\textmd{X}}}_{1\times}=\bm{\textmd{X}}_{1\times}$ and $\bm{\hat{\textmd{X}}}_{0.5\times}=f_{s}(\bm{\textmd{X}}_{1\times})$. Likewise, for the second cell (\emph{i.e.}, $t=2$), its input features come from the first cell have two resolution scales, but after the multi-scale transition module, there will bring in three resolution scales,~\emph{i.e.}, $\bm{\hat{\textmd{X}}}_{1\times}=\bm{\textmd{X}}_{1\times}$, $\bm{\hat{\textmd{X}}}_{0.5\times}=\bm{\textmd{X}}_{0.5\times}$, $\bm{\hat{\textmd{X}}}_{0.25\times}=f_{s}(\bm{\textmd{X}}_{0.5\times})$.



\subsubsection{Multi-scale Parallel Module}
The multi-scale parallel module is responsible for processing the multi-scale features in parallel, where the resolution of features in each parallel line remains unchanged via a residual block~\cite{He2016Deep}, which can be formulated as:
\begin{equation}
	\begin{split}
		\label{equ:esim2}
		&	\bm{\hat{\textmd{X}}}_{1\times}=\mathcal{F}(\bm{\textmd{X}}_{1\times}) + \bm{\textmd{X}}_{1\times}, \\
		&	\;\;\;\;\;\;\;\;\vdots \\
		&	\bm{\hat{\textmd{X}}}_{0.5^{t-1}\times}=\mathcal{F}(\bm{\textmd{X}}_{0.5^{t-1}\times}) + \bm{\textmd{X}}_{0.5^{t-1}\times}, \\
		&	\bm{\hat{\textmd{X}}}_{0.5^{t}\times}=\mathcal{F}(\bm{\textmd{X}}_{0.5^{t}\times}) + \bm{\textmd{X}}_{0.5^{t}\times},  
	\end{split}
\end{equation}
where $\mathcal{F}(\cdot)$ represents a residual function which is learned by $3\times3$ convolution layers.

\subsubsection{Multi-scale Fusion Module}
The multi-scale fusion module is responsible for fusing the features at different resolution scales via strided convolution, upsampling, and none operation. The formulation of this module can be written as:
\begin{small}
\begin{equation}
	\begin{split}
		\label{equ:esim3}
		&	\!\bm{\hat{\textmd{X}}}_{1\times}\!=\!\sigma\big(\bm{\textmd{X}}_{1\times}\!+\!{\rm US }(\bm{\textmd{X}}_{0.5^{1}\times})\!+\!\cdots\!+\!{\rm US}(\bm{\textmd{X}}_{0.5^{t}\times})\big), \\
		&	\;\;\;\;\;\;\;\;\vdots \\
		&\!\bm{\hat{\textmd{X}}}_{0.5^{t-1}\!\times\!}\!=\!\sigma\big(f_{s}(\bm{\textmd{X}}_{1\!\times\!})\!+\!\cdots\!+\!f_{s}(\bm{\textmd{X}}_{0.5^{t-2}\times})\!+\!\bm{\textmd{X}}_{0.5^{t-1}\times}\!+\!{\rm US}(\bm{\textmd{X}}_{0.5^{t}\times})\big), \\
		&	\bm{\hat{\textmd{X}}}_{0.5^{t}\times}\!=\!\sigma\big(f_{s}(\bm{\textmd{X}}_{1\times})\!+\!\cdots\!+\!f_{s}(\bm{\textmd{X}}_{0.5^{t-1}\times})\!+\!\bm{\textmd{X}}_{0.5^{t}\times}\big),
	\end{split}
\end{equation}
\end{small}
where $\rm US(\cdot)$ represents an up-sampling operation to the input features, and $\sigma(\cdot)$ is a Rectified Linear Unit (ReLU)~\cite{Krizhevsky2012Imagenet} for non-linearity.


\subsubsection{Multi-scale Attention Module}
The multi-scale attention module is responsible for adaptively attending to important regions within a context at different resolution scales. 
As shown in Fig.~\ref{fig:modules}, 
the multi-scale attention module first divides the input features $\bm{\textmd{X}}$ (here we omit its scale subscript for brevity) into two groups along the channel dimension,  
leading to 
two groups of split features,~\emph{i.e.}, $\{\bm{\textmd{X}}_1,\bm{\textmd{X}}_2\}$. 
After that, a series of attention operations are applied on $\{\bm{\textmd{X}}_1,\bm{\textmd{X}}_2\}$, producing the corresponding intermediate features $\{\bm{\textmd{Y}}_1, \bm{\textmd{Y}}_2\}$:
\begin{equation}
	\begin{split}
		\label{equ:esim4}
		&\bm{\textmd{Y}}_{1}=\Phi(\bm{\textmd{X}}_{1}), \\
		&\bm{\textmd{Y}}_{2}=\Phi(\bm{\textmd{X}}_{2})+\Phi(\bm{\textmd{Y}}_{1}), 
	\end{split}
\end{equation} 
where $\Phi(\cdot)$ denotes an attention operation. Finally, the resulting intermediates features $\{\bm{\textmd{Y}}_1, \bm{\textmd{Y}}_2\}$ are concatenated along the channel axis to yield the attentive output:
\begin{equation}
	\begin{split}
		\label{equ:esim5}
		\bm{\hat{\textmd{X}}}=f_{1c}([\bm{\textmd{Y}}_{1};\bm{\textmd{Y}}_{2};\overline{\bm{\textmd{Y}}}]), 
	\end{split}
\end{equation}
where $\overline{\bm{\textmd{Y}}}=\bm{\textmd{Y}}_{1}+\bm{\textmd{Y}}_{2}$, $\big[;\big]$ represents the concatenation along the channel axis, and $f_{1c}(\cdot)$ represents the mapping function learned by $1 \times 1$ convolutional layers. 

For the multi-scale paradigm, the attentive outputs are: 
\begin{equation}
	\begin{split}
		\label{equ:esim6}
		&\bm{\hat{\textmd{X}}}_{1\times}=f_{1c}([\bm{\textmd{Y}}_{1,1\times};\bm{\textmd{Y}}_{2,1\times};\overline{\bm{\textmd{Y}}}_{1\times}]),  \\
		&	\;\;\;\;\;\;\;\;\vdots \\
		& \bm{\hat{\textmd{X}}}_{0.5^{t-1}\times}=f_{1c}([\bm{\textmd{Y}}_{1,0.5^{t-1}\times};\bm{\textmd{Y}}_{2,0.5^{t-1}\times};\overline{\bm{\textmd{Y}}}_{0.5^{t-1}\times}]), \\
		& \bm{\hat{\textmd{X}}}_{0.5^{t}\times}=f_{1c}([\bm{\textmd{Y}}_{1,0.5^{t}\times};\bm{\textmd{Y}}_{2,0.5^{t}\times};\overline{\bm{\textmd{Y}}}_{0.5^{t}\times}]).
	\end{split}
\end{equation} 
It is worthwhile of mentioning that all of the attention operations in this module can be searched automatically, thus an attention search sub-space used in~\cite{Ma2020Auto} is also incorporated into our search space. The formulation of each attention operation in the sub-space is documented in Table.~\ref{tab:attention}.
\begin{table}[t]
	\footnotesize
	\renewcommand{\arraystretch}{1.3}
	\caption{Various attention operations in attention search sub-space, where $\bm{\textmd{Z}}_{i}$ denotes the $i$-th split group features $\bm{\textmd{X}}_{i}$ or its corresponding intermediate features $\bm{\textmd{Y}}_{i}$, $\bm{\textmd{Z}}_{i,\text{avg}}$ denotes the global spatial average pooled features from the input $\bm{\textmd{Z}}_{i}$, $\otimes$ denotes the Kronecker Products, $\bm{\textmd{Z}}_{i,\text{max}}$ denotes the global spatial max pooled features, $\phi(\cdot)$ denotes a multilayer perceptron, $\delta(\cdot)$ denotes the sigmoid activation function, $f_{d}(\cdot)$ denotes a mapping function learned by a $3\times3$ depth-wise convolutional layer.}
	\label{tab:attention}
	\tabcolsep0.03cm
	\centering
	\begin{tabular}{c|cc}
		\Xhline{1.3pt}
		Name               &   Definition     \\
		\hline
		Channel Attention V1                  &   $\Phi_{0}(\bm{\textmd{Z}}_{i})=\delta(\phi(\bm{\textmd{Z}}_{i,\text{avg}})) \otimes \bm{\textmd{T}}_{i}$  \\
		\hline
		Channel Attention V2                   &   $\Phi_{1}(\bm{\textmd{Z}}_{i})=\delta\big(\phi(\bm{\textmd{Z}}_{i,\text{avg}})+\phi(\bm{\textmd{Z}}_{i,\text{max}})\big) \otimes \bm{\textmd{Z}}_{i}$   \\
		\hline
		Spatial Attention                   &   $\Phi_{2}(\bm{\textmd{Z}}_{i})=\delta(f_{3c}([\bm{\textmd{Z}}_{i,\text{avg}};\bm{\textmd{Z}}_{i,\text{max}}])) \otimes \bm{\textmd{Z}}_{i}$   \\
		\hline
		Normalization Attention                   &   $\Phi_{3}(\bm{\textmd{Z}}_{i})=\delta(f_{d}(\bm{\textmd{Z}}_{i})) \otimes \bm{\textmd{Z}}_{i}$  \\
		\hline
		Convolutional Block Attention                &  $\Phi_{4}(\bm{\textmd{Z}}_{i})=\Phi_{3}(\Phi_{2}(\bm{\textmd{Z}}_{i}))$    \\
		\hline
		Identity Attention                   &   $\Phi_{5}(\bm{\textmd{Z}}_{i})=\bm{\textmd{Z}}_{i} \otimes \textbf{1}$  \\
		\hline
		Zero Attention                   &   $\Phi_{6}(\bm{\textmd{Z}}_{i})=\bm{\textmd{Z}}_{i} \otimes \textbf{0}$   \\
		\Xhline{1.3pt}  
	\end{tabular}
\end{table}
\subsubsection{Multi-scale Attentive Cells}
With the above-defined multi-scale modules, we further use them to build our multi-scale attentive cells. As the upper part of Fig.~\ref{fig:framework} shows, where we take the first cell as a showcase, the multi-scale attentive cell starts with a multi-scale transition module, followed by a structure with $4$ columns (each column contains a multi-scale fusion module or a multi-scale parallel module), and finally ends with a multi-scale attention module. The motivation to build our cell with such a layout is based on the consideration that it can bring in a series of high-scale to low-scale attentive sub-networks, as the color-filled boxes in the upper part of Fig.~\ref{fig:framework} show, this could effectively strengthen the representation
ability of the de-raining network. 


\subsubsection{Multi-scale Attentive Image De-raining Network}
With the multi-scale attentive cells, we further use them to construct a powerful multi-scale attentive image de-raining network, as the bottom part of Fig.~\ref{fig:framework} shows. Following the idea in~\cite{Gou2020CLEARER}, we first exploit two residual blocks~\cite{He2016Deep} cascaded together to receive the rainy image $\textmd{I}_{1\times}$ and keep its original scale unchanged. After that, we introduce several multi-scale attentive cells connected one-by-one to extract the desirable multi-scale attentive features. The last cell is followed by a multi-scale fusion module which is responsible for fusing the multi-scale attentive features into the single-scale ones. And at the end of our de-raining network, a $1\times 1$ convolutional layer is employed to output the final de-rained image $\textmd{O}_{1\times}$. Note that, as we increase the number of cells, more and more high-scale to low-scale attentive sub-networks will be added into the de-raining network, this can significantly facilitate the learning of highly representative and discriminative features for image de-raining. 

\subsection{Multi-scale Attentive Neural Architecture Search}
\label{ssec:search}

With the comprehensive coverage of the multi-scale attention search space, we then give a method for the multi-scale attentive neural architecture search. Our ultimate objective in searching is to find the best path in each multi-scale attentive cell, so as to determine the internal architecture of the cell.
For convenience, we still take the first cell in the de-raining network as an example, which is shown in the upper part of Fig.~\ref{fig:framework}. As shown, we need to search two types of path to determine the internal architecture of the cell. Among them, the first type of path is searched to determine whether the multi-scale~\emph{parallel} module or the multi-scale~\emph{fusion} module is chosen at each column, while the second type of path is searched to determine what kind of attention operations should be applied to our multi-scale attention module.

Specifically, let $\bm{\textmd{X}}^{m}$ be the input of the multi-scale parallel module or the multi-scale fusion module at the $m$-th column, where $m\in\{1,2,3,4\}$, 
$\rm P^{m}(\cdot)$ and $\rm F^{m}(\cdot)$ be the multi-scale parallel module and the multi-scale fusion module at the $m$-th column.
Then, we have the first type of path search: 
\begin{equation}
	\begin{split}
		\label{equ:esim9}
		\bm{\textmd{X}}^{m+1}=\rm P^{m}(\bm{\textmd{X}}^{m}|\alpha^{m}_{0})+\rm F^{m}(\bm{\textmd{X}}^{m}|\alpha^{m}_{1}),
	\end{split}
\end{equation}
where $\alpha^{m}_{0}$, $\alpha^{m}_{1}\in\{0,1\}$ denote the architecture parameters in the $\rm P^{m}(\cdot)$ and $\rm F^{m}(\cdot)$, respectively. If $\alpha^{m}_{0} =1$, $\alpha^{m}_{1}=0$, then the path to the multi-scale parallel module $\rm P^{m}(\cdot)$ is chosen. Otherwise, $\alpha^{m}_{0} =0$, $\alpha^{m}_{1}=1$ means that the path to the multi-scale fusion module $\rm F^{m}(\cdot)$ is chosen.

Regarding the multi-scale attention module, let $\{\textmd{X}_{1}, \textmd{X}_{2}\}$ be the split group features, then we have the second type of path search:
\begin{equation}
	\begin{split}
		\label{equ:esim10}
		& \bm{\textmd{Y}}_{1}=\Phi_{k}(\bm{\textmd{X}}_{1}|\beta_{k}) \\
		&\bm{\textmd{Y}}_{2}=\Phi_{k}(\bm{\textmd{X}}_{2}|\beta_{k})+\Phi_{k}(\bm{\textmd{Y}}_{1}|\beta_{k})
	\end{split}
\end{equation}
where $\beta_{k}$ represents the architecture parameters in the multi-scale attention module. Likewise, if their values are equal to one, then their corresponding attention operations,~\emph{i.e.}, $\Phi_{k}(\cdot)$, are chosen, as the red arrows in the upper right part of Fig.~\ref{fig:framework} show. $\{\textmd{Y}_{1},\textmd{Y}_{2}\}$ denotes the intermediate features resulted from the attention operations. $k\in\{0,1,\ldots,6\}$ is the index of the specific attention operations listed in Table~\ref{tab:attention}.

Obviously, all of the architecture parameters are limited to a binary constraint,~\emph{i.e.}, $\alpha_{0}$, $\alpha_{1},\beta_{k}\in\{0,1\}$, this means that they are discrete. In order to integrate our search method into a differentiable manner, like~\cite{Liu2018Darts,Gou2020CLEARER,Ma2020Auto}, we 
conduct continuous relaxation on the architecture parameters using softmax, such that: 
\begin{equation}
		\begin{split}
			\label{equ:esim11-1}
			&\alpha_{i}=\frac{\text{exp}(\mu_{i})}{\sum^{1}_{i=0}\text{exp}(\mu_{i})}, \\ 
			&\beta_{k}=\frac{\text{exp}(\nu_{k})}{\sum^{6}_{k=0}\text{exp}(\nu_{k})}, \\
		\end{split}
\end{equation}
where $\mu_{i}$ and $\nu_{k}$ denote the learnable architecture weights corresponding to $\alpha_{i}$ and $\beta_{k}$, respectively. 

For brevity, we use $\theta$ to represent all the architecture weights $\mu_{i}$ and $\nu_{k}$ in the de-raining network. Accordingly, our search problem turns to optimizing the architecture weights $\theta$ and the network weights $\omega$, as follows:
\begin{equation}
	\begin{split}
			\label{equ:esim12}
			& \;\;\;\;\;\;\;\;\;\; \min_{\theta}\mathcal{L}_{tainB}(\omega^{\ast}(\theta),\theta) \\
			&s.t. \; \omega^{\ast}(\theta) = \mathop{\arg\min}_{\omega}\mathcal{L}_{trainA}(\omega,\theta),
	\end{split}
\end{equation}
where $\mathcal{L}_{trainA}$ denotes the loss function for optimizing the network weights $\omega$, and $\mathcal{L}_{trainB}$ denotes the loss function for optimizing the architecture weights $\theta$. To solve this optimization problem~\ref{equ:esim12}, we adopt bi-level optimization~\cite{Ma2020Auto}, as shown in Algorithm 1.

Finally, when finishing the architecture search training, all of architecture parameters $\{\alpha_{i}, \beta_{k}\}$ are encoded into binary, such that: 
\begin{equation}
	\label{equ:esim120}
	\begin{split}
		\alpha_{i}=\left\{
		\begin{aligned}
			1 & , \;\;\; i=\mathop{\arg\max}_{i}\mu_{i}, \\
			0 & , \;\;\; \text{otherwise}, 
		\end{aligned}
		\right. 
	\end{split}
\end{equation}

\begin{equation}
	\label{equ:esim121}
	\begin{split}
		\beta_{k}=\left\{
		\begin{aligned}
			1 & , \;\;\; k=\mathop{\arg\max}_{k}\nu_{k}, \\
			0 & , \;\;\; \text{otherwise}, 
		\end{aligned}
		\right. 
	\end{split}
\end{equation}
At this point, the best paths in the de-raining network are determined.






\subsection{Multi-to-one Training Strategy}
\label{ssec:training}
\begin{figure*}[t]
	\centering
	\includegraphics[width=0.8\linewidth]{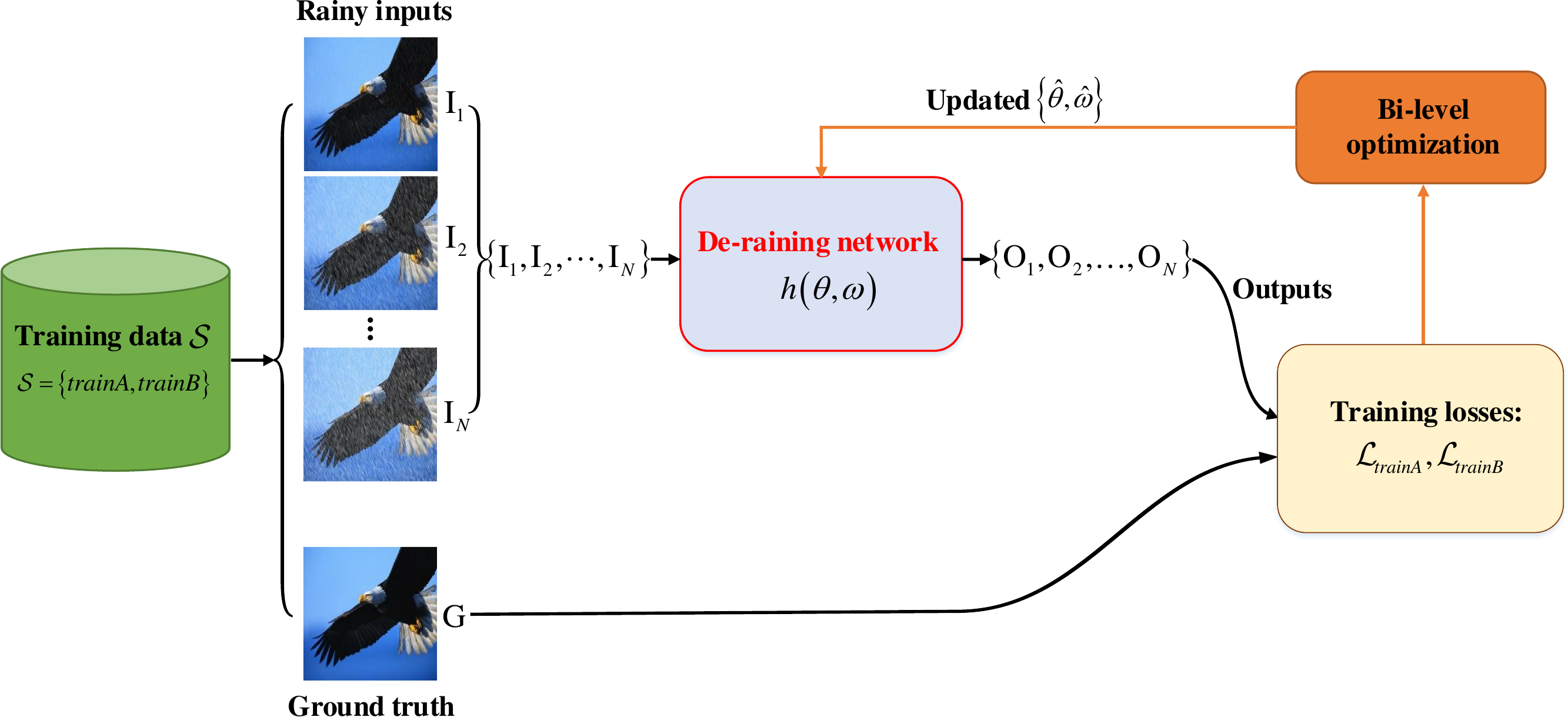}
	\caption{The schematic illustration of our multi-to-one training strategy. Multi-to-one image pairs $\{\textmd{I}_{1},\textmd{I}_{2},\ldots,\textmd{I}_{N}, G\}$ are used for training. The de-raining network takes multiple rainy images $\{\textmd{I}_{1},\textmd{I}_{2},\ldots,\textmd{I}_{N}\}$ with the same background $G$ as input, and estimates the outputs $\{\textmd{O}_{1},\textmd{O}_{2},\ldots,\textmd{O}_{N}\}$ to calculate the training losses $\mathcal{L}_{trainA}$ and $\mathcal{L}_{trainB}$. The resulting $\mathcal{L}_{trainA}$ and $\mathcal{L}_{trainB}$ are then feed back to the de-raining network to update the architecture weight $\theta$ and the network weight $\omega$ through the bi-level optimization.
	}
	\label{fig:training}
\end{figure*}
The schematic illustration of the proposed~\emph{multi-to-one training strategy} is depicted in Fig.~\ref{fig:training}. 
As shown, the training data set $\mathcal{S}$ is first partitioned into two disjoint training sets ${trainA}$ and ${trainB}$, both of them consist of multi-to-one image pairs and each pair have multiple rainy images $\{\textmd{I}_{1},\textmd{I}_{2},\ldots,\textmd{I}_{N}\}$ with the same clean background (\emph{i.e.}, the ground truth image $G$).
Then, the de-raining network is driven to process the multi-to-one image pairs to calculate the loss $\mathcal{L}_{trainA}$ on the set ${trainA}$ and the loss $\mathcal{L}_{trainB}$ on the set ${trainB}$. Finally, the resulting losses $\mathcal{L}_{trainA}$ and $\mathcal{L}_{trainB}$ are fed back to the de-raining network to optimize the architecture weights $\theta$ and the network weights $\omega$ alternately through bi-level optimization. In the following, we will elaborate the formulation of the losses $\mathcal{L}_{trainA}$ and $\mathcal{L}_{trainB}$, respectively.


\begin{table}[t]
	\renewcommand{\arraystretch}{1.2}
	\tabcolsep0.0001cm
	\centering
	\label{tab:bi-level}       
	\begin{tabular}{p{0.95\columnwidth}}
		\Xhline{1.3pt}
		\textbf{Algorithm 1:} Bi-level optimization \\
		\hline
		\textbf{Require:} the training sets ${trainA}$ and ${trainB}$, the de-raining network $h(\theta,\omega)$ with architecture weight $\theta$ and network weight $\omega$, the number of training iterations $J$. \\
		1: Initialize $\theta$ and $\omega$         \\
		2: \textbf{for} $j=1$ \textbf{to} $J$ \textbf{do}  \\
		3: \;\;\;\; Sample a group of multi-to-one image pairs from ${trainA}$  \\
		4: \;\;\;\; Update $\hat{\omega}\leftarrow \omega - \frac{\partial\mathcal{L}_{trainA}}{\partial\omega}$                                \\
		5: \;\;\;\; Sample a group of multi-to-one image pairs from ${trainB}$ \\
		6: \;\;\;\; Update $\hat{\theta}\leftarrow \theta - \frac{\partial\mathcal{L}_{trainB}}{\partial\theta}$   \\
		7: \textbf{end for} \\
		\textbf{Return:} the optimized weights $\{\theta^{\ast},\omega^{\ast}\}$  \\
		\Xhline{1.3pt}
	\end{tabular}
\end{table}



\subsubsection{$\mathcal{L}_{trainA}$ for Optimizing $\omega$}
Since we train our de-raining network in a supervised manner, 
we can measure the distance between the de-rained outputs $\{\bm{\textmd{O}}_{1},\bm{\textmd{O}}_{2},\ldots,\bm{\textmd{O}}_{N}\}$ and the ground truth $G$ to form an~\emph{external loss} $\mathcal{L}_{\text{ext}}$. Specifically, $\mathcal{L}_{\text{ext}}$ is defined as a composition of the MSE loss and SSIM loss~\cite{Wang2004Image}, as follows:
\begin{equation}
	\begin{split}
		\label{equ:esim13}
		\mathcal{L}_{\text{ext}}\!=\!\frac{1}{N}\sum_{i\in \mathcal{N}}\! \lVert\bm{\textmd{O}}_{i}-\bm{\textmd{G}}\rVert^{2}_{2} \!+\!\frac{1}{N} \sum_{i\in \mathcal{N}}\!\big(1\!-\!\text{SSIM}(\bm{\textmd{O}}_{i},\bm{\textmd{G}})\big),
	\end{split}
\end{equation}
where $\mathcal{N}=\{1,2,\ldots, N\}$, and $\bm{\textmd{O}}_{i}\in\{\bm{\textmd{O}}_{1},\bm{\textmd{O}}_{2},\ldots,\bm{\textmd{O}}_{N}\}$.




On the other hand, for multi-to-one image pairs, the background in each pair is invariant. Accordingly, we can employ this prior knowledge to formulate an~\emph{internal loss} among the outputs $\{\bm{\textmd{O}}_{1},\bm{\textmd{O}}_{2},\ldots,\bm{\textmd{O}}_{N}\}$ to make them close as much as possible. In this work, we simply use MSE to measure the distance of all pairs $\{\bm{\textmd{O}}_{i},\bm{\textmd{O}}_{j}\}_{i\neq j}$ in $\{\bm{\textmd{O}}_{1},\bm{\textmd{O}}_{2},\ldots,\bm{\textmd{O}}_{N}\}$, as follows:
\begin{equation}
	\begin{split}
		\label{equ:esim14}
		\mathcal{L}_{\text{int}}=\frac{1}{C^{2}_{N}}\sum_{i\in \mathcal{N}} \sum_{j\in \mathcal{N}, i<j}\lVert\bm{\textmd{O}}_{i}-\bm{\textmd{O}}_{j}\rVert^{2}_{2},
	\end{split}
\end{equation}
where $C^{2}_{N}$ is the number of combinations,~\emph{i.e.}, the number of selecting two different elements from the set $\{\bm{\textmd{O}}_{1},\bm{\textmd{O}}_{2},\ldots,\bm{\textmd{O}}_{N}\}$. 

Therefore, the overall formulation of $\mathcal{L}_{\text{trainA}}$ for optimizing the network weights $\omega$ is as follows:
\begin{equation}
	\begin{split}
		\label{equ:esim116}
		\mathcal{L}_{trainA}=\mathcal{L}_{\text{ext}}+\mathcal{L}_{\text{int}}.
	\end{split}
\end{equation}

\subsubsection{$\mathcal{L}_{trainB}$ for Optimizing $\theta$}
The performance of image de-raining is sensitive to neural architectures, according to the previous study on image de-raining. Thus, when formulating the loss $\mathcal{L}_{trainB}$ for optimizing the architecture weight $\theta$, we also incorporate the~\emph{external loss} and the~\emph{internal loss} into $\mathcal{L}_{trainB}$ to facilitate the searching of the high-performance multi-scale attentive image de-raining networks. Meanwhile, besides the de-raining performance, 
a desirable de-raining network should also take into account the model complexity, since it is crucial to many resource-constrained scenarios such as mobile phones. 
To this end, we formulate a~\emph{model complexity loss} $\mathcal{L}_{\text{comp}}$ to control the complexity of our de-raining model, as follows:
\begin{equation}
	\begin{split}
		\label{equ:esim117}
		\!\mathcal{L}_{\text{comp}}\!=\!\frac{1}{U\!+\!V}\!\sum\!(\alpha_{0}\Omega_{0}\!+\!\alpha_{1}\Omega_{1}\!+\!\beta_{0}\Lambda_{0}\!+\!\ldots\!+\!\beta_{6}\Lambda_{6}),
	\end{split}
\end{equation} 
where $\Omega_{0}$ and $\Omega_{1}$ are the size of the multi-scale parallel module and the multi-scale fusion module, respectively, whereas $\Lambda_{0},\Lambda_{1},\ldots,\Lambda_{6}$ are the size of the attention operations listed in Table~\ref{tab:attention}. $U$ is the number of $\alpha$, and $V$ is the number of $\beta$.

On the other hand, although we have made our search method differentiable, there still exists a problem caused by the continuous relaxation using softmax. To be more specific, it could lead to a~\emph{trivial} solution about the architecture parameters, such that $\alpha_{0}=\alpha_{1}$, and $\beta_{0}=\beta_{1}=\ldots=\beta_{6}$. This means that these searched candidate modules as well as the searched candidate attention operations are indistinguishable. To tackle this problem, like~\cite{Gou2020CLEARER}, we introduce the~\emph{architecture regularization loss} $\mathcal{L}_{\text{arch}}$ to regularize the architecture parameters, as follows:
\begin{equation}
	\begin{split}
		\label{equ:esim118}
		&\mathcal{L}_{\text{arch}}\!=\!-\frac{1}{U}\sum_{\alpha\in\{\alpha_{0},\alpha_{1}\}}\big(\alpha\log\alpha\!+\!(1-\alpha)\log(1\!-\!\alpha)\big) \\
		&\;\;\;\;-\frac{1}{V}\sum_{\beta\in\{\beta_{0},\beta_{1},\ldots,\beta_{6}\}}\big(\beta\log\beta\!+\!(1\!-\!\beta)\log(1\!-\!\beta)\big).
	\end{split}
\end{equation}
Accordingly, based on Eq.~\ref{equ:esim11-1} and Eq.\ref{equ:esim118}, we relax the discrete architecture parameters into a continuous distribution which approximates to either $0$ or $1$.

By jointly considering the~\emph{external loss},~\emph{internal loss},~\emph{architecture regularization loss}, and~\emph{model complexity loss}, the overall formulation of $\mathcal{L}_{trainB}$ for optimizing the architecture weight $\theta$ is expressed as:
\begin{equation}
	\begin{split}
		\label{equ:esim119}
		\mathcal{L}_{trainB}=\mathcal{L}_{\text{ext}}+\mathcal{L}_{\text{int}}+\lambda_{\text{arch}}\mathcal{L}_{\text{arch}}+\lambda_{\text{comp}}\mathcal{L}_{\text{comp}},
	\end{split}
\end{equation}
where $\lambda_{\text{arch}}$ and $\lambda_{\text{comp}}$ denote the hyper-parameters to balance the corresponding losses $\mathcal{L}_{\text{arch}}$ and $\mathcal{L}_{\text{comp}}$.

\section{Experiments}
\label{sec:exp}
\subsection{Experimental Protocol}
\label{ssec:pro}
\subsubsection{Datasets}
\label{sssec:dataset}
In this research, we adopt two different synthetic datasets to validate the superiority of our proposed MANAS method for image de-raining. 

\begin{figure}[!t]
	\centering
	\subfigure[\scriptsize Light Rain]{
		\begin{minipage}[t]{0.233\linewidth}
			\centering
			\centerline{\includegraphics[width=0.83in]{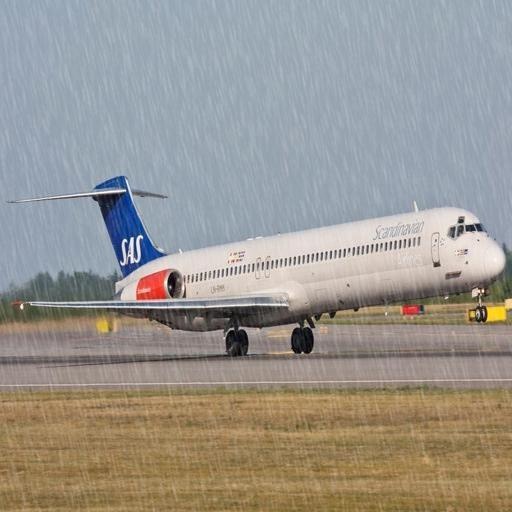}}
		\end{minipage}%
	}%
	\subfigure[\scriptsize Medium Rain]{
		\begin{minipage}[t]{0.233\linewidth}
			\centering
			\centerline{\includegraphics[width=0.83in]{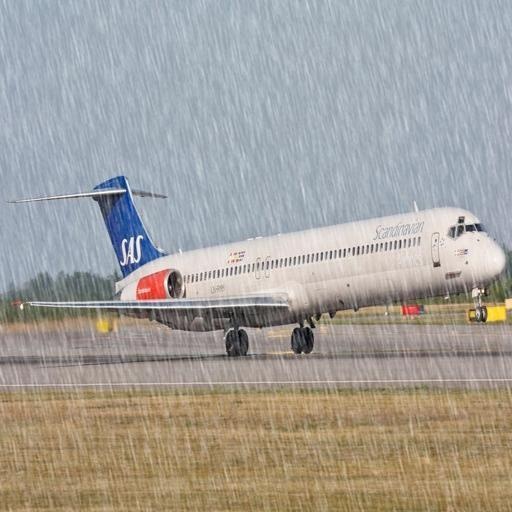}}
		\end{minipage}%
	}%
	\subfigure[\scriptsize Heavy Rain]{
		\begin{minipage}[t]{0.233\linewidth}
			\centering
			\centerline{\includegraphics[width=0.83in]{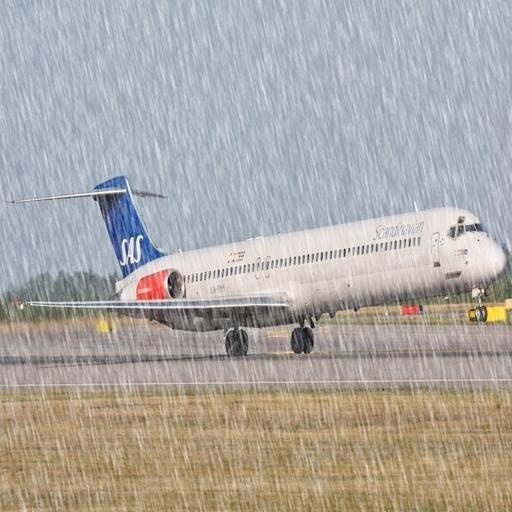}}
		\end{minipage}%
	}%
	\subfigure[\scriptsize Ground Truth]{
		\begin{minipage}[t]{0.233\linewidth}
			\centering
			\centerline{\includegraphics[width=0.83in]{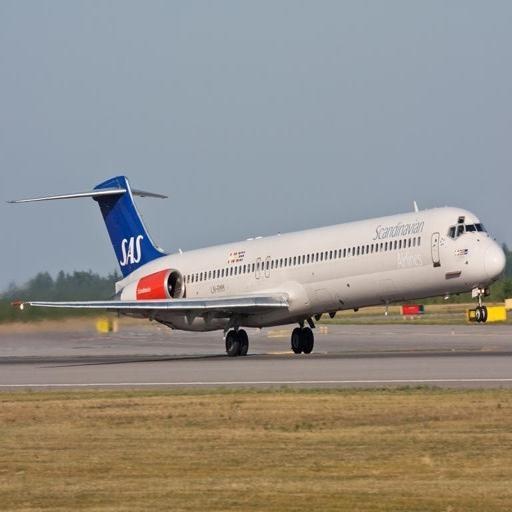}}
		\end{minipage}%
	}%
	\centering
	\caption{An example of three-to-one rainy/clean image pair from DID-MDN~\cite{Zhang2018Density} dataset.}
	\label{fig:DID_MDN}
\end{figure}

\begin{figure}[!t]
	\centering
	\subfigure[\scriptsize Light Fog]{
		\begin{minipage}[t]{0.48\linewidth}
			\centering
			\centerline{\includegraphics[width=1.685in]{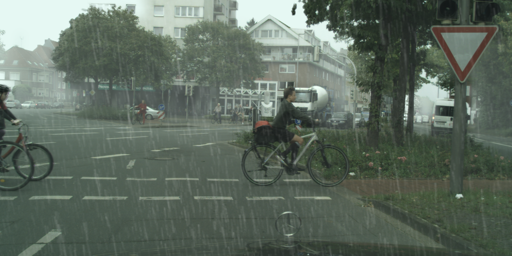}}
		\end{minipage}%
	}%
	\subfigure[\scriptsize Medium Fog]{
		\begin{minipage}[t]{0.48\linewidth}
			\centering
			\centerline{\includegraphics[width=1.685in]{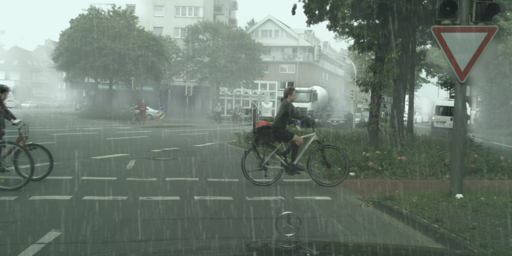}}	
		\end{minipage}%
	}%
	\quad
	\subfigure[\scriptsize Heavy Fog]{
		\begin{minipage}[t]{0.48\linewidth}
			\centering
			\centerline{\includegraphics[width=1.685in]{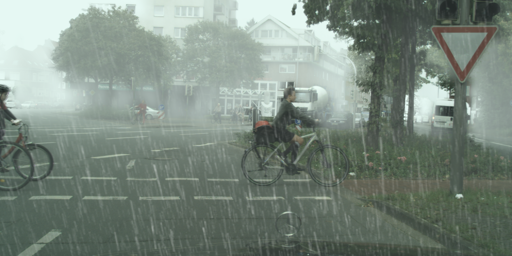}}
		\end{minipage}%
	}%
	\subfigure[\scriptsize Ground Truth]{
		\begin{minipage}[t]{0.48\linewidth}
			\centering
			\centerline{\includegraphics[width=1.685in]{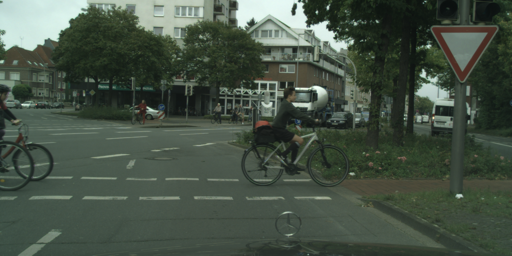}}
		\end{minipage}%
	}%
	\centering
	\caption{An example of three-to-one rainy/clean image pair from RainCityscapes~\cite{Hu2021Single} dataset.}
	\label{fig:Rain_fog}
\end{figure}

The first dataset, called~\emph{DID-MDN}~\cite{Zhang2018Density}, was created according to the~\emph{additive composite model}. The rainy image in this dataset is a simple
superimposition of the background layer and the rain streak
layer, without considering the rain accumulation that could
produce veiling/fog effect. The DID-MDN dataset consists of $4400$ three-to-one rainy/clean image pairs,~\emph{i.e.}, every three rainy images correspond to a common clean background image, and each rainy image in the pair represents a rain-density level such as light rain, medium rain, and heavy rain, as shown in Fig.~\ref{fig:DID_MDN}. We utilized $4000$ three-to-one rainy/clean image pairs (${trainA}=2000$ and ${trainB}=2000$) to train and find the best multi-scale attentive neural architecture for the de-raining network, and then test it with $400$ three-to-one rainy/clean image pairs to report the final de-raining performance. Note that our MANAS was still trained and tested according to the original training and testing split in the DID-MDN dataset,~\emph{i.e.}, $12000$ rainy images and $4000$ clean background images (equal to $4000$ three-to-one rainy/clean image pairs) for training, and $1200$ rainy images and $400$ clean background images (equal to $400$ three-to-one rainy/clean image pairs) for testing.

The second dataset, called~\emph{RainCityscapes}~\cite{Hu2021Single}, was created according to the~\emph{depth-aware rain model}, in which the depth information is used to distribute different degrees of rain streaks and fog in the images. The RainCityscapes dataset consists of $10620$ rainy images with $9432$ for training and $1188$ for testing. 
In this work, we partitioned $9432$ training images and $1188$ testing images into three fog-thickness levels~\footnote{Note that the fog in the images can be viewed as the rain, due to the fact that the fog effect in the real world is caused by the rain streak accumulation~\cite{Yang2019Joint},.}, according to the fog-thickness attenuation coefficient provided by~\cite{Hu2021Single},~\emph{i.e.}, light fog, medium fog, and heavy fog, as shown in Fig.~\ref{fig:Rain_fog}. Likewise, we still followed the original training and testing split in the RainCityscapes dataset. That is we employed $3144$ three-to-one rainy/clean image pairs (${trainA}=1572$ and ${trainB}=1572$), corresponding to $9432$ rainy images and $3144$ clean background images in the training set of RainCityscapes dataset, to train and find the best multi-scale attentive neural architecture for the de-raining network, and $396$ three-to-one rainy/clean image pairs, corresponding to $1188$ rainy images and $396$ clean background images in the testing set of RainCityscapes dataset, to report the corresponding results.


\subsubsection{Evaluation Metrics}
\label{sssec:metrics}
For the performance evaluation, we adopted three commonly-used metrics including Peak Signal to Noise Ration (PSNR), Structure Similarity Index (SSIM)~\cite{Wang2004Image}, and Natural Image Quality Evaluator (NIQE)~\cite{Mittal2012Making}, to evaluate the performance of rain removal on synthesized datasets. Generally speaking, higher PSNR and SSIM while lower NIQE values indicate better de-raining results. Due to the fact that the real-world rainy images have no corresponding clean ground truth images, we thus adopted a blind de-raining quality assessment model B-FEN~\cite{Wu2020Subjective} to estimate scores for the de-rained images. A higher score for the de-rained image generally means it has better visual quality.




\subsection{Implementation Details}
\label{ssec:implement}
Our MANAS method was trained by the specifically-designed~\emph{multi-to-one training strategy} in a stage-wise manner, which involves two training stages, namely the~\emph{architecture search stage} and the~\emph{model training stage}. In the architecture search stage, the proposed MANAS method was trained to optimize the architecture weights $\theta$ and the network weights $\omega$. For optimizing $\omega$, we used a standard SGD optimizer with the momentum of $0.9$ and the weight decay of $3\times10^{-4}$. And the learning rate is automatically decayed from $2\times10^{-3}$ to $1\times10^{-4}$ by the cosine anneal strategy~\cite{Loshchilov2016Sgdr}. For optimizing $\theta$, we adopted the Adam optimizer~\cite{Kingma2014Adam} with the basic learning rate of $3\times10^{-4}$ and the weight decay of $1\times10^{-3}$. To make a fair comparison with other competing methods, we ignored the model complexity by simply picking the hyper-parameter $\mathcal{\lambda}_{\text{comp}}$ at $0$. In our later experiments, we will show the effect of this hyper-parameter on balancing the de-raining performance and the model complexity. For $\mathcal{\lambda}_{\text{arch}}$, we fixed it at $0.01$. In the model training stage, we encoded the architecture parameters into binary and froze them, and then fine-tune the network weights $\omega$ with the overall training data set $\mathcal{S}$. We adopted the Adam optimizer to optimize the network weights $\omega$. And the learning rate is decayed from $1\times10^{-3}$ to $0$ via the cosine annealing strategy. We assigned three multi-scale attentive cells (\emph{i.e.}, $\text{T}=3$) in the de-raining network to have a comparison with other competing methods. In our later ablation experiments, we will study the influence of the cell number on de-raining performance.

\begin{figure}[!t]
	\centering
	\subfigure[\scriptsize DID-MDN dataset]{
		\begin{minipage}[t]{0.48\linewidth}
			\centering
			\centerline{\includegraphics[width=1.85in]{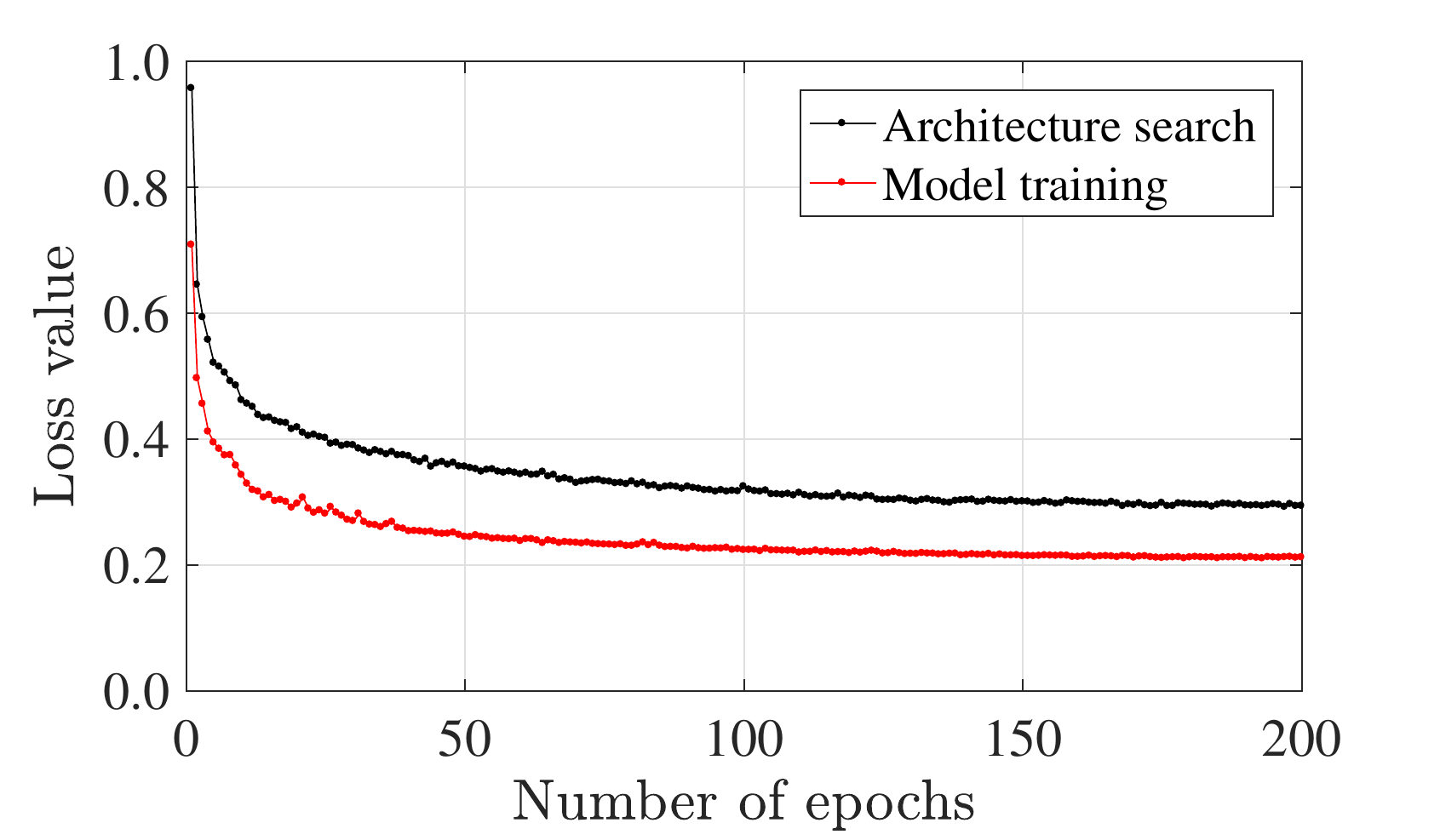}}
		\end{minipage}%
	}%
	\subfigure[\scriptsize RainCityscapes dataset]{
		\begin{minipage}[t]{0.48\linewidth}
			\centering
			\centerline{\includegraphics[width=1.85in]{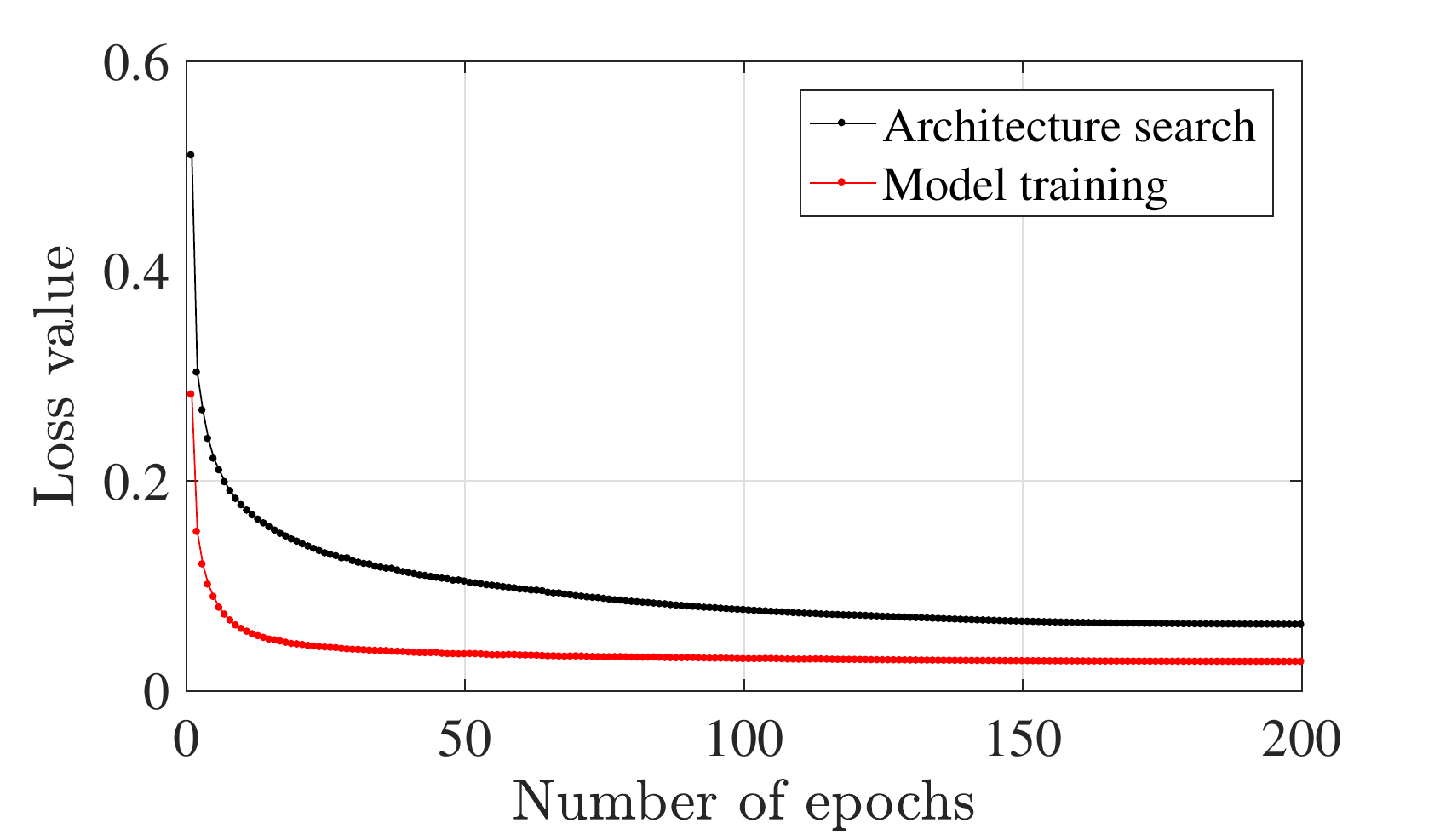}}	
		\end{minipage}%
	}%
	\caption{Training convergence curves of our MANAS on (a) DID-MDN~\cite{Zhang2018Density} and (b) RainCityscapes~\cite{Hu2021Single} datasets.}
	\label{fig:convergence}
\end{figure}

For DID-MDN~\cite{Zhang2018Density} dataset, we followed the same practice in~\cite{Gou2020CLEARER,Suganuma2018Exploiting}, randomly cropping the $512\times512$ original resolution images to $64\times64$ image patches to augment the data and improve the search efficiency. For RainCityscapes~\cite{Hu2021Single} dataset, since the scene depth plays an important role in distributing different degrees of rain streaks and fog in the image, randomly cropping the original resolution image will inevitably destroy the scene depth, and in turn breaks the spatial distribution of rain streaks and fog in the image. With this consideration, we followed the practice in~\cite{Hu2019Depth,Hu2021Single}, scaling the $1024\times2048$ original resolution images to the $128\times256$ versions to maintain the distribution pattern of the rain streaks and fog in the image.
In addition, during the training procedure, we employed the image flip to augment the training data. All experiments are conducted on a server with NVIDIA 
GTX 2080Ti and Intel(R) Xeon(R) CPU E5-2630 2.20GHz. In our single GPU implementation, the training~\footnote{The training includes the architecture search and the model training.} of the proposed MANAS on DID-MDN and RainCityscapes datasets respectively take $102$ and $91$ hours. The training convergence curves of MANAS on these two datasets are shown in Fig.~\ref{fig:convergence}. As can be seen, the architecture search and the model training both converge within $200$ epochs, and the model training can present a faster convergence rate and lower training loss.

\begin{table*}[t]
	\renewcommand{\arraystretch}{1.1}
	\tabcolsep0.2cm
	\centering
	\caption{Comparison of PSNR, SSIM, and NIQE results by different methods on the test set of DID-MDN~\cite{Zhang2018Density} and RainCityscapes~\cite{Hu2021Single}, respectively. The best results are marked in bold.}
	\label{tab:synthetic}       
	\begin{tabular}{ccccccccccc}
		\Xhline{1.3pt}
		\multicolumn{1}{c}{\multirow{3}{*}{Method}} & \multicolumn{1}{c}{Dataset} &  \multicolumn{3}{c}{DID-MDN~\cite{Zhang2018Density}}   & \multicolumn{3}{c}{RainCityscapes~\cite{Hu2021Single}}   \\
		\cmidrule{2-8}
		& \multicolumn{1}{|c|}{Year} & \multicolumn{1}{c}{PSNR $\uparrow$}    & \multicolumn{1}{c}{SSIM $\uparrow$} & \multicolumn{1}{c|}{NIQE $\downarrow$} & \multicolumn{1}{c}{PSNR $\uparrow$}    & \multicolumn{1}{c}{SSIM $\uparrow$} & \multicolumn{1}{c}{NIQE $\downarrow$} \\
		\hline
		\multirow{1}{*}{DSC~\cite{Luo2015Removing}} &  \multicolumn{1}{|c|}{2015}  & \multicolumn{1}{c}{21.44}         & \multicolumn{1}{c}{0.790}    & \multicolumn{1}{c|}{5.56}    & \multicolumn{1}{c}{16.25}         & \multicolumn{1}{c}{0.775}  & \multicolumn{1}{c}{5.94}  \\
		\hline
		\multirow{1}{*}{GMM~\cite{Li2016Rain}} &  \multicolumn{1}{|c|}{2016}  & \multicolumn{1}{c}{22.75}         & \multicolumn{1}{c}{0.835}  & \multicolumn{1}{c|}{5.37}  & \multicolumn{1}{c}{17.80}         & \multicolumn{1}{c}{0.817}  & \multicolumn{1}{c}{5.88}  \\
		\hline
		\multirow{1}{*}{JCAS~\cite{Gu2017Joint}} &  \multicolumn{1}{|c|}{2017}  & \multicolumn{1}{c}{23.62}         & \multicolumn{1}{c}{0.777}  & \multicolumn{1}{c|}{4.16}  & \multicolumn{1}{c}{15.66}         & \multicolumn{1}{c}{0.771}  & \multicolumn{1}{c}{6.01} \\
		\hline
		\multirow{1}{*}{DID-MDN~\cite{Zhang2018Density}} &  \multicolumn{1}{|c|}{2018}  & \multicolumn{1}{c}{27.95}         & \multicolumn{1}{c}{0.909}  & \multicolumn{1}{c|}{4.04}  & \multicolumn{1}{c}{28.43}         & \multicolumn{1}{c}{0.935}  & \multicolumn{1}{c}{4.40} \\
		\hline
		\multirow{1}{*}{PReNet~\cite{Ren2019Progressive}} &  \multicolumn{1}{|c|}{2019}  & \multicolumn{1}{c}{30.93}         & \multicolumn{1}{c}{0.905}    & \multicolumn{1}{c|}{3.71}    & \multicolumn{1}{c}{30.60}         & \multicolumn{1}{c}{0.967}  & \multicolumn{1}{c}{3.88}   \\
		\hline
		\multirow{1}{*}{MSPFN~\cite{Jiang2020Multi}} &  \multicolumn{1}{|c|}{2020}  & \multicolumn{1}{c}{28.46}         & \multicolumn{1}{c}{0.845}     & \multicolumn{1}{c|}{3.96}    & \multicolumn{1}{c}{25.26}         & \multicolumn{1}{c}{0.934}      & \multicolumn{1}{c}{4.31}  \\
		\hline
		\multirow{1}{*}{CLEARER~\cite{Gou2020CLEARER}} &  \multicolumn{1}{|c|}{2020}  & \multicolumn{1}{c}{31.43}         & \multicolumn{1}{c}{0.893}     & \multicolumn{1}{c|}{3.62}    & \multicolumn{1}{c}{32.01}         & \multicolumn{1}{c}{0.976}      & \multicolumn{1}{c}{3.68}  \\
		\hline
		\multirow{1}{*}{DGNL-Net-fast~\cite{Hu2021Single}} &  \multicolumn{1}{|c|}{2021}   & \multicolumn{1}{c}{31.21}     & \multicolumn{1}{c}{0.877}         & \multicolumn{1}{c|}{3.64}     & \multicolumn{1}{c}{30.59}     & \multicolumn{1}{c}{0.956}    & \multicolumn{1}{c}{4.15}   \\
		\hline
		\multirow{1}{*}{DGNL-Net~\cite{Hu2021Single}} &  \multicolumn{1}{|c|}{2021}   & \multicolumn{1}{c}{32.41}     & \multicolumn{1}{c}{0.910}         & \multicolumn{1}{c|}{\textbf{3.52}}     & \multicolumn{1}{c}{32.44}     & \multicolumn{1}{c}{0.973}    & \multicolumn{1}{c}{4.01}  \\
		\hline
		\multirow{1}{*}{$\textbf{MANAS (Ours)}$} &  \multicolumn{1}{|c|}{-} & \multicolumn{1}{c}{\textbf{32.60}}  & \multicolumn{1}{c}{\textbf{0.922}}  & \multicolumn{1}{c|}{3.59}  & \multicolumn{1}{c}{\textbf{35.19}} & \multicolumn{1}{c}{\textbf{0.984}}  & \multicolumn{1}{c}{\textbf{3.38}}  \\
		\Xhline{1.3pt}
	\end{tabular}
\end{table*}
\begin{figure*}[!t]
	\centering
	\subfigure{
		\begin{minipage}[t]{0.135\linewidth}
			\centering
			\centerline{\includegraphics[width=0.98in]{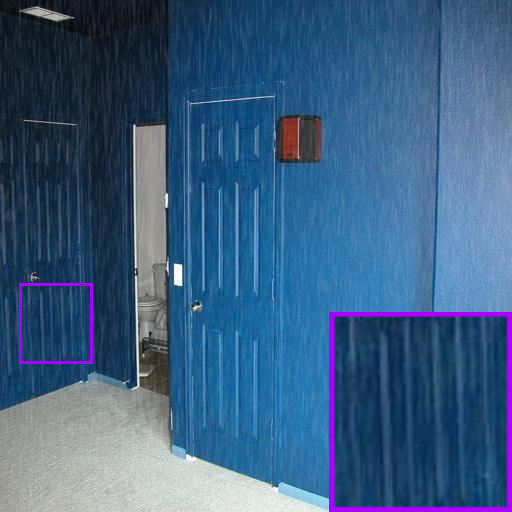}}
			{\tiny{Light Rain: 30.06/0.828}}
			\vspace{4pt}
			\vspace{1.7pt}
			\centerline{\includegraphics[width=0.98in]{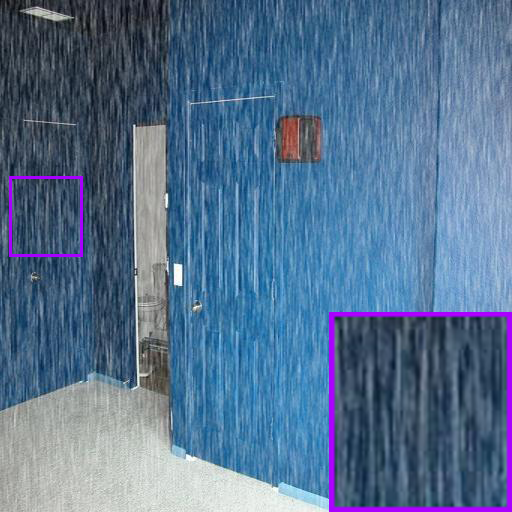}}
			{\tiny{Medium Rain: 18.54/0.481}}
			\vspace{4pt}
			\vspace{2.6pt}
			\centerline{\includegraphics[width=0.98in]{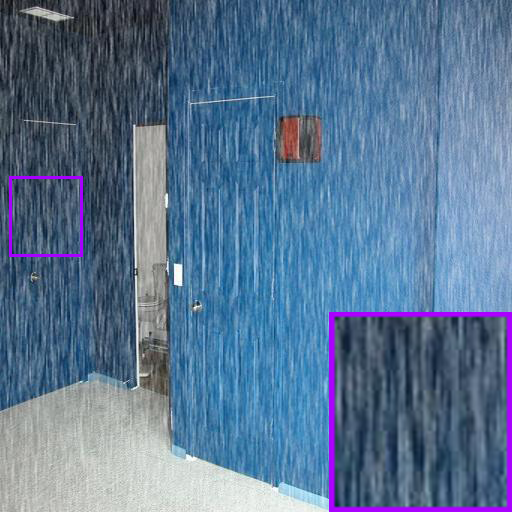}}
			{\tiny{Heavy Rain: 17.67/0.461}}
		\end{minipage}%
	}%
	\subfigure{
		\begin{minipage}[t]{0.135\linewidth}
			\centering
			\centerline{\includegraphics[width=0.98in]{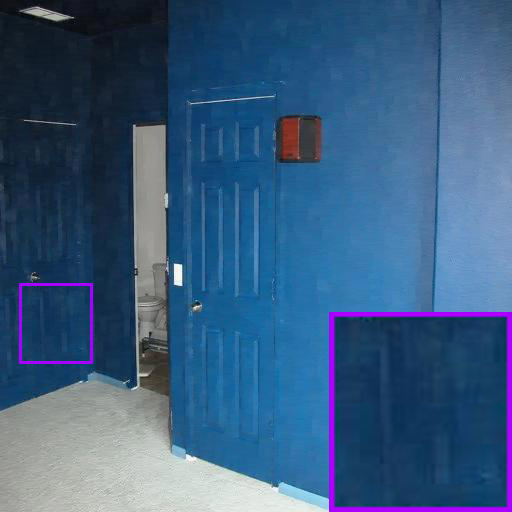}}
			{\tiny{JCAS~\cite{Gu2017Joint}: 30.47/0.844}}
			\vspace{4pt}
			\vspace{2pt}
			\centerline{\includegraphics[width=0.98in]{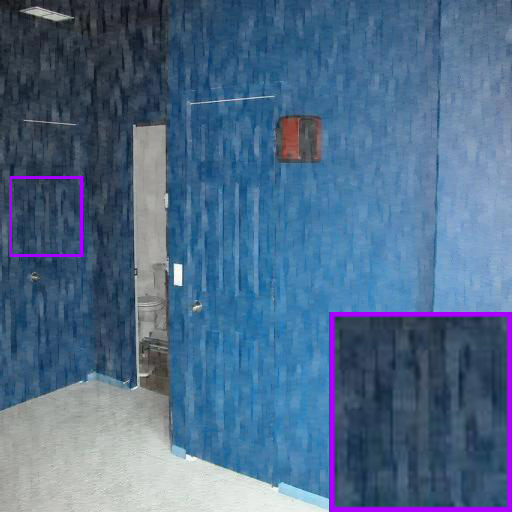}}
			{\tiny{JCAS~\cite{Gu2017Joint}: 20.66/0.654}}
			\vspace{4pt}
			\vspace{2pt}
			\centerline{\includegraphics[width=0.98in]{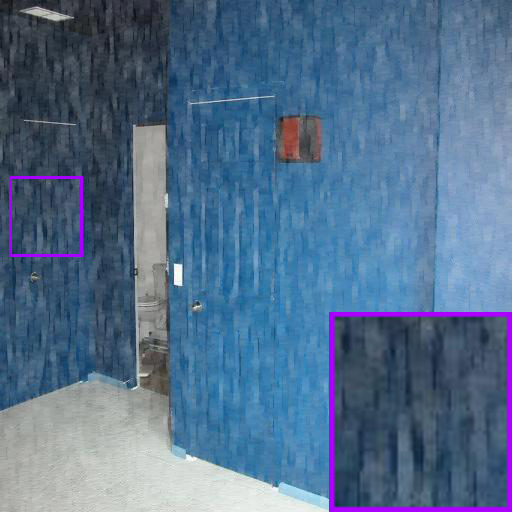}}
			{\tiny{JCAS~\cite{Gu2017Joint}: 19.63/0.632}}
		\end{minipage}%
	}%
	\subfigure{
		\begin{minipage}[t]{0.135\linewidth}
			\centering
			\centerline{\includegraphics[width=0.98in]{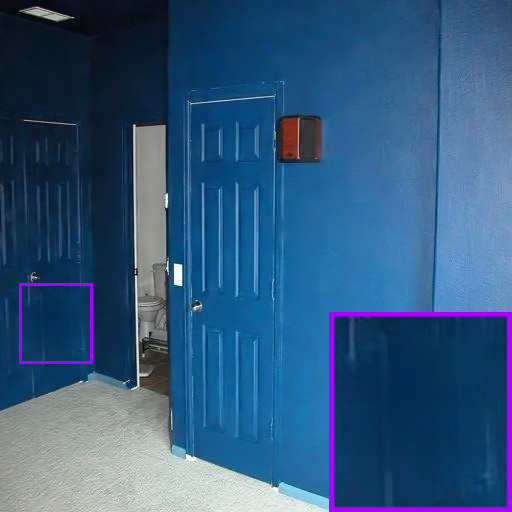}}
			{\tiny{PReNet~\cite{Ren2019Progressive}: 35.10/0.919}}
			\vspace{4pt}
			\vspace{2pt}
			\centerline{\includegraphics[width=0.98in]{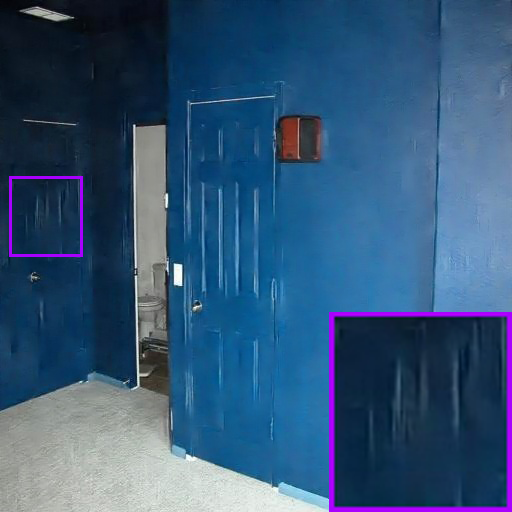}}
			{\tiny{PReNet~\cite{Ren2019Progressive}: 29.75/0.836}}
			\vspace{4pt}
			\vspace{2pt}
			\centerline{\includegraphics[width=0.98in]{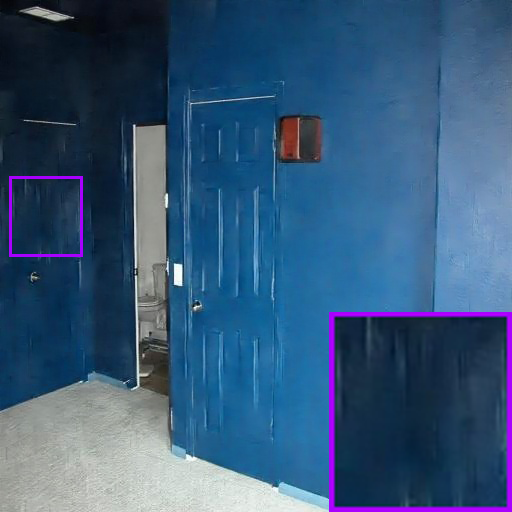}}
			{\tiny{PReNet~\cite{Ren2019Progressive}: 29.79/0.836}}
		\end{minipage}%
	}%
	\subfigure{
		\begin{minipage}[t]{0.135\linewidth}
			\centering
			\centerline{\includegraphics[width=0.98in]{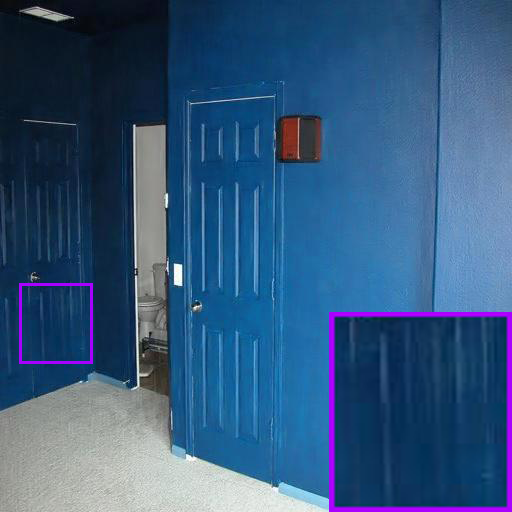}}
			{\tiny{CLEARER~\cite{Gou2020CLEARER}: 35.38/0.912}}
			\vspace{4pt}
			\vspace{2pt}
			\centerline{\includegraphics[width=0.98in]{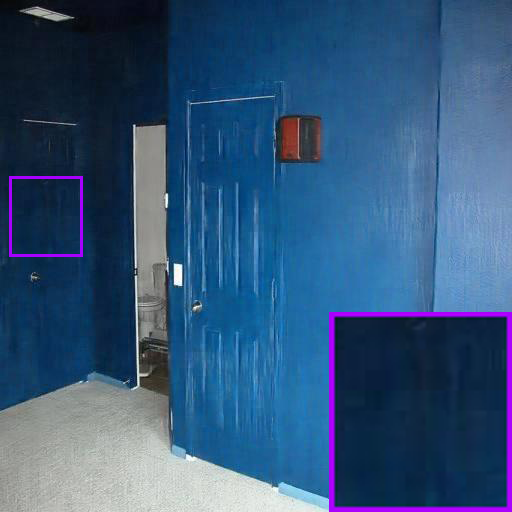}}
			{\tiny{CLEARER~\cite{Gou2020CLEARER}: 30.53/0.819}}
			\vspace{4pt}
			\vspace{2pt}
			\centerline{\includegraphics[width=0.98in]{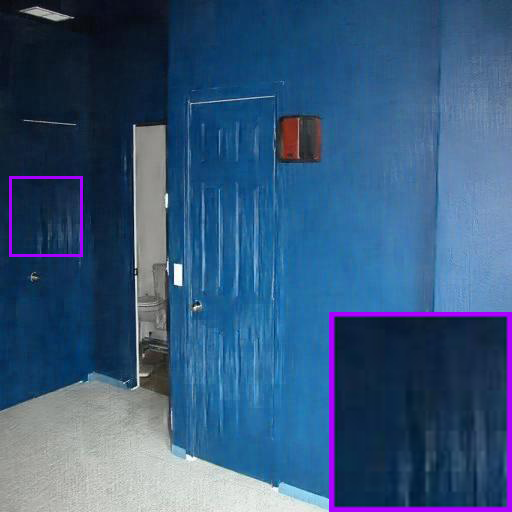}}
			{\tiny{CLEARER~\cite{Gou2020CLEARER}: 29.95/0.813}}
		\end{minipage}%
	}%
	\subfigure{
		\begin{minipage}[t]{0.135\linewidth}
			\centering
			\centerline{\includegraphics[width=0.98in]{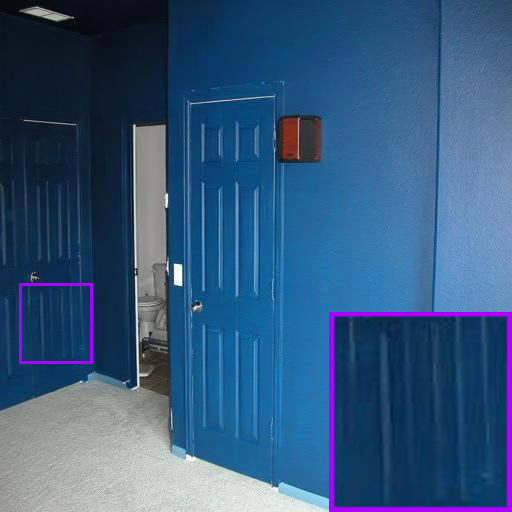}}
			{\tiny{DGNL-Net~\cite{Hu2021Single}: \textbf{36.45}/0.926}}
			\vspace{4pt}
			\vspace{2pt}
			\centerline{\includegraphics[width=0.98in]{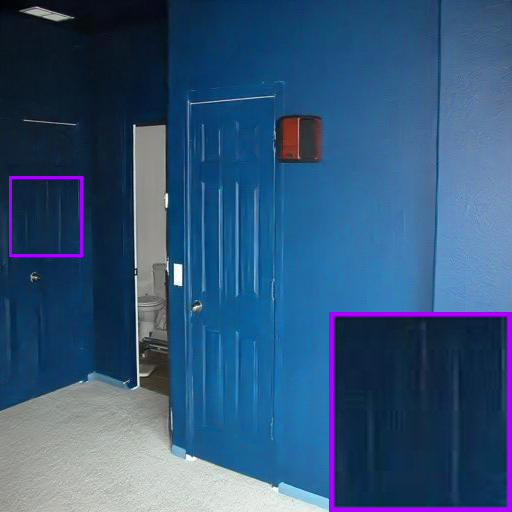}}
			{\tiny{DGNL-Net~\cite{Hu2021Single}: 31.79/0.851}}
			\vspace{4pt}
			\vspace{2pt}
			\centerline{\includegraphics[width=0.98in]{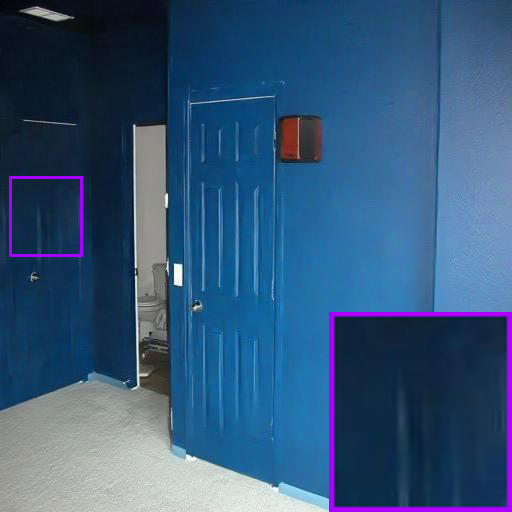}}
			{\tiny{DGNL-Net~\cite{Hu2021Single}: 32.10/0.853}}
		\end{minipage}%
	}%
	\subfigure{
		\begin{minipage}[t]{0.135\linewidth}
			\centering
			\centerline{\includegraphics[width=0.98in]{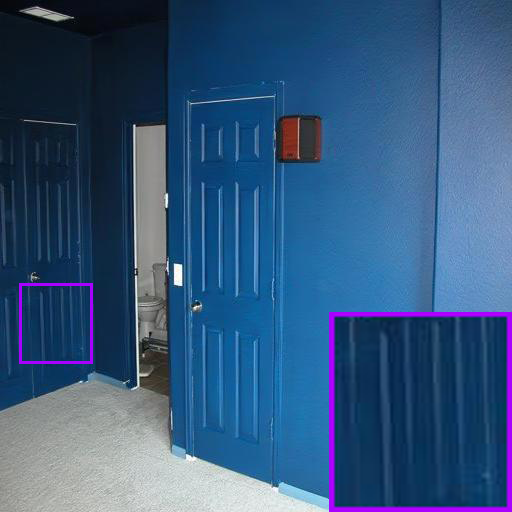}}
			{\tiny{\textbf{MANAS (Ours):} 35.95/\textbf{0.934}}}
			\vspace{4pt}
			\vspace{2pt}
			\centerline{\includegraphics[width=0.98in]{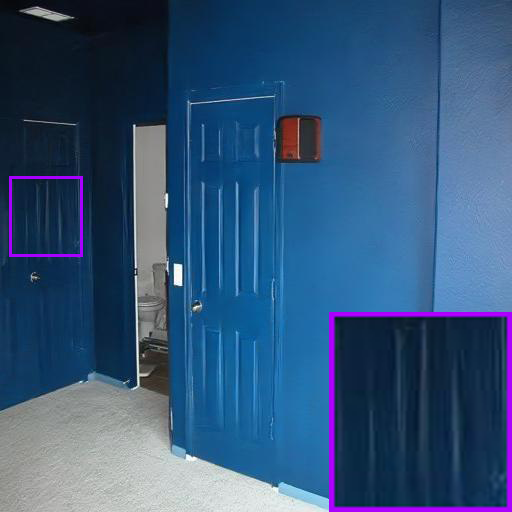}}
			{\tiny{\textbf{MANAS (Ours): 31.86/0.868}}}
			\vspace{4pt}
			\vspace{2pt}
			\centerline{\includegraphics[width=0.98in]{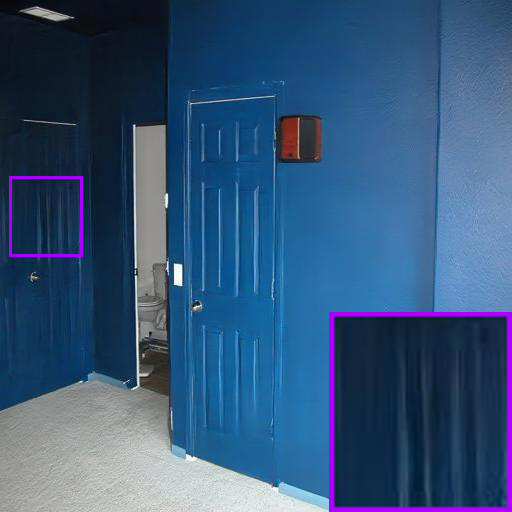}}
			{\tiny{\textbf{MANAS (Ours): 32.36/0.869}}}
		\end{minipage}%
	}%
	\subfigure{
		\begin{minipage}[t]{0.135\linewidth}
			\centering
			\centerline{\includegraphics[width=0.98in]{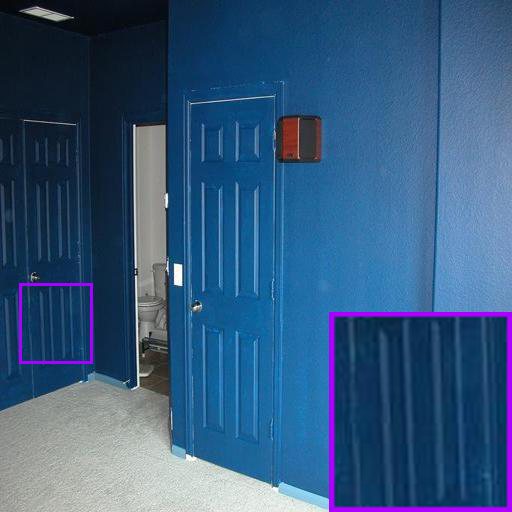}}
			{\tiny{Ground Truth: inf/1.000}}
			\vspace{4pt}
			\vspace{2.7pt}
			\centerline{\includegraphics[width=0.98in]{./Figs/gt_17.png}}
			{\tiny{Ground Truth: inf/1.000}}
			\vspace{4pt}
			\vspace{2pt}
			\centerline{\includegraphics[width=0.98in]{./Figs/gt_17.png}}
			{\tiny{Ground Truth: inf/1.000}}
		\end{minipage}%
	}%
	\centering
	\caption{Comparison of qualitative and quantitative results on three synthetic rainy images with the same background from the test set of DID-MDN~\cite{Zhang2018Density}. The purple boxes remark the local details after de-raining. Quantitative results PSNR(dB)/SSIM are documented below each image, and the best results are marked in bold.}
	\label{fig:DID-MDN}
\end{figure*} 
\subsection{Comparison with the State-of-the-Art}
\label{ssec:Comparison}

\subsubsection{Results on Synthetic Rain Removal}
\label{sssec:Synthetic}
\begin{figure*}[!t]
	\centering
	\subfigure{
		\begin{minipage}[t]{0.135\linewidth}
			\centering
			\centerline{\includegraphics[width=0.985in]{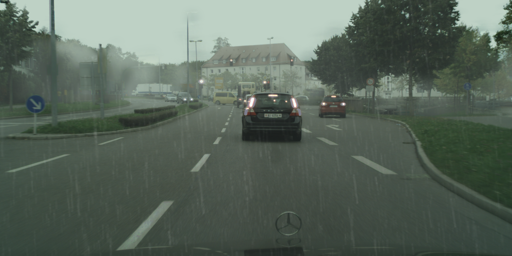}}
			{\tiny{Light Fog: 18.35/0.850}}
			\vspace{4pt}
			\vspace{2pt}
			\centerline{\includegraphics[width=0.985in]{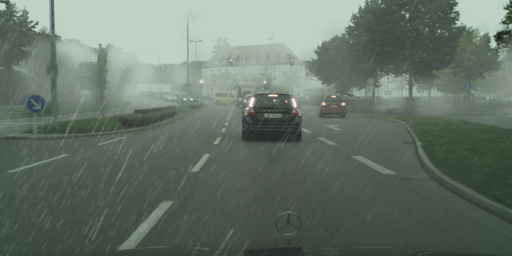}}
			{\tiny{Medium Fog: 14.58/0.751}}
			\vspace{4pt}
			\vspace{1.7pt}
			\centerline{\includegraphics[width=0.985in]{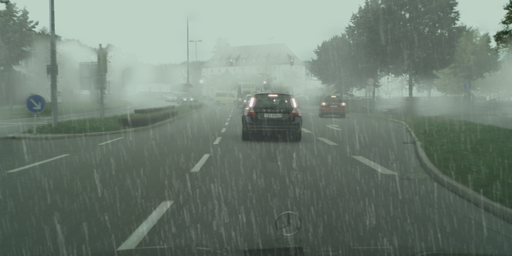}}
			{\tiny{Heavy Fog: 12.67/0.666}}
			\vspace{4pt}
		\end{minipage}%
	}%
	\subfigure{
		\begin{minipage}[t]{0.135\linewidth}
			\centering
			\centerline{\includegraphics[width=0.985in]{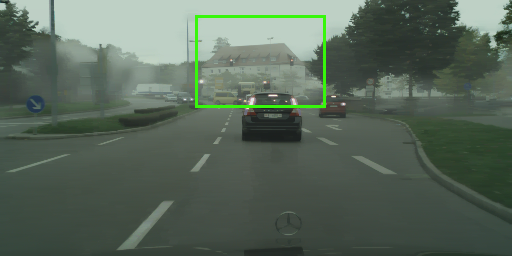}}
			{\tiny{JCAS~\cite{Gu2017Joint}: 18.43/0.841}}
			\vspace{4pt}
			\vspace{2.3pt}
			\centerline{\includegraphics[width=0.985in]{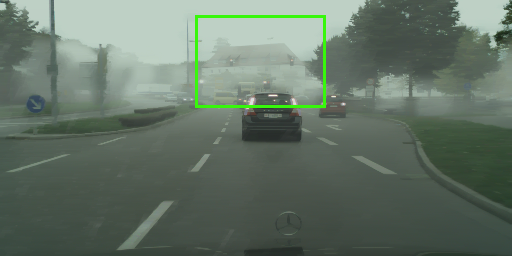}}
			{\tiny{JCAS~\cite{Gu2017Joint}: 14.64/0.761}}
			\vspace{4pt}
			\vspace{2pt}
			\centerline{\includegraphics[width=0.985in]{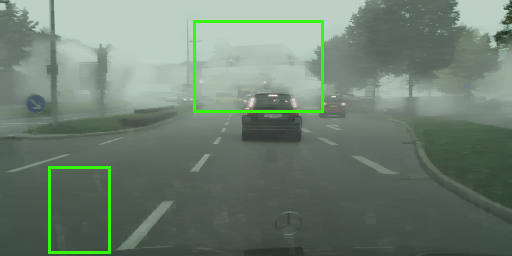}}
			{\tiny{JCAS~\cite{Gu2017Joint}: 12.73/0.710}}
			\vspace{4pt}
		\end{minipage}%
	}%
	\subfigure{
		\begin{minipage}[t]{0.135\linewidth}
			\centering
			\centerline{\includegraphics[width=0.985in]{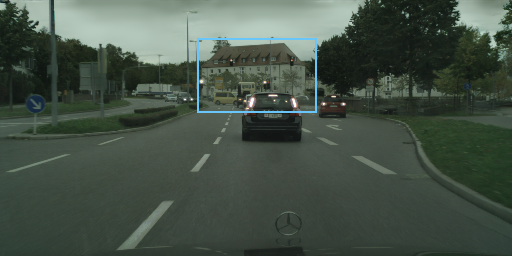}}
			{\tiny{PReNet~\cite{Ren2019Progressive}: 29.98/0.958}}
			\vspace{4pt}
			\vspace{2.3pt}
			\centerline{\includegraphics[width=0.985in]{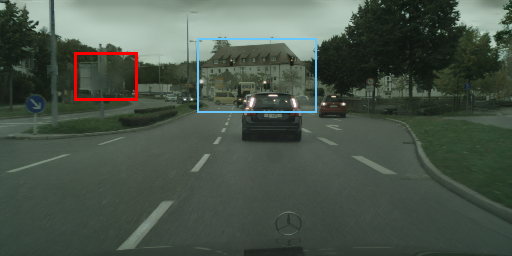}}
			{\tiny{PReNet~\cite{Ren2019Progressive}: 28.40/0.943}}
			\vspace{4pt}
			\vspace{2pt}
			\centerline{\includegraphics[width=0.985in]{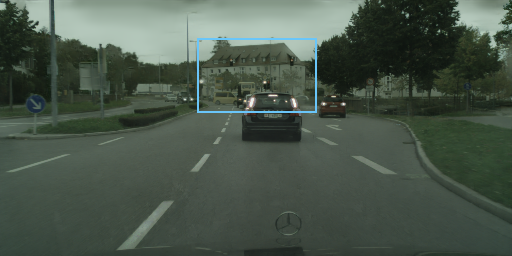}}
			{\tiny{PReNet~\cite{Ren2019Progressive}: 27.61/0.934}}
			\vspace{4pt}
		\end{minipage}%
	}%
	\subfigure{
		\begin{minipage}[t]{0.135\linewidth}
			\centering
			\centerline{\includegraphics[width=0.985in]{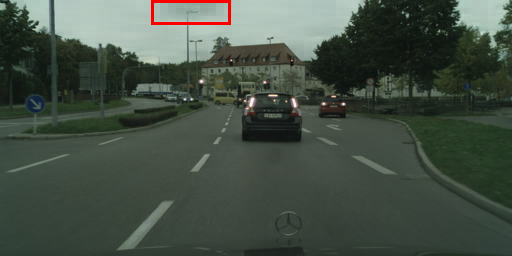}}
			{\tiny{CLEARER~\cite{Gou2020CLEARER}: 32.90/0.973}}
			\vspace{4pt}
			\vspace{2.3pt}
			\centerline{\includegraphics[width=0.985in]{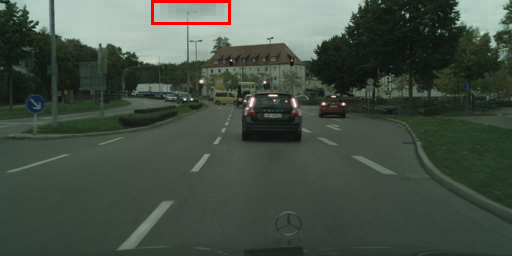}}
			{\tiny{CLEARER~\cite{Gou2020CLEARER}: 32.81/0.971}}
			\vspace{4pt}
			\vspace{2pt}
			\centerline{\includegraphics[width=0.985in]{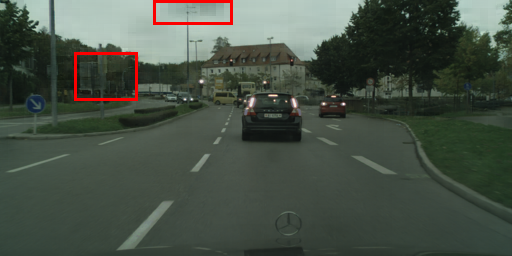}}
			{\tiny{CLEARER~\cite{Gou2020CLEARER}: 30.99/0.954}}
			\vspace{4pt}
		\end{minipage}%
	}%
	\subfigure{
		\begin{minipage}[t]{0.135\linewidth}
			\centering
			\centerline{\includegraphics[width=0.98in]{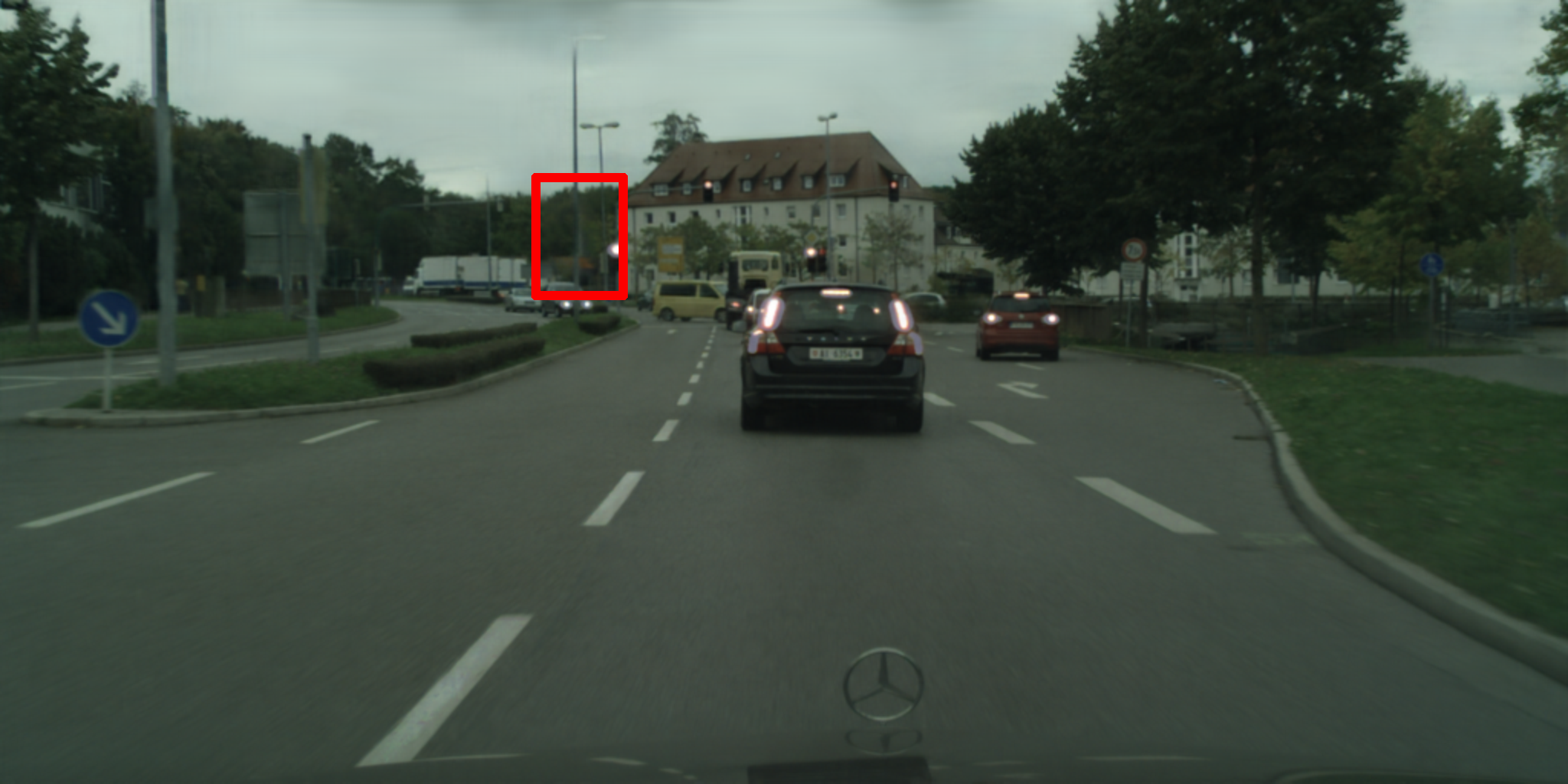}}
			{\tiny{DGNL-Net~\cite{Hu2021Single}: 33.50/0.968}}
			\vspace{4pt}
			\vspace{2.3pt}
			\centerline{\includegraphics[width=0.985in]{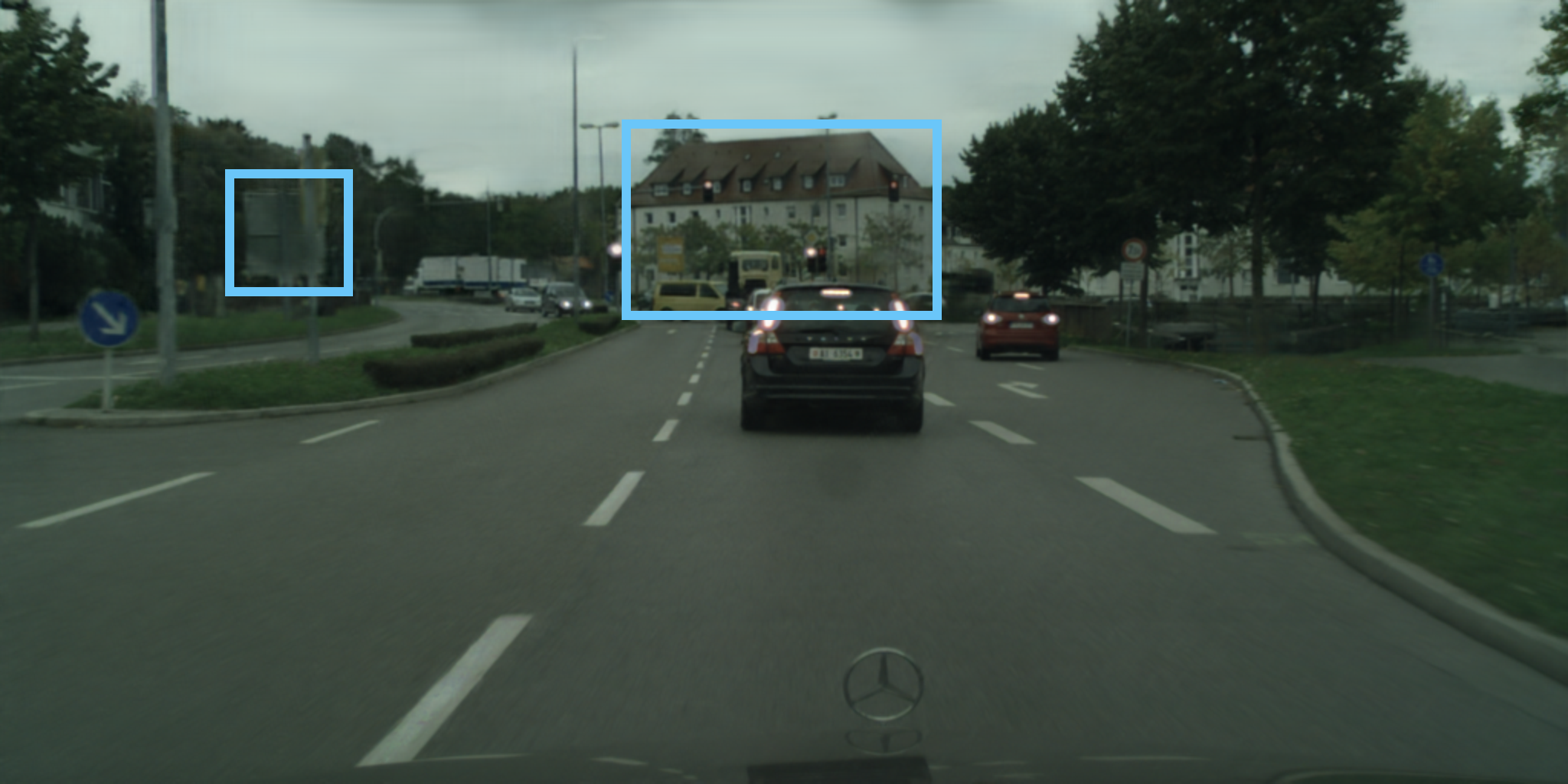}}
			{\tiny{DGNL-Net~\cite{Hu2021Single}: 32.14/0.959}}
			\vspace{4pt}
			\vspace{2pt}
			\centerline{\includegraphics[width=0.985in]{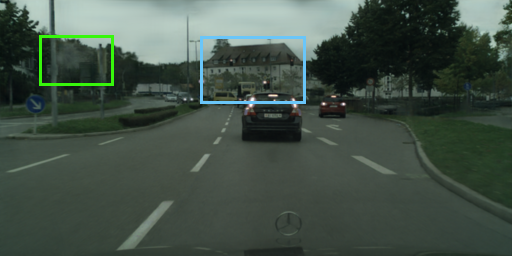}}
			{\tiny{DGNL-Net~\cite{Hu2021Single}: 30.38/0.944}}
			\vspace{4pt}
		\end{minipage}%
	}%
	\subfigure{
		\begin{minipage}[t]{0.135\linewidth}
			\centering
			\centerline{\includegraphics[width=0.985in]{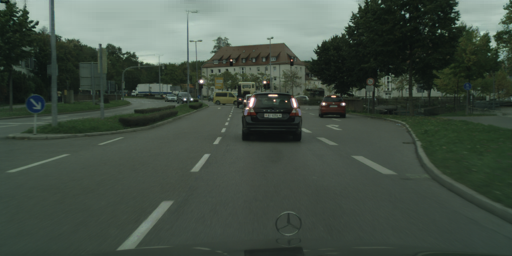}}
			{\tiny{\textbf{MANAS (Ours)}: \textbf{35.75/0.980}}}
			\vspace{4pt}
			\vspace{2.3pt}
			\centerline{\includegraphics[width=0.985in]{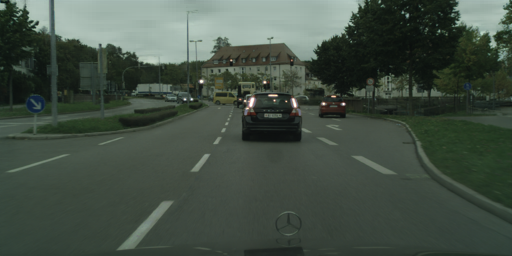}}
			{\tiny{\textbf{MANAS (Ours)}: \textbf{34.57/0.974}}}
			\vspace{4pt}
			\vspace{1.8pt}
			\centerline{\includegraphics[width=0.985in]{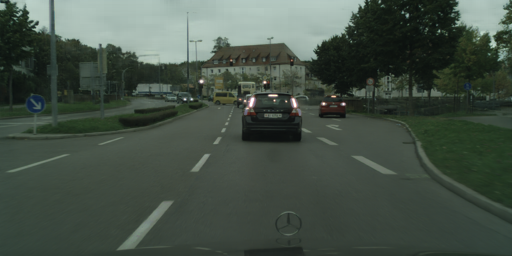}}
			{\tiny{\textbf{MANAS (Ours)}: \textbf{32.75/0.963}}}
			\vspace{4pt}
		\end{minipage}%
	}%
	\subfigure{
		\begin{minipage}[t]{0.135\linewidth}
			\centering
			\centerline{\includegraphics[width=0.985in]{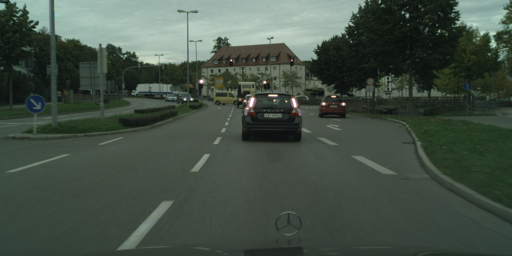}}
			{\tiny{Ground Truth: inf/1.000}}
			\vspace{4pt}
			\vspace{3.0pt}
			\centerline{\includegraphics[width=0.985in]{./Figs/gt_568.png}}
			{\tiny{Ground Truth: inf/1.000}}
			\vspace{4pt}
			\vspace{2.6pt}
			\centerline{\includegraphics[width=0.985in]{./Figs/gt_568.png}}
			{\tiny{Ground Truth: inf/1.000}}
			\vspace{4pt}
		\end{minipage}%
	}%
	\centering
	\caption{Comparison of qualitative and quantitative results on three synthetic rainy images with the same background from the test set of RainCistyscapes~\cite{Hu2021Single}. The green boxes remark the regions where remain some rain streaks or fog, the blue boxes remark the regions with color distortion, and the red boxes remark the regions with artifacts. Please zoom in to see the details.}
	\label{fig:RainCityscapes}
\end{figure*} 
\begin{table*}[t]
	\renewcommand{\arraystretch}{1.2}
	\tabcolsep0.15cm
	\centering
	\caption{Comparison of average score results by different methods on DQA~\cite{Wu2020Subjective} database.}
	\label{tab:score}       
	\begin{tabular}{cccccccccccc}
		\Xhline{1.3pt}
		\multirow{1}{*}{Method} 
		& 
		\multicolumn{1}{|c}{Rainy inputs} &
		\multicolumn{1}{|c}{JCAS~\cite{Gu2017Joint}}    & \multicolumn{1}{c}{PReNet~\cite{Ren2019Progressive}} & \multicolumn{1}{c}{CLEARER~\cite{Gou2020CLEARER}}  
		& \multicolumn{1}{c}{DGNL-Net~\cite{Hu2021Single}} 
		& \multicolumn{1}{c}{\textbf{MANAS (Ours)}} \\
		\hline
		\multirow{1}{*}{Average Score $\uparrow$} 
		& \multicolumn{1}{|c}{0.2997}  
		& \multicolumn{1}{|c}{0.3202} 
		& \multicolumn{1}{c}{0.3025} 
		& \multicolumn{1}{c}{0.3216} 
		& \multicolumn{1}{c}{0.3109} 
		& \multicolumn{1}{c}{\textbf{0.3281}} \\ 
		\Xhline{1.3pt}
	\end{tabular}
\end{table*}

\begin{figure*}[!t]
	\centering
	\subfigure{
		\begin{minipage}[t]{0.16\linewidth}
			\centering
			\centerline{\includegraphics[width=1.17in]{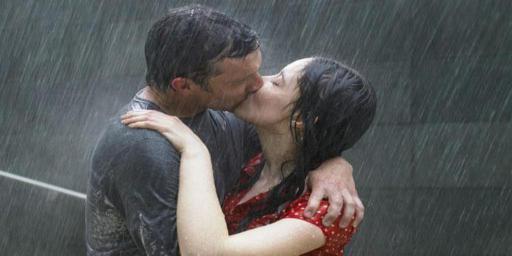}}
			{\scriptsize{Rainy Image: 0.3304}}
			\vspace{4pt}
			\vspace{1.7pt}
			\centerline{\includegraphics[width=1.17in]{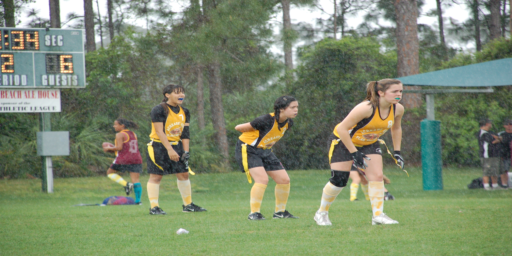}}
			{\scriptsize{Rainy Image: 0.3939}}
			\vspace{4pt}
			\vspace{1.5pt}
			\centerline{\includegraphics[width=1.17in]{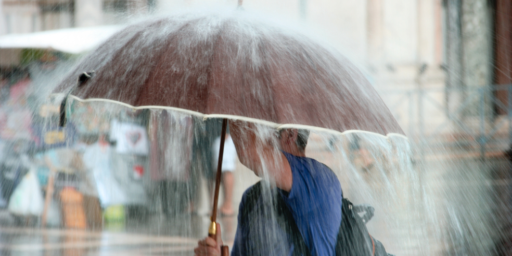}}
			{\scriptsize{Rainy Image: 0.3575}}
		\end{minipage}%
	}%
	\subfigure{
		\begin{minipage}[t]{0.16\linewidth}
			\centering
			\centerline{\includegraphics[width=1.17in]{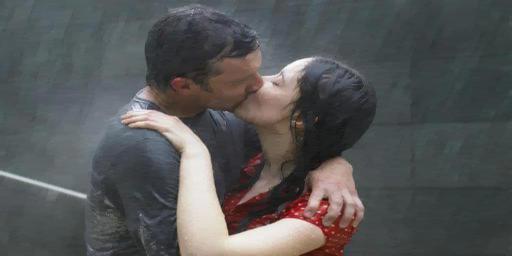}}
			{\scriptsize{JCAS~\cite{Gu2017Joint}: 0.5279}} 
			\vspace{4pt}
			\vspace{2pt}
			\centerline{\includegraphics[width=1.17in]{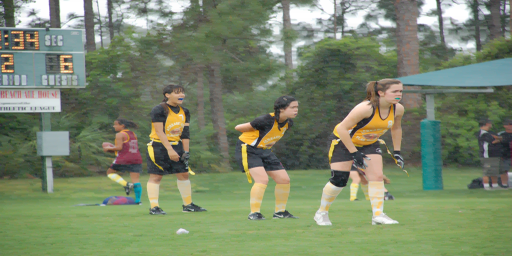}}
			{\scriptsize{JCAS~\cite{Gu2017Joint}: 0.4131}}
			\vspace{4pt}
			\vspace{2pt}
			\centerline{\includegraphics[width=1.17in]{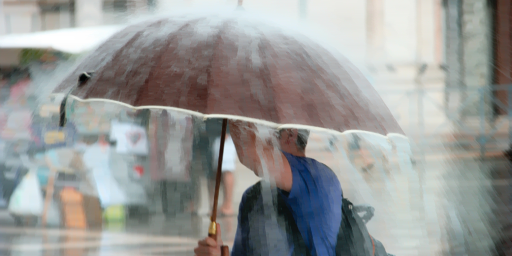}}
			{\scriptsize{JCAS~\cite{Gu2017Joint}: 0.2638}}
		\end{minipage}%
	}%
	\subfigure{
		\begin{minipage}[t]{0.16\linewidth}
			\centering
			\centerline{\includegraphics[width=1.17in]{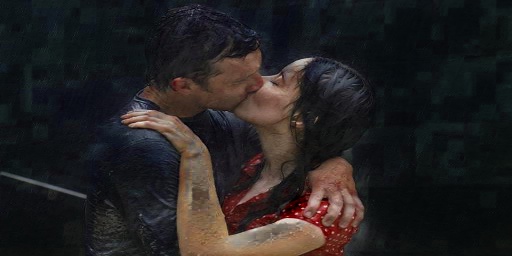}}
			{\scriptsize{PReNet~\cite{Ren2019Progressive}: 0.5317}} 
			\vspace{4pt}
			\vspace{2pt}
			\centerline{\includegraphics[width=1.17in]{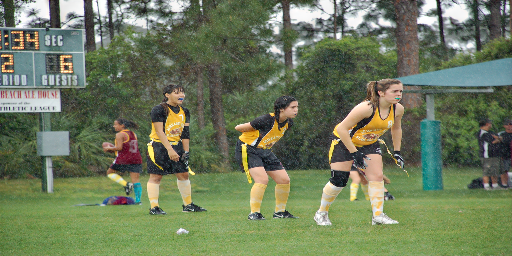}}
			{\scriptsize{PReNet~\cite{Ren2019Progressive}: 0.3023}}
			\vspace{4pt}
			\vspace{2pt}
			\centerline{\includegraphics[width=1.17in]{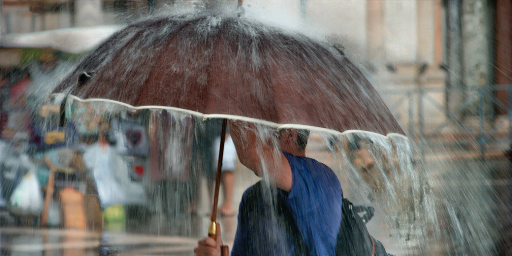}}
			{\scriptsize{PReNet~\cite{Ren2019Progressive}: 0.3298}}
		\end{minipage}%
	}%
	\subfigure{
		\begin{minipage}[t]{0.16\linewidth}
			\centering
			\centerline{\includegraphics[width=1.17in]{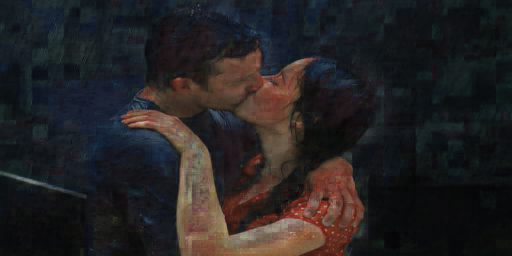}}
			{\scriptsize{CLEARER~\cite{Gou2020CLEARER}: 0.4780}}
			\vspace{4pt}
			\vspace{2pt}
			\centerline{\includegraphics[width=1.17in]{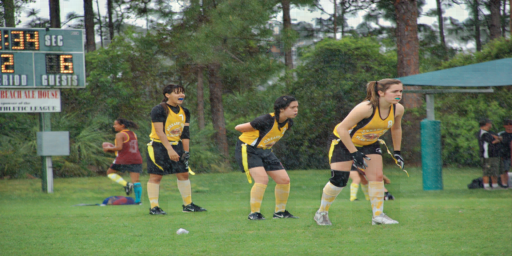}}
			{\scriptsize{CLEARER~\cite{Gou2020CLEARER}: 0.4086}}
			\vspace{4pt}
			\vspace{2pt}
			\centerline{\includegraphics[width=1.17in]{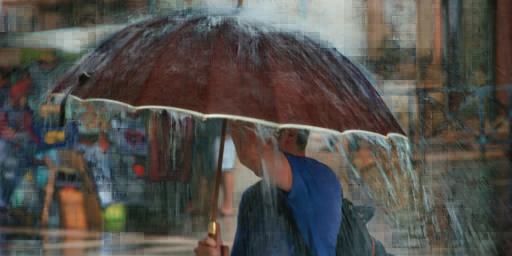}}
			{\scriptsize{CLEARER~\cite{Gou2020CLEARER}: 0.3307}}
		\end{minipage}%
	}%
	\subfigure{
		\begin{minipage}[t]{0.16\linewidth}
			\centering
			\centerline{\includegraphics[width=1.17in]{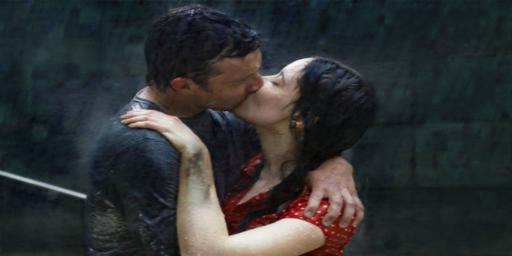}}
			{\scriptsize{DGNL-Net~\cite{Hu2021Single}: 0.4874}}
			\vspace{4pt}
			\vspace{2pt}
			\centerline{\includegraphics[width=1.17in]{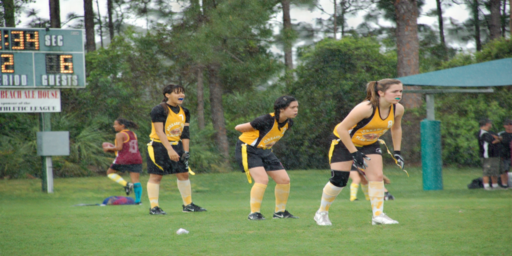}}
			{\scriptsize{DGNL-Net~\cite{Hu2021Single}: 0.3962}}
			\vspace{4pt}
			\vspace{2pt}
			\centerline{\includegraphics[width=1.17in]{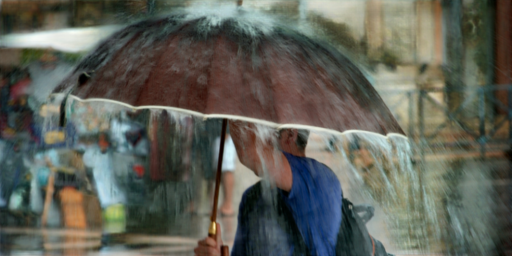}}
			{\scriptsize{DGNL-Net~\cite{Hu2021Single}: 0.3182}}
		\end{minipage}%
	}%
	\subfigure{
		\begin{minipage}[t]{0.16\linewidth}
			\centering
			\centerline{\includegraphics[width=1.17in]{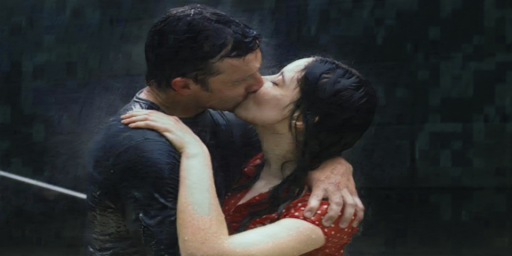}}
			{\scriptsize{\textbf{MANAS (Ours): 0.5466}}}
			\vspace{4pt}
			\vspace{1.8pt}
			\centerline{\includegraphics[width=1.17in]{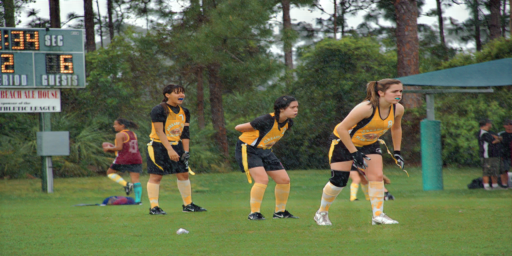}}
			{\scriptsize{\textbf{MANAS (Ours): 0.4409}}}
			\vspace{4pt}
			\vspace{2.0pt}
			\centerline{\includegraphics[width=1.17in]{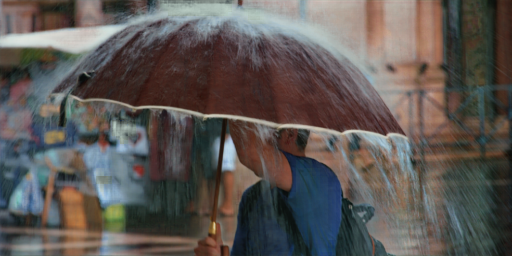}}
			{\scriptsize{\textbf{MANAS (Ours): 0.3823}}}
		\end{minipage}%
	}%
	\centering
	\caption{Comparison of qualitative and quantitative results on real rainy images from DQA~\cite{Wu2020Subjective} database. All of the de-rained images were derived from the de-raining models that had been trained on the RainCityscapes~\cite{Hu2021Single} dataset, except JCAS~\cite{Gu2017Joint}. Quantitative results (\emph{i.e.}, predicted quality score by B-FEN model~\cite{Wu2020Subjective}) are documented below each image, and the best results are marked in bold.}
	\label{fig:DQA}
\end{figure*} 

\begin{figure*}[!t]
	\centering
	\subfigure{
		\begin{minipage}[t]{0.16\linewidth}
			\centering
			\centerline{\includegraphics[width=1.17in]{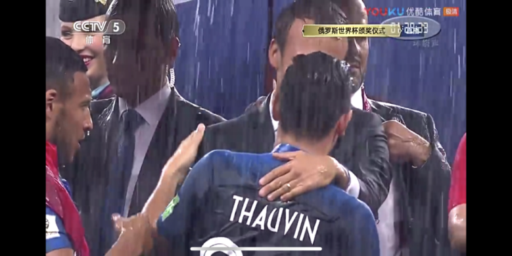}}
			{\scriptsize{Rainy Image: 0.3386}}
			\vspace{4pt}
			\vspace{1.7pt}
			\centerline{\includegraphics[width=1.17in]{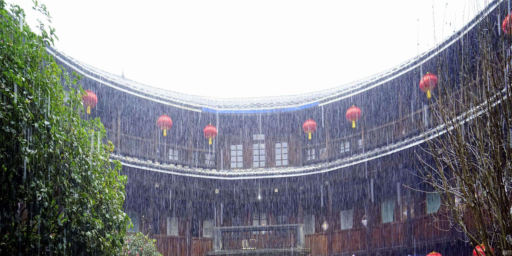}}
			{\scriptsize{Rainy Image: 0.3294}}
			\vspace{4pt}
			\vspace{1.5pt}
			\centerline{\includegraphics[width=1.17in]{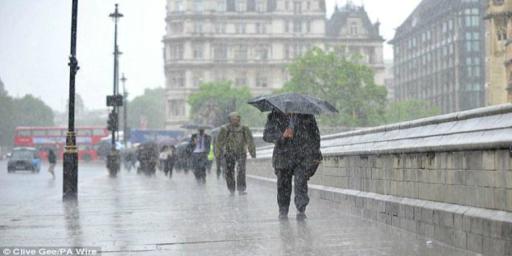}}
			{\scriptsize{Rainy Image: 0.2952}}
		\end{minipage}%
	}%
	\subfigure{
		\begin{minipage}[t]{0.16\linewidth}
			\centering
			\centerline{\includegraphics[width=1.17in]{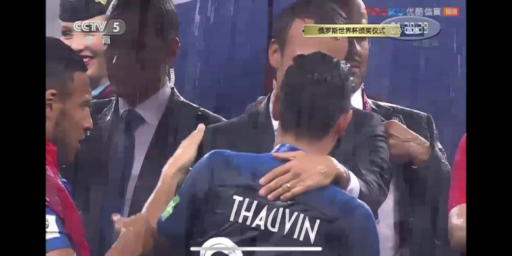}}
			{\scriptsize{JCAS~\cite{Gu2017Joint}: 0.3505}}  
			\vspace{4pt}
			\vspace{2pt}
			\centerline{\includegraphics[width=1.17in]{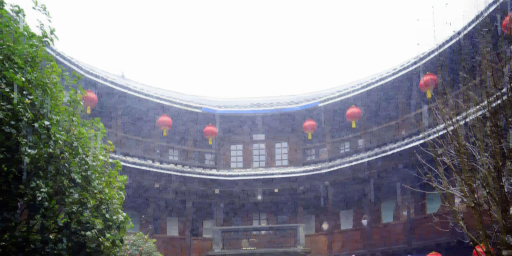}}
			{\scriptsize{JCAS~\cite{Gu2017Joint}: 0.3564}}
			\vspace{4pt}
			\vspace{2pt}
			\centerline{\includegraphics[width=1.17in]{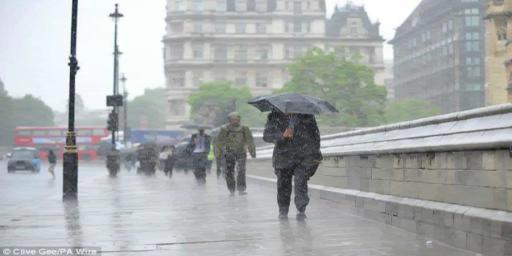}}
			{\scriptsize{JCAS~\cite{Gu2017Joint}: 0.2864}}
		\end{minipage}%
	}%
	\subfigure{
		\begin{minipage}[t]{0.16\linewidth}
			\centering
			\centerline{\includegraphics[width=1.17in]{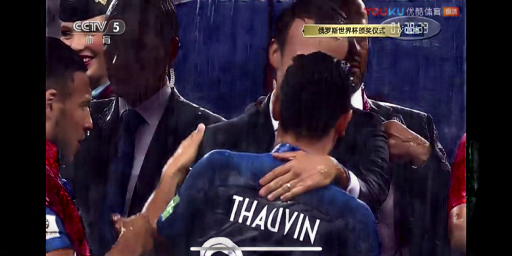}}
			{\scriptsize{PReNet~\cite{Ren2019Progressive}: 0.3697}} 
			\vspace{4pt}
			\vspace{2pt}
			\centerline{\includegraphics[width=1.17in]{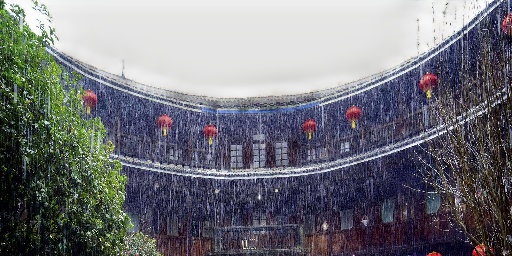}}
			{\scriptsize{PReNet~\cite{Ren2019Progressive}: 0.3432}}
			\vspace{4pt}
			\vspace{2pt}
			\centerline{\includegraphics[width=1.17in]{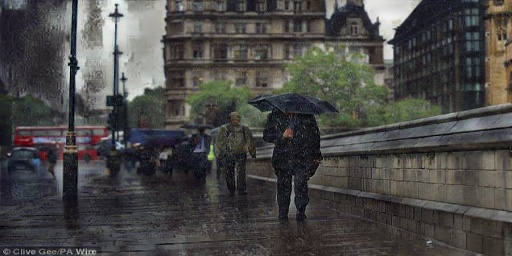}}
			{\scriptsize{PReNet~\cite{Ren2019Progressive}: 0.3196}}
		\end{minipage}%
	}%
	\subfigure{
		\begin{minipage}[t]{0.16\linewidth}
			\centering
			\centerline{\includegraphics[width=1.17in]{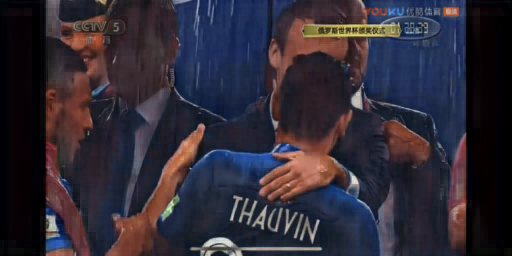}}
			{\scriptsize{CLEARER~\cite{Gou2020CLEARER}: 0.3388}}
			\vspace{4pt}
			\vspace{2pt}
			\centerline{\includegraphics[width=1.17in]{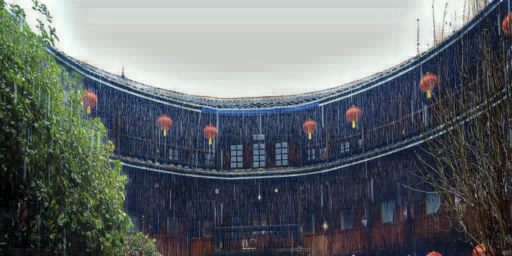}}
			{\scriptsize{CLEARER~\cite{Gou2020CLEARER}: 0.3641}}
			\vspace{4pt}
			\vspace{2pt}
			\centerline{\includegraphics[width=1.17in]{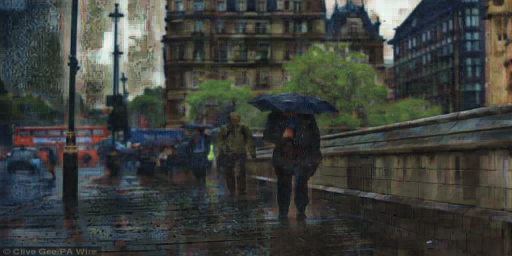}}
			{\scriptsize{CLEARER~\cite{Gou2020CLEARER}: 0.3219}}
		\end{minipage}%
	}%
	\subfigure{
		\begin{minipage}[t]{0.16\linewidth}
			\centering
			\centerline{\includegraphics[width=1.17in]{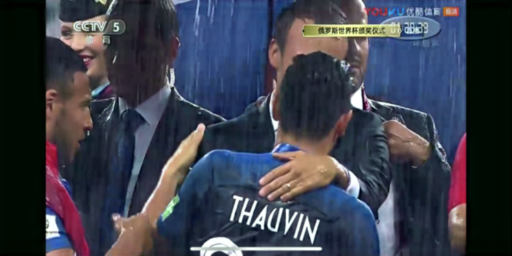}}
			{\scriptsize{DGNL-Net~\cite{Hu2021Single}: 0.3477}}
			\vspace{4pt}
			\vspace{2pt}
			\centerline{\includegraphics[width=1.17in]{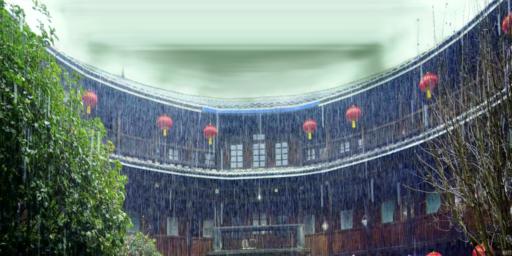}}
			{\scriptsize{DGNL-Net~\cite{Hu2021Single}: 0.3339}}
			\vspace{4pt}
			\vspace{2pt}
			\centerline{\includegraphics[width=1.17in]{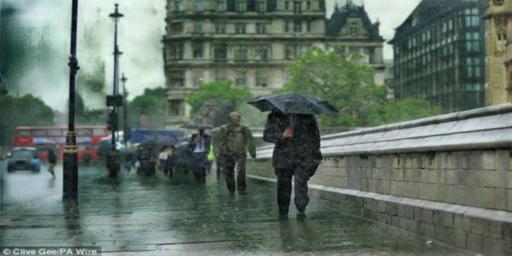}}
			{\scriptsize{DGNL-Net~\cite{Hu2021Single}: 0.3038}}
		\end{minipage}%
	}%
	\subfigure{
		\begin{minipage}[t]{0.16\linewidth}
			\centering
			\centerline{\includegraphics[width=1.17in]{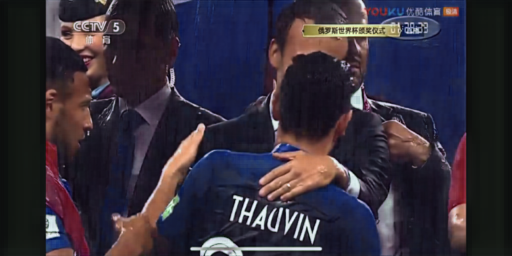}}
			{\scriptsize{\textbf{MANAS (Ours): 0.4283}}}
			\vspace{4pt}
			\vspace{2pt}
			\centerline{\includegraphics[width=1.17in]{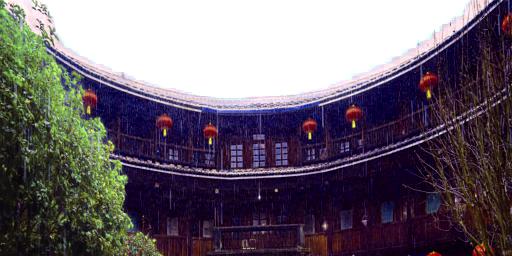}}
			{\scriptsize{\textbf{MANAS (Ours): 0.3672}}}
			\vspace{4pt}
			\vspace{2pt}
			\centerline{\includegraphics[width=1.17in]{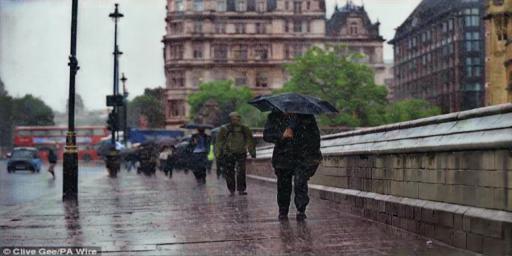}}
			{\scriptsize{\textbf{MANAS (Ours): 0.3263}}}
		\end{minipage}%
	}%
	\centering
	\caption{Comparison of qualitative and quantitative results on real rainy images collected from the Internet. Likewise, quantitative results (\emph{i.e.}, predicted quality score by B-FEN model~\cite{Wu2020Subjective}) are documented below each image.}
	\label{fig:real_derain}
\end{figure*} 

Table~\ref{tab:synthetic} shows the quantitative results of the proposed method and previous image de-raining methods
on two synthetic datasets,~\emph{i.e.}, DID-MDN~\cite{Zhang2018Density} and RainCityscapes~\cite{Hu2021Single}. Regarding the previous image de-raining methods, if authors have publicly released their de-raining models which had been trained with the training set of DID-MDN~\cite{Zhang2018Density} and RainCityscapes~\cite{Hu2021Single} datasets, respectively, then they were directly used to evaluate the performance on the corresponding testing set. These pre-trained models were trained and tested according to the original settings (\emph{e.g.}, hyper-parameters and input size) mentioned in their corresponding published papers. 
On the other hand, for those de-raining methods whose authors did not provide the pre-trained models, we re-trained models from scratch with their publicly released codes and followed their original settings under these two datasets. After training, we used the well-trained models to evaluate the performance on the corresponding testing set.


As shown in Table~\ref{tab:synthetic}, our MANAS method exceeds other competing methods and achieves the best average performance. In particular, the proposed MANAS method is superior to the second best method,~\emph{i.e.}, DGNL-Net~\cite{Hu2021Single},
by $1.47$ dB, $0.011$, and $0.28$ in terms of average PSNR, SSIM, and NIQE, respectively. 
In addition to quantitative comparisons, our method also outperforms other competing methods in qualitative comparisons. As shown in Fig.~\ref{fig:DID-MDN}, three representative synthetic rainy images with the same background scene were selected from DID-MDN dataset to perform the rain streak removal. It can be observed that our proposed MANAS method removes rain streaks more clearly, and meanwhile preserve image details better than previous image de-raining methods. Furthermore, we also adopted three representative synthetic rainy images selected from RainCityscapes~\cite{Hu2021Single} dataset to conduct the rain streak and fog removal. As shown in Fig.~\ref{fig:RainCityscapes}, our method better removes the rain streaks
and fog in the line of sight, while other de-raining methods could still remain some rain streaks or fog, or produce undesirable local artifacts and color distortion in the de-rained images, please zoom in to see the details.   
The superior performance on quantitative and qualitative evaluations consistently demonstrates the effectiveness of our MANAS method on synthetic rain removal.



\subsubsection{Results on Real-world Rain Removal}
\label{sssec:Realistic}
To demonstrate the practicality, we conduct real-world rain removal on the De-raining Quality Assessment (DQA)~\cite{Wu2020Subjective} database which contains $206$ real rainy images. To make a fair comparison, except the hand-crafted prior-based de-raining methods requiring no training, all of the competing methods were trained on the RainCityscapes dataset and then applied to remove rain from the real rainy images. The quantitative results, \emph{i.e.}, average score over $206$ de-rained images, are tabulated in Table~\ref{tab:score}. As can be seen, our MANAS achieves a higher average score than other de-raining methods compared against.
Besides, six real-world rainy images were selected from the DQA database and the Internet to conduct de-raining visual comparisons. As shown in Fig.~\ref{fig:DQA} and Fig.~\ref{fig:real_derain}, the JCAS~\cite{Gu2017Joint}, which belongs to the hand-crafted prior-based de-raining method, fails to remove the fog that comes with the rain. And other competing methods either cannot well operate the rain streaks in the images or more easily produce undesirable artifacts and color distortion. In contrast, our method can more sufficiently remove the rain streaks and fog in the real-world rainy images, leading to vivid and clear de-rained results. This implies that our MANAS method is also effective for real-world rain removal and outperforms previous de-raining methods.




\subsubsection{Preprocessing for High-level Vision Tasks}
\label{sssec:Application}
\begin{table}[t]
	\renewcommand{\arraystretch}{1.2}
	\tabcolsep0.15cm
	\centering
	\caption{Comparison of mAP ($\%$) and mIoU ($\%$) results on RainCityscapes~\cite{Hu2021Single}.}
	\label{tab:object}       
	\begin{tabular}{cccccccccccc}
		\Xhline{1.3pt}
		\multirow{1}{*}{Method} 
		& \multicolumn{1}{|c}{mAP $\uparrow$}    & \multicolumn{1}{c}{mIoU $\uparrow$}   \\
		\hline
		\multirow{1}{*}{Rainy images} & \multicolumn{1}{|c}{25.25}  & \multicolumn{1}{c}{40.89}   \\ 
		\hline
		\multirow{1}{*}{De-rained images (by MANAS)} & \multicolumn{1}{|c}{\textbf{32.88}}  & \multicolumn{1}{c}{\textbf{46.61}}   \\
		\hline
		\multirow{1}{*}{Ground-truth images} & \multicolumn{1}{|c}{34.79}  & \multicolumn{1}{c}{48.39}  \\
		\Xhline{1.3pt}
	\end{tabular}
\end{table}
\begin{figure}[!t]
	\begin{center}
		\centering
		\subfigure[Rainy Image]{
			\begin{minipage}[t]{0.48\linewidth}
				\centering
				\centerline{\includegraphics[width=1.68in]{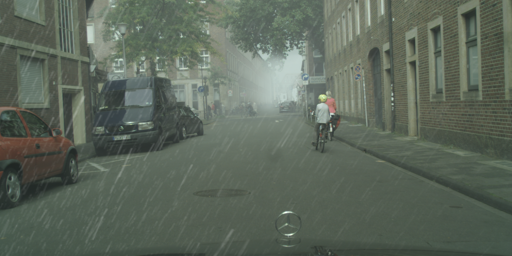}}
				\vspace{4pt}
				\centerline{\includegraphics[width=1.68in]{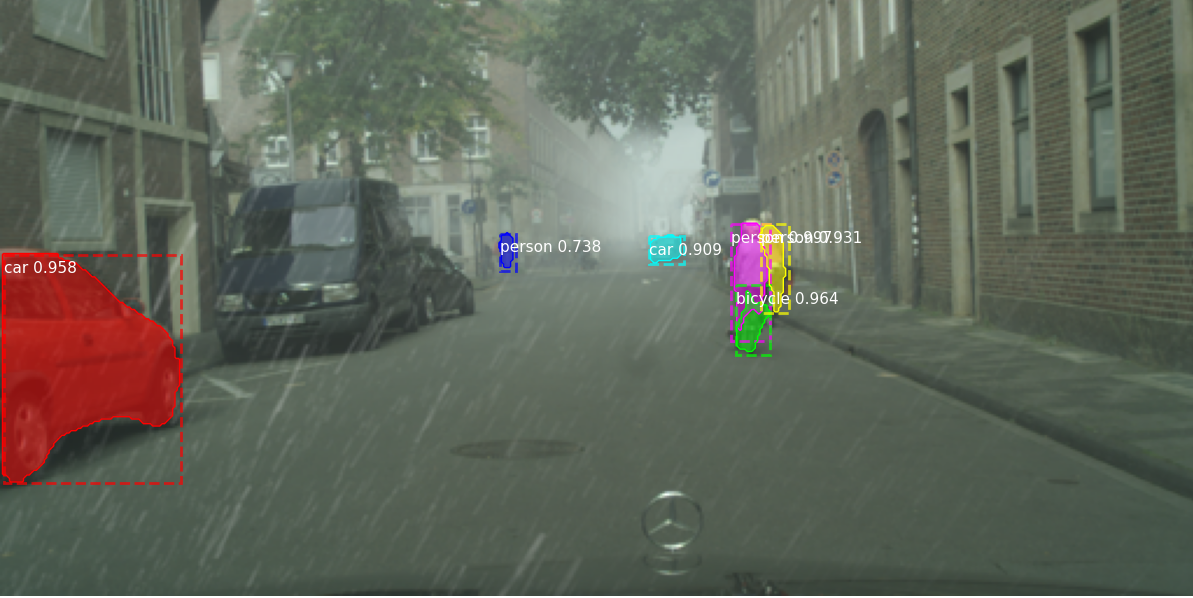}}
				\vspace{4pt}
			\end{minipage}%
		}%
		\subfigure[De-rained Image]{
			\begin{minipage}[t]{0.48\linewidth}
				\centering
				\centerline{\includegraphics[width=1.68in]{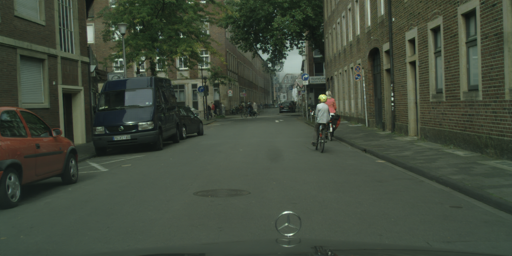}}
				\vspace{4pt}
				\centerline{\includegraphics[width=1.68in]{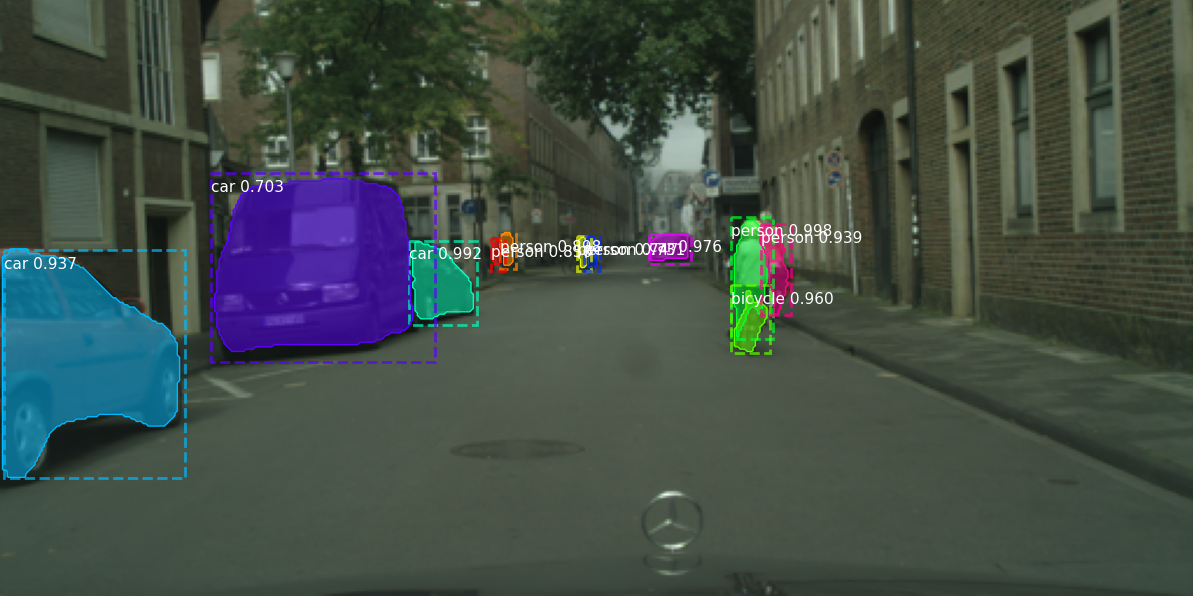}}
				\vspace{4pt}
			\end{minipage}%
		}%
		\centering
		\caption{Object detection and segmentation results on a (a) synthetic rainy image from the RainCityscapes~\cite{Hu2021Single}, and on its corresponding (b) de-rained image produced by our MANAS method. 
			The first row shows the reference images before conducting the object detection and segmentation task.}
		\label{fig:applications}
	\end{center}
\end{figure} 
\begin{figure}[!t]
	\centering
	\subfigure[Rainy Image]{
		\begin{minipage}[t]{0.48\linewidth}
			\centering
			\centerline{\includegraphics[width=1.68in]{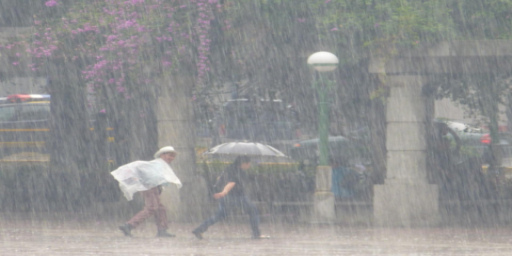}}
			\vspace{4pt}
			\centerline{\includegraphics[width=1.68in]{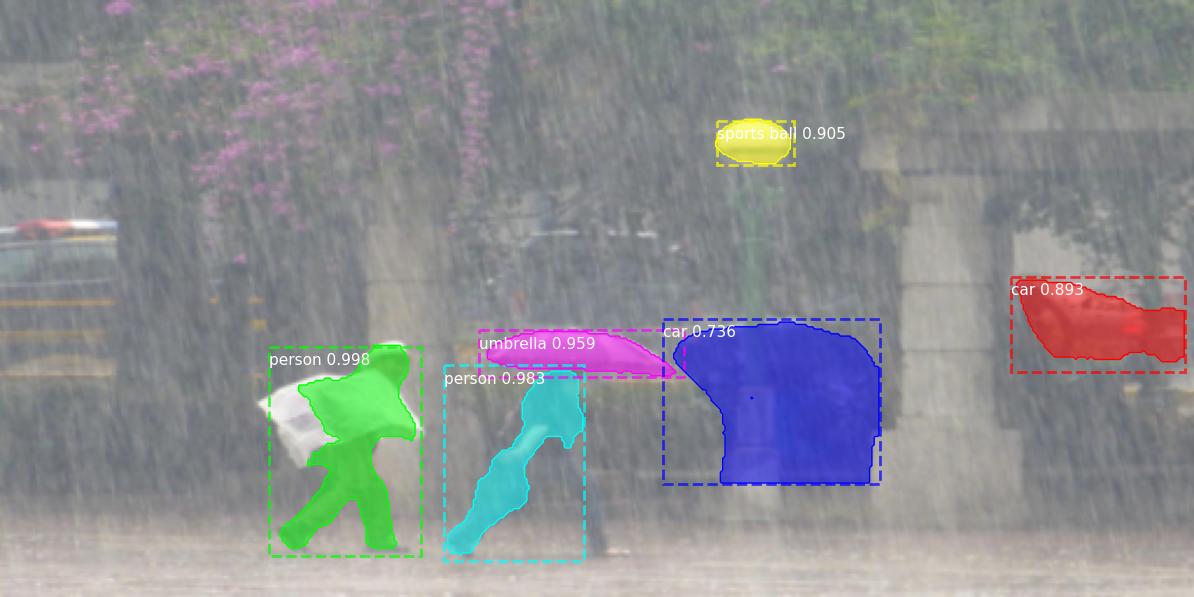}}
			\vspace{4pt}
		\end{minipage}%
	}%
	\subfigure[De-rained Image]{
		\begin{minipage}[t]{0.48\linewidth}
			\centering
			\centerline{\includegraphics[width=1.68in]{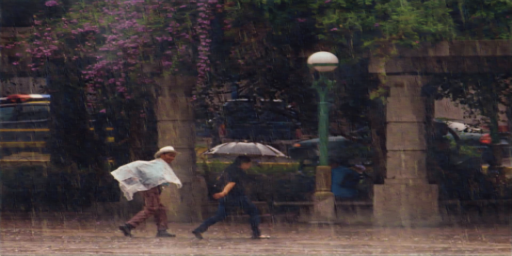}}
			\vspace{4pt}
			\centerline{\includegraphics[width=1.68in]{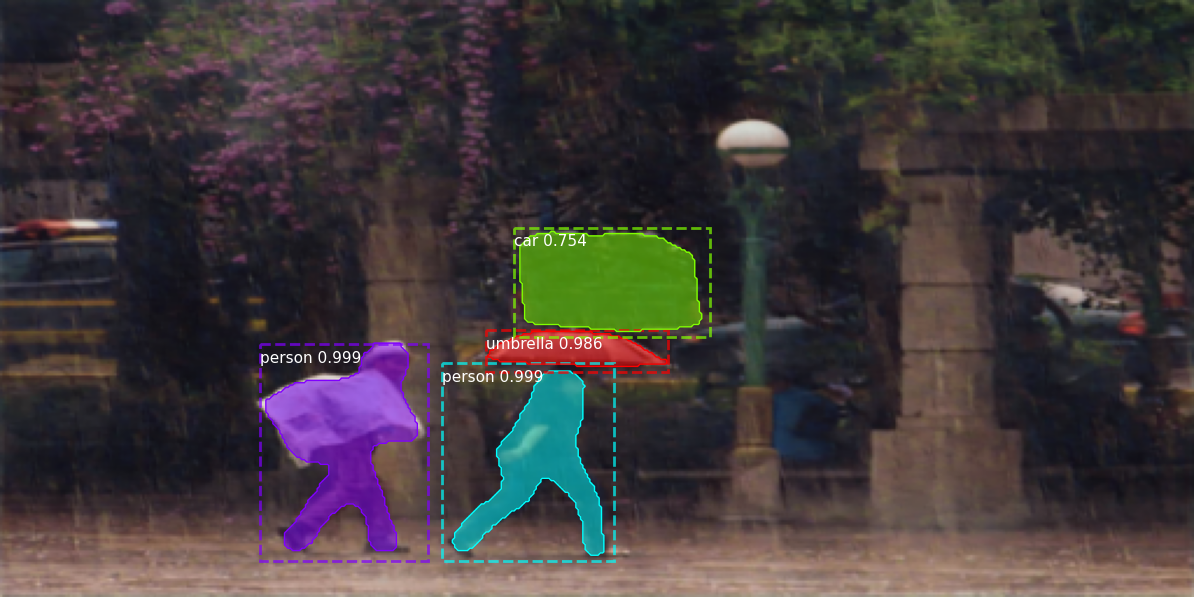}}
			\vspace{4pt}
		\end{minipage}%
	}%
	\centering
	\caption{Object detection and segmentation results on a (a) real rainy image, and on its corresponding (b) de-rained image produced by our MANAS method. The first row shows the reference images before conducting the object detection and segmentation task.}
	\label{fig:applications_real}
\end{figure}
Image de-raining can be applied to enhance the image content that has been degraded by the rain, thus such a technique has great potential for improving the visibility of objects under various rainy weather conditions~\cite{Cai2021Joint}. 
In this section, we investigate how our MANAS method contributes to improving the performance of object detection and segmentation. Specifically, we downloaded the publicly released Mask R-CNN model~\cite{He2017Mask} which had been pre-trained on the COCO~\cite{Lin2014Microsoft} dataset, and then used it to detect and segment the objects in rainy images, de-rained images (by MANAS), and rain-free ground-truth images inside the RainCityscapes~\cite{Hu2021Single} dataset. Table~\ref{tab:object} reports the quantitative results in terms of mean Average Precision (mAP) and mean Intersection over Union (mIoU). It is obvious that, after applying our method for image de-raining, the mAP and mIoU of de-rained images show notable improvements over those of original rainy images by $7.63\%$ and $5.72\%$, respectively. 
In addition, Fig.~\ref{fig:applications} and Fig.~\ref{fig:applications_real} show the visual results of the object detection and segmentation before and after applying our MANAS method for removing the rain streaks and fog in the images. It can be intuitively observed that our method helps improve the performance of object detection and segmentation.
This study implies that the proposed MANAS has the potential to be used as a pre-processing step for the object detection and segmentation task.


\subsection{Validation of the Proposed MANAS Method}
\label{ssec:Ablation}
To further validate the proposed MANAS method, we conduct deeper analyses on DID-MDN~\cite{Zhang2018Density} dataset. 
Specifically, we have investigated the following three aspects: 1) the importance of the multi-scale attentive neural architecture for image de-raining; 2) the effectiveness of our multi-to-one training strategy for improving the robustness; 3) the role of the model complexity loss $\mathcal{L}_{\text{comp}}$ on controlling the trade-off between performance and model size.


\subsubsection{How Important Is the Multi-scale Attentive Neural Architecture for Image De-raining?}
As mentioned earlier, all of the multi-scale attentive cells in the de-raining network integrate the novel multi-scale attentive neural architecture, and the more cells in the de-raining network, the more high-scale to low-scale attentive sub-networks will be incorporated into the de-raining network.
Accordingly, to demonstrate the importance of the multi-scale attentive neural architecture for image de-raining, we have conducted an ablation experiment to evaluate the quantitative performance after we gradually reduce the number of cells in the de-raining network. 
The corresponding results are shown in Table.~\ref{tab:cell_increase}. We observe that as we reduced the number of multi-scale attentive cells, the performances on PSNR, SSIM, and NIQE are consistently degraded. This is due to the fact that the fewer multi-scale attentive cells will lead to fewer high-scale to low-scale attentive sub-networks, which in turn weakens the representation ability of the de-raining network to learn the highly representative and discriminative multi-scale attentive features to handle the image de-raining task. 
Furthermore, once we remove all of the multi-scale attentive cells (\emph{i.e.}, $\text{T}=0$), the performance drops drastically, as shown in Table.~\ref{tab:cell_increase}, only $25.66$ (dB)/$0.825$/$3.79$ on PSNR/SSIM/NIQE, much lower than those with multi-scale attentive cells in the de-raining network. This study clearly demonstrates the importance of the multi-scale attentive neural architecture for image de-raining.
\begin{table}[t]
	\renewcommand{\arraystretch}{1.2}
	\tabcolsep0.15cm
	\centering
	\caption{PSNR, SSIM, and NIQE comparison for different cell numbers $\text{T}=3,2,1,0$.}
	\label{tab:cell_increase}       
	\begin{tabular}{cccccccccccc}
		\Xhline{1.3pt}
		\multirow{1}{*}{Method} 
		& \multicolumn{1}{|c}{PSNR $\uparrow$}    & \multicolumn{1}{c}{SSIM $\uparrow$} & \multicolumn{1}{c}{NIQE $\downarrow$}  \\
		\hline
		\multirow{1}{*}{$\text{MANAS}_{\text{T}=3}$} & \multicolumn{1}{|c}{\textbf{32.60}}  & \multicolumn{1}{c}{\textbf{0.922}}  & \multicolumn{1}{c}{\textbf{3.59}}  \\ 
		\hline
		\multirow{1}{*}{$\text{MANAS}_{\text{T}=2}$} & \multicolumn{1}{|c}{31.98}  & \multicolumn{1}{c}{0.913}  & \multicolumn{1}{c}{3.61}   \\
		\hline
		\multirow{1}{*}{$\text{MANAS}_{\text{T}=1}$} & \multicolumn{1}{|c}{31.19}  & \multicolumn{1}{c}{0.904}  & \multicolumn{1}{c}{3.65}  \\
		\hline
		\multirow{1}{*}{$\text{MANAS}_{\text{T}=0}$} & \multicolumn{1}{|c}{25.66}  & \multicolumn{1}{c}{0.825}  & \multicolumn{1}{c}{3.79}   \\
		\Xhline{1.3pt}
	\end{tabular}
\end{table}

\subsubsection{Whether the Multi-to-One Training Strategy Could Make Our Model Robust?}
\begin{table}[t]
	\renewcommand{\arraystretch}{1.2}
	\tabcolsep0.1cm
	\centering
	\caption{Comparison of PSNR, SSIM, and NIQE performances using one-to-one and multi-to-one training strategies.}
	\label{tab:training_paradigm}       
	\begin{tabular}{cccccccccc}
		\Xhline{1.3pt}
		\multirow{1}{*}{Method} & \multicolumn{2}{|c}{PSNR $\uparrow$}    & \multicolumn{2}{c}{SSIM $\uparrow$}  & \multicolumn{2}{c}{NIQE $\downarrow$}   \\
		\hline
		\multirow{1}{*}{One-to-one training}  & \multicolumn{2}{|c}{31.52}  & \multicolumn{2}{c}{0.905}  & \multicolumn{2}{c}{3.62}   \\
		\hline
		\multirow{1}{*}{Multi-to-one training} & \multicolumn{2}{|c}{\textbf{32.60}}  & \multicolumn{2}{c}{\textbf{0.922}}  & \multicolumn{2}{c}{\textbf{3.59}}  \\
		\Xhline{1.3pt}
	\end{tabular}
\end{table}
\begin{figure}[!t]
	\centering
	\subfigure{
		\begin{minipage}[t]{0.23\linewidth}
			\centering
			\centerline{\includegraphics[width=0.82in]{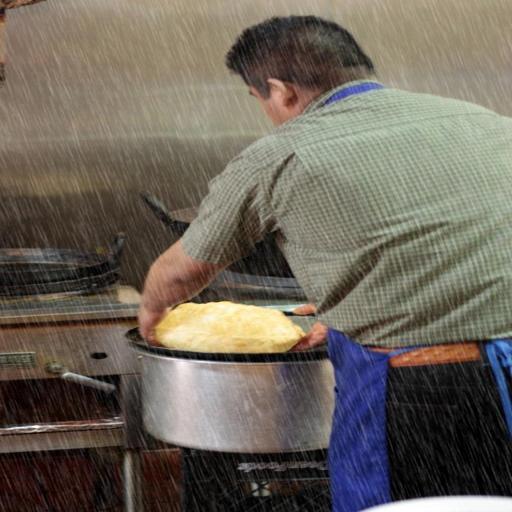}}
			\vspace{2pt}
			\centerline{\includegraphics[width=0.82in]{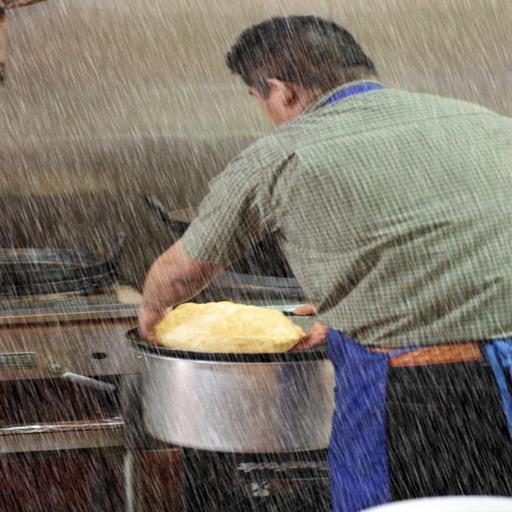}}
			\vspace{2pt}
			\centerline{\includegraphics[width=0.82in]{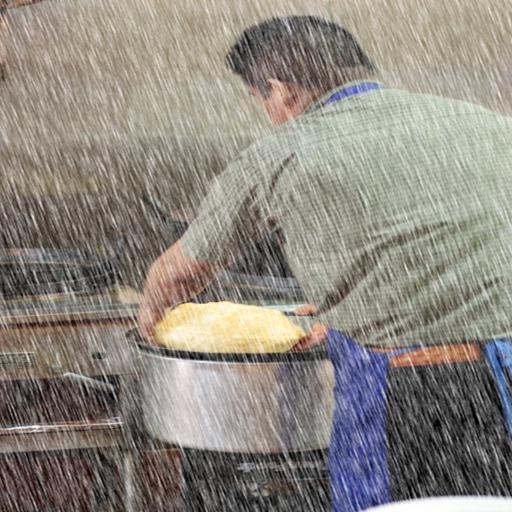}}
			{\scriptsize{Rainy Image}}
		\end{minipage}%
	}%
	\subfigure{
		\begin{minipage}[t]{0.23\linewidth}
			\centering
			\centerline{\includegraphics[width=0.82in]{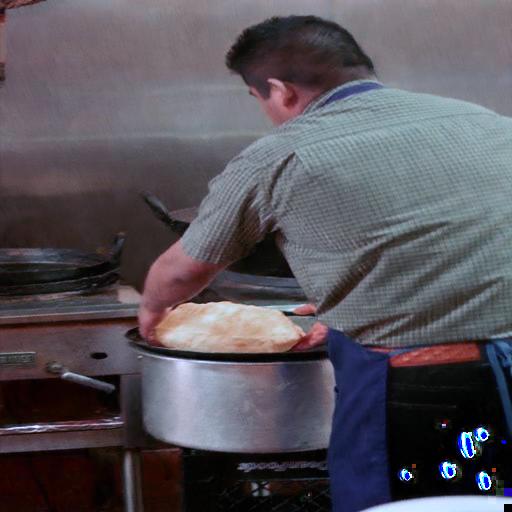}}
			\vspace{2pt}
			\centerline{\includegraphics[width=0.82in]{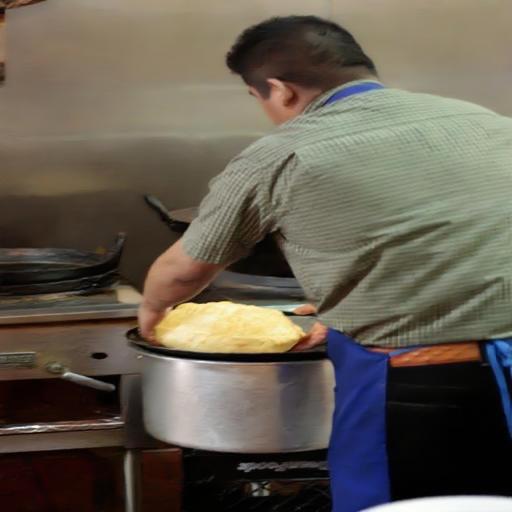}}
			\vspace{2pt}
			\centerline{\includegraphics[width=0.82in]{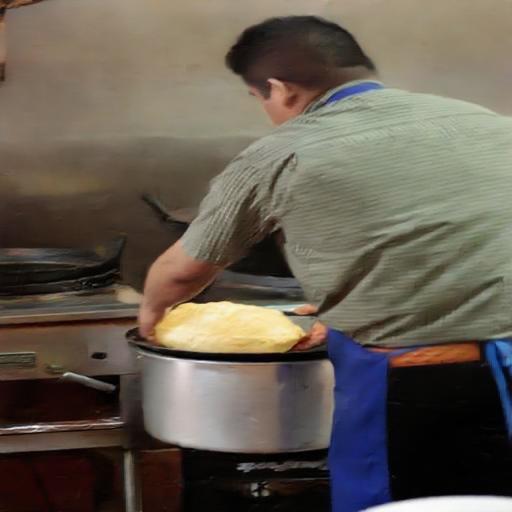}}
			{\scriptsize{One-to-one}}
		\end{minipage}%
	}%
	\subfigure{
		\begin{minipage}[t]{0.23\linewidth}
			\centering
			\centerline{\includegraphics[width=0.82in]{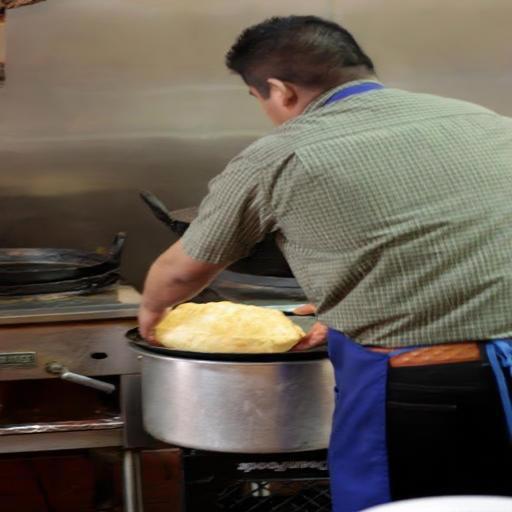}}
			\vspace{2pt}
			\centerline{\includegraphics[width=0.82in]{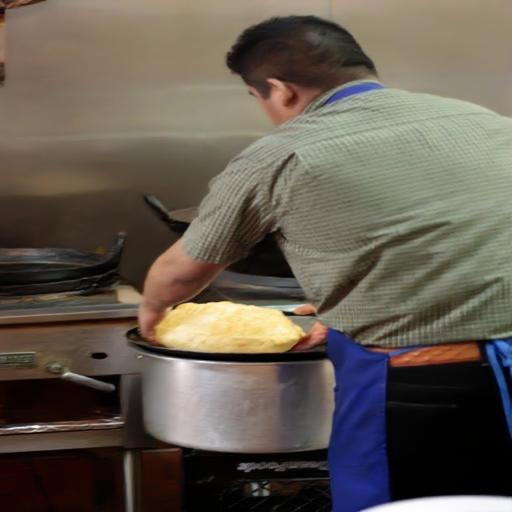}}
			\vspace{2pt}
			\centerline{\includegraphics[width=0.82in]{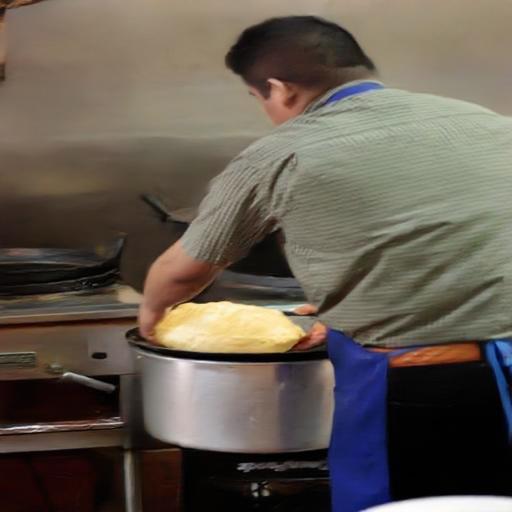}}
			{\scriptsize{Multi-to-one}}
		\end{minipage}%
	}%
	\subfigure{
		\begin{minipage}[t]{0.23\linewidth}
			\centering
			\centerline{\includegraphics[width=0.82in]{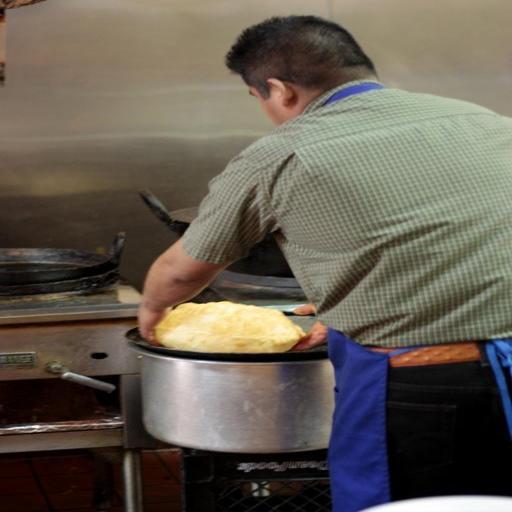}}
			\vspace{2pt}
			\centerline{\includegraphics[width=0.82in]{./Figs/1_gt.jpg}}
			\vspace{2pt}
			\centerline{\includegraphics[width=0.82in]{./Figs/1_gt.jpg}}
			{\scriptsize{Ground Truth}}
		\end{minipage}%
	}%
	\centering
	\caption{Three synthetic rainy images with the same background from DID-MDN~\cite{Zhang2018Density} dataset, and their corresponding de-rained images resulting from one-to-one and multi-to-one training strategies.}
	\label{fig:consistency_2}
\end{figure} 
To answer this question, we also trained our MANAS method with a one-to-one training strategy which was commonly used in previous deep-learning-based de-raining methods. 
The one-to-one training strategy adopts
one-to-one image pairs for training, and each pair contains only one rainy image and one ground truth image, thus this training strategy cannot use the~\emph{internal loss} $\mathcal{L}_{\text{int}}$ like our multi-to-one training strategy.
On the other hand, the one-to-one training paradigm does not mean that fewer rainy samples were used for training. Instead, all of the rainy images in the training set were used. The quantitative comparisons between one-to-one and multi-to-one training strategies are shown in Table.~\ref{tab:training_paradigm}. As can be seen, better performances are obtained by our multi-to-one training strategy. Furthermore, 
in Fig.~\ref{fig:consistency_2}, we showcase three synthetic rainy images that have the same background but are degraded by light rain, medium rain, and heavy rain, respectively (from top to bottom), as well as their corresponding de-rained versions resulting from the one-to-one and our multi-to-one training strategies. We observe that the one-to-one training strategy could lead to parts of the de-rained images presenting serious distortions,~\emph{i.e.}, color inconsistency and local artifacts, while the proposed multi-to-one training strategy can well avoid these issues, see the second and third columns in Fig.~\ref{fig:consistency_2} for a comparison.
Intuitively, this could be due to the fact that the~\emph{internal loss} $\mathcal{L}_{\text{int}}$ in the multi-to-one training strategy can well constrain the de-rained images that have the same background to be close with each other, thus this to some extent prevents the distortion after de-raining. This study shows that our multi-to-one training strategy can indeed make our de-raining model robust, and outperforms the previously commonly-used one-to-one training strategy.


\subsubsection{How Does $\mathcal{L}_{\text{comp}}$ Control the Trade-off Between Performance and Model Size?}
\label{sssec:Loss}

\begin{table}[t]
	\renewcommand{\arraystretch}{1.2}
	\tabcolsep0.1cm
	\centering
	\caption{PSNR, SSIM, and NIQE comparison for different $\lambda_{\text{comp}}$ values.}
	\label{tab:comp}       
	\begin{tabular}{ccccccccccccc}
		\Xhline{1.3pt}
		\multirow{1}{*}{Configurations} & \multicolumn{2}{|c}{PSNR $\uparrow$}  & \multicolumn{2}{c}{SSIM $\uparrow$} & \multicolumn{2}{c}{NIQE $\downarrow$} & \multicolumn{2}{c}{Model Size (M) $\downarrow$}  \\
		\hline
		\multirow{1}{*}{$\lambda_{\text{comp}}=0.00$}  & \multicolumn{2}{|c}{32.60}  & \multicolumn{2}{c}{0.922}  & \multicolumn{2}{c}{3.59} & \multicolumn{2}{c}{8.19}  \\
		\hline
		\multirow{1}{*}{$\lambda_{\text{comp}}=0.01$}  & \multicolumn{2}{|c}{32.58}   & \multicolumn{2}{c}{0.921}  & \multicolumn{2}{c}{3.59} & \multicolumn{2}{c}{7.96}  \\
		\hline
		\multirow{1}{*}{$\lambda_{\text{comp}}=0.10$}  & \multicolumn{2}{|c}{32.37}   & \multicolumn{2}{c}{0.919}  & \multicolumn{2}{c}{3.60} & \multicolumn{2}{c}{6.79}  \\
		\hline
		\multirow{1}{*}{$\lambda_{\text{comp}}=1.00$}  & \multicolumn{2}{|c}{32.14}   & \multicolumn{2}{c}{0.914}  & \multicolumn{2}{c}{3.62} & \multicolumn{2}{c}{5.85} \\
		\Xhline{1.3pt}
	\end{tabular}
\end{table}
To study this point, we set $\lambda_{\text{comp}}$ to different values to verify the role of the model complexity loss,~\emph{i.e.}, $\mathcal{L}_{\text{comp}}$, on balancing the performance and the model size. The corresponding results are shown in Table.~\ref{tab:comp}. It is interesting to see that as we increase the value of $\lambda_{\text{comp}}$, the model size will become smaller, and meanwhile the performance will be worse but still maintain a relatively high quantitative results.

\subsection{Limitations}
\label{ssec:discussion}
Although the proposed MANAS framework can automatically search and integrate the~\emph{multi-scale attentive neural architecture} for image de-raining, there still exist some limitations in this proposed method. For example, the number and width of the multi-scale attentive cells in the de-raining network are required to be set manually. Recently, Fang~\emph{et al.}~\cite{Fang2020Densely} proposed to automatically search block counts and block widths in a dense super network using neural architecture search (NAS). Inspired by their work, the number and width of the multi-scale attentive cells may be able to be searched alternatively or synchronously in the de-raining network. We leave this study as future work. 


\section{Conclusion}
In this work, we proposed a novel~\emph{multi-scale attentive neural architecture search} (MANAS) framework for image de-raining. In brief, we formulated a new~\emph{multi-scale attention search space} consisting of multiple typical basic modules that are beneficial for image de-raining. Meanwhile, we integrated the multi-scale attention search space into a differentiable form through a continuous relaxation operation, and exploited a gradient-based search algorithm to automatically search the internal~\emph{multi-scale attentive architecture} for the de-raining network. On the other hand, to generate a robust de-raining model, we also specifically devised a practical and effective~\emph{multi-to-one training strategy} for image de-raining. We conducted extensive image de-raining experiments on both synthetic and realistic datasets, and also applied our MANAS method to high-level vision applications,~\emph{i.e.}, object detection and segmentation. The experimental results consistently demonstrate the superiority of our method.





%

\appendices




\ifCLASSOPTIONcaptionsoff
  \newpage
\fi



%
\bibliographystyle{IEEEbib}

\bibliography{IRNet}

\begin{IEEEbiography}[{\includegraphics[width=1in,height=1.25in,clip,keepaspectratio]{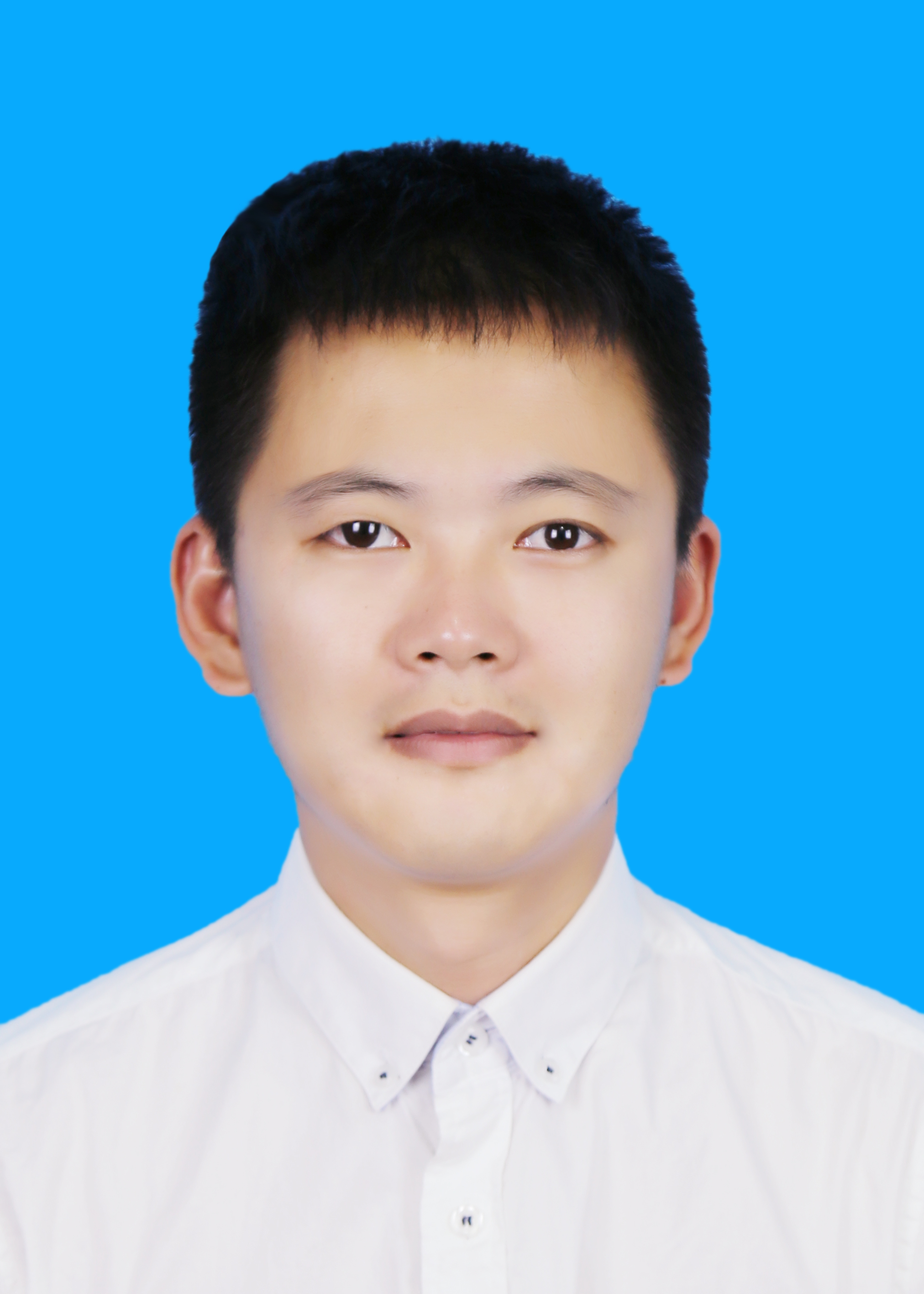}}]{Lei Cai} received the M.S. degree in communication and information system from the School of Information Science and Engineering, Huaqiao University, Xiamen, China, in 2018. He is currently pursuing the Ph.D degree at the School of Electronic and Information Engineering, South China University of Technology. His research interests include pedestrian gender recognition, compressed sensing, and image restoration.
\end{IEEEbiography}

\begin{IEEEbiography}[{\includegraphics[width=1in,height=1.25in,clip,keepaspectratio]{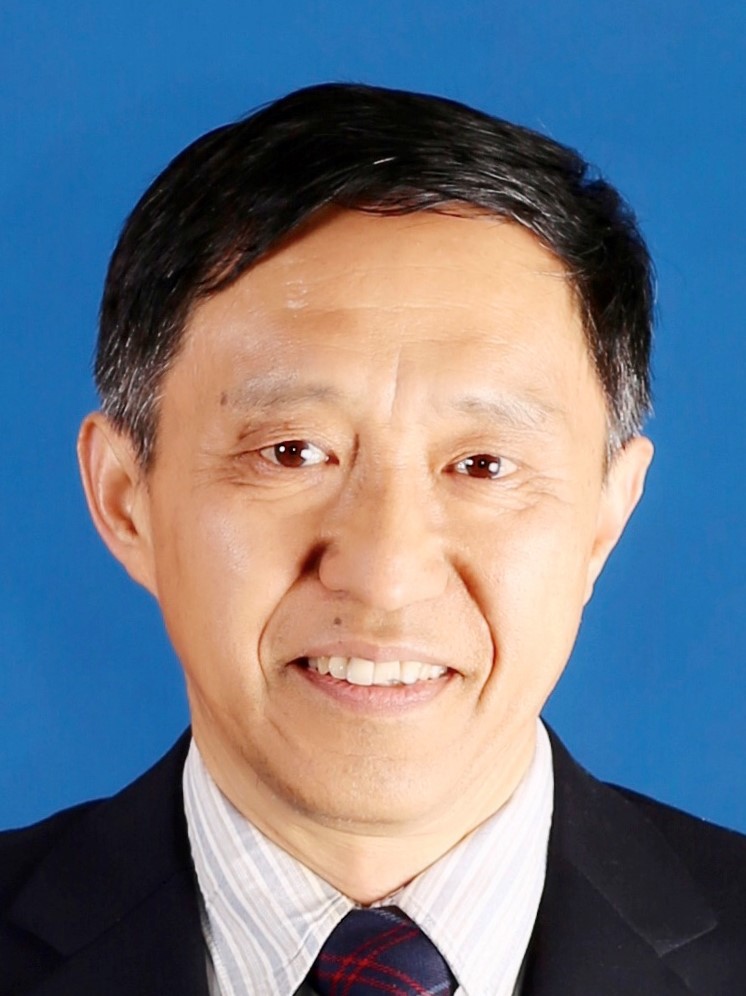}}]{Yuli Fu} received the Ph.D. degree from the Huazhong University of Science and Technology, Wuhan, China, in 2000. He is currently a Full Professor with the School of Electronic and Information Engineering, South China University of Technology, Guangzhou, China. He has authored or coauthored about 80 journal papers on control theory, neural networks and signal processing. His current research interests include adaptive signal processing, artificial intelligence, and pattern recognition.
\end{IEEEbiography}

\begin{IEEEbiography}[{\includegraphics[width=1in,height=1.25in,clip,keepaspectratio]{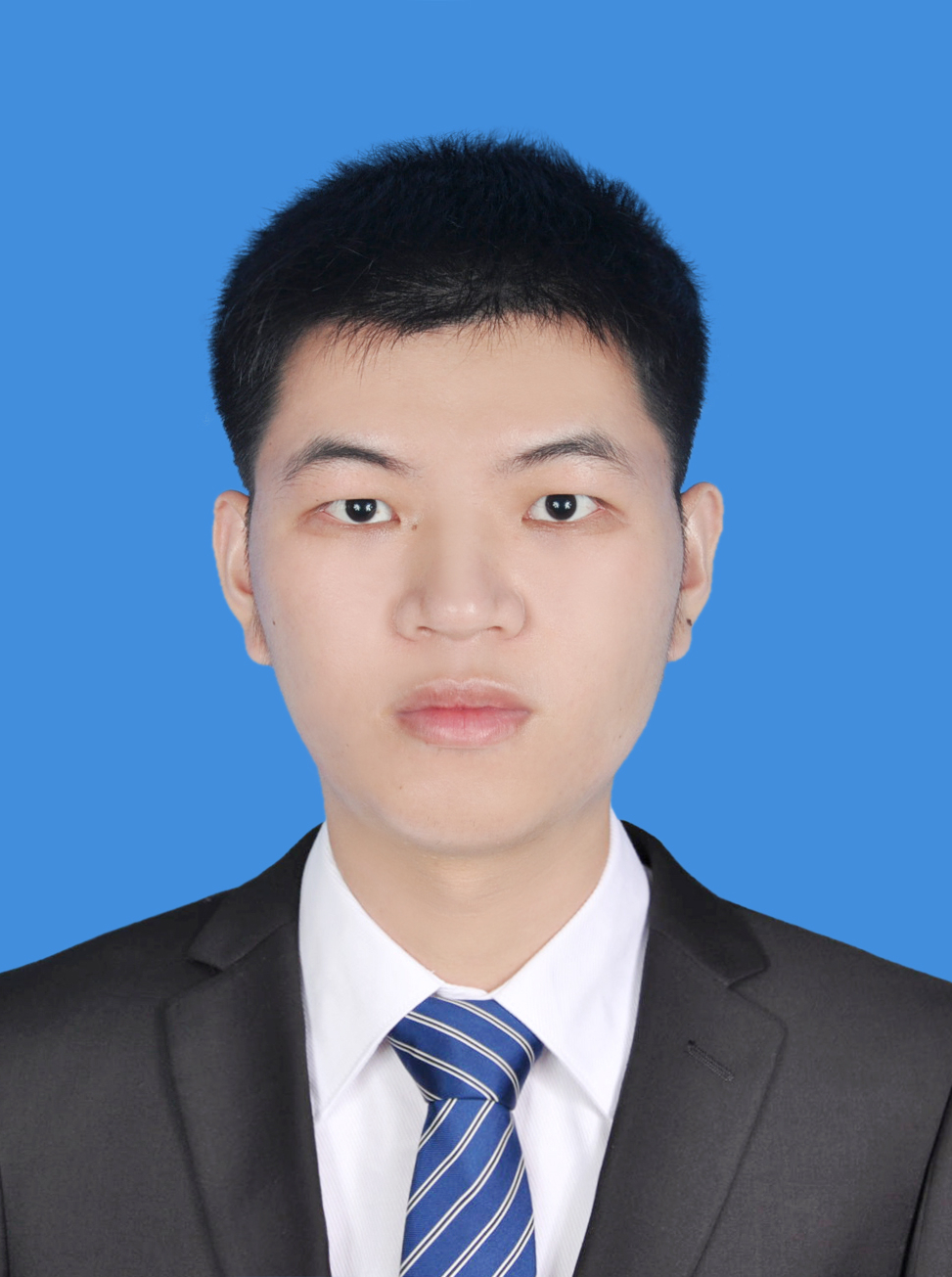}}]{Wanliang Huo} received the B.E. degree from the South China University of Technology, Guangzhou, China, in 2020. He is currently pursuing the M.S. degree at the School of Electronic and Information Engineering. His research interests include machine learning and computer vision, particularly focusing on the image de-raining problem. 
\end{IEEEbiography}

\begin{IEEEbiography}[{\includegraphics[width=1in,height=1.25in,clip,keepaspectratio]{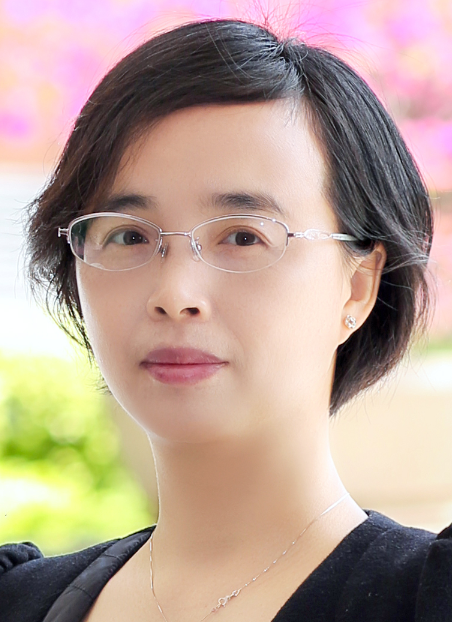}}]{Youjun Xiang} received the M.S. degree in pattern recognition and intelligent system from Xidian University, Xi'an, China, and the Ph.D. degree in signal and information processing from South China University of Technology, Guangzhou, China. She is currently an Associate Professor with the South China University of Technology. Her research interests include video coding, artificial intelligence, and pattern recognition.
\end{IEEEbiography}

\begin{IEEEbiography}[{\includegraphics[width=1in,height=1.25in,clip,keepaspectratio]{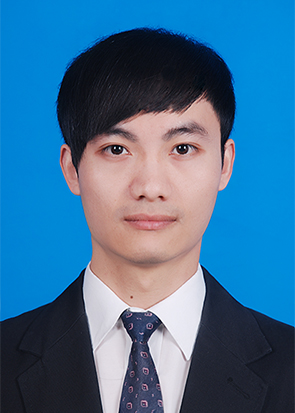}}]{Tao Zhu} received the Ph.D degree in signal and information processing from the School of Electronic and Information Engineering, South China University of Technology, Guangzhou, China, in 2021. His research interests include inverse problem, dictionary learning, and independent component analysis.
\end{IEEEbiography}

\begin{IEEEbiography}[{\includegraphics[width=1in,height=1.25in,clip,keepaspectratio]{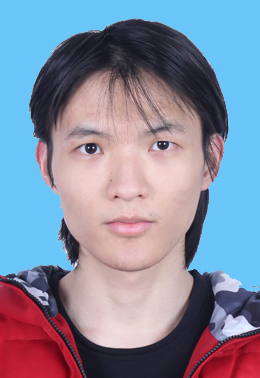}}]{Ying Zhang} received the B.E. degree from the South China University of Technology, Guangzhou, China, in 2020. He is currently pursuing the M.S. degree at the School of Electronic and Information Engineering, South China University of Technology. His research interests include image de-raining and related image restoration problems. 
\end{IEEEbiography}

\begin{IEEEbiography}[{\includegraphics[width=1in,height=1.25in,clip,keepaspectratio]{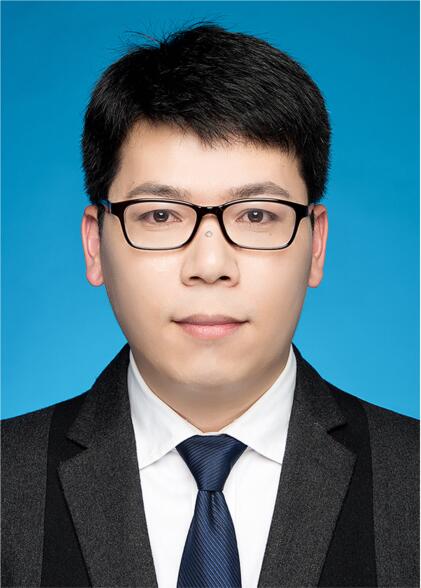}}]{Huanqiang Zeng} (S'10-M'13-SM'18)
	received the B.S. and M.S. degrees in electrical engineering from Huaqiao University, China, and the Ph.D. degree in electrical engineering from Nanyang Technological University, Singapore.
	
	He is currently a Full Professor at the School of Engineering and the School of Information Science and Engineering, Huaqiao University. Before that, he was a Postdoctoral Fellow at The Chinese University of Hong Kong, Hong Kong. He has published more than 100 papers in well-known journals and conferences, including three best poster/paper awards (in the International Forum of Digital TV and Multimedia Communication 2018 and the Chinese Conference on Signal Processing 2017/2019). His research interests include image processing, video coding, machine learning, and computer vision. He has also been actively serving as the General Co-Chair for IEEE International Symposium on Intelligent Signal Processing and Communication Systems 2017 (ISPACS2017), the Co-Organizer for ICME2020 Workshop on 3D Point Cloud Processing, Analysis, Compression, and Communication, the Technical Program Co-Chair for Asia–Pacific Signal and Information Processing Association Annual Summit and Conference 2017 (APSIPA-ASC2017), the Area Chair for IEEE International Conference on Visual Communications and Image Processing (VCIP2015 and VCIP2020), and a technical program committee member for multiple flagship international conferences. He has been actively serving as an Associate Editor for IEEE TRANSACTIONS ON IMAGE PROCESSING, IEEE TRANSACTIONS ON CIRCUITS AND SYSTEMS FOR VIDEO TECHNOLOGY, and Electronics Letters (IET). He has been actively serving as a Guest Editor for Journal of Visual Communication and Image Representation, Multimedia Tools and Applications, and Journal of Ambient Intelligence and Humanized Computing.
\end{IEEEbiography}

\begin{IEEEbiography}[{\includegraphics[width=1in,height=1.25in,clip,keepaspectratio]{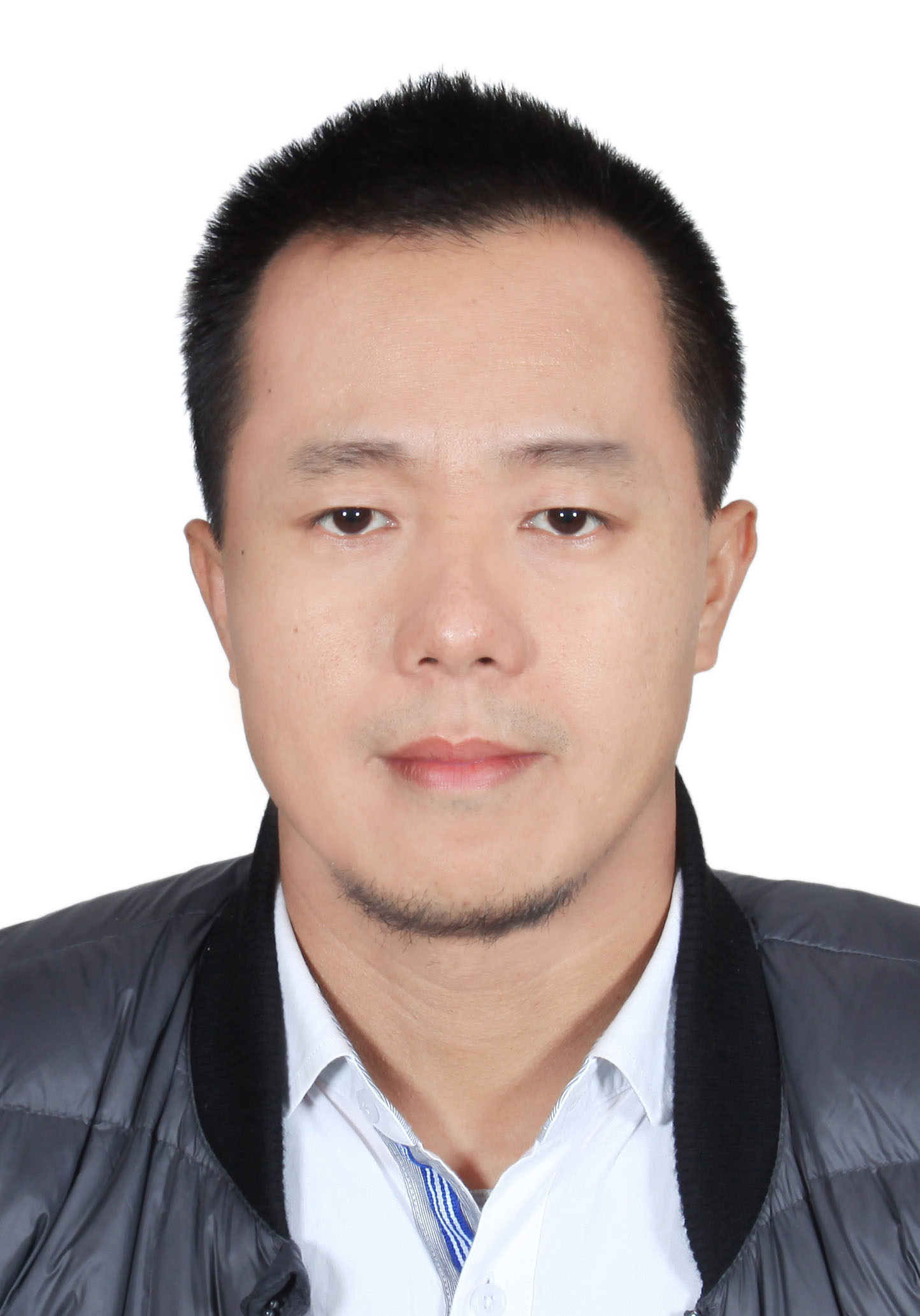}}]{Delu Zeng} received his Bachelor degree in applied mathematics and his Phd. degree in signal and information processing in South China University of Technology (SCUT) in June 2003 and June 2010, respectively. He has been visiting scholar in Columbia University, University of Waterloo, University of Oulu. And He is currently a full professor with the School of Electronic and Information Engineering in South China University of Technology (SCUT) in Guangzhou. His research interests focus on statistics learning, image and speech processing, computational intelligence, machine learning, fitting and approximation, and their applications to communications, industrial intelligence, etc.. 
\end{IEEEbiography}

%








\end{document}